\documentclass[a4paper,10pt]{elsarticle}
\usepackage{lineno,hyperref}
\usepackage{amsmath}
\usepackage{amsfonts}
\usepackage{amssymb}
\usepackage{pdflscape}
\usepackage{threeparttable}
\usepackage{mathrsfs}
\usepackage{bm}
\usepackage{geometry}
\usepackage{rotating}

\usepackage[dvips]{epsfig}
\usepackage{epsfig, epsf}
\usepackage[dvips]{color}

\usepackage{algorithm}
\usepackage{algorithmic}
\usepackage{multirow}

\usepackage{longtable}

\usepackage{booktabs}

\usepackage{subfig}

\modulolinenumbers[5]

\journal{European Journal of Operational Research}

\newdefinition{definition}{Definition}

\newproof{pf}{Proof}

\usepackage{setspace}
\begin{document}
	
	\begin{frontmatter}
		
		\title{A preference learning framework for multiple criteria sorting\\ with diverse additive value models and valued assignment examples}
		
		\author[mymainaddress]{Jiapeng Liu\corref{mycorrespondingauthor}}
		\cortext[mycorrespondingauthor]{Corresponding author}
		\ead{jiapengliu@mail.xjtu.edu.cn}
		
		\author[mysecondaddress]{Mi{\l}osz Kadzi{\'n}ski}
		\ead{milosz.kadzinski@cs.put.poznan.pl}
		
		\author[mymainaddress]{Xiuwu Liao}
		\ead{liaoxiuwu@mail.xjtu.edu.cn}
		
		\author[mymainaddress]{Xiaoxin Mao}
		\ead{maoxiaoxin29@stu.xjtu.edu.cn}	
		
		\author[mymainaddress]{Yao Wang}
		\ead{yao.s.wang@gmail.com}	
		
		\address[mymainaddress]{Center of Intelligent Decision-making and Machine Learning, School of Management, Xi'an Jiaotong University, Xi'an, 710049, Shaanxi, P.R. China}
		
		\address[mysecondaddress]{Institute of Computing Science, Poznan University of Technology, Piotrowo 2, 60-965 Pozna{\'n}, Poland}
		
		\begin{abstract}
			We present a preference learning framework for multiple criteria sorting. We consider sorting procedures applying an additive value model with diverse types of marginal value functions (including linear, piecewise-linear, splined, and general monotone ones) under a~unified analytical framework. Differently from the existing sorting methods that infer a preference model from crisp decision examples, where each reference alternative is assigned to a unique class, our framework allows to consider valued assignment examples in which a reference alternative can be classified into multiple classes with respective credibility degrees. We propose an optimization model for constructing a preference model from such valued examples by maximizing the credible consistency among reference alternatives. To improve the predictive ability of the constructed model on new instances, we employ the regularization techniques. Moreover, to enhance the capability of addressing large-scale datasets, we introduce a state-of-the-art algorithm that is widely used in the machine learning community to solve the proposed optimization model in a computationally efficient way. Using the constructed additive value model, we determine both crisp and valued assignments for non-reference alternatives. Moreover, we allow the Decision Maker to prioritize importance of classes and give the method a flexibility to adjust classification performance across classes according to the specified priorities. The practical usefulness of the analytical framework is demonstrated on a~real-world dataset by comparing it to several existing sorting methods.		
		\end{abstract}
		
		
	\end{frontmatter}

	\section{Introduction}
	
	\noindent With a~rapid development of information technology, organizations have accumulated and stored a vast quantity of data from various sources, such as manufacturing, marketing, finance, tourism, agriculture, transportation, or ecosystem. The availability of data resources helps organizations mine useful information and make better informed decisions, including optimizing operations, deepening customers engagement, preventing threats and fraud, and capitalizing on new sources of revenue. Many of such decisions concern classification of a set of alternatives into pre-defined and preference-ordered classes according to their evaluations on multiple criteria. Such a scenario is of interest in \emph{sorting} or ordinal classification problems~\cite{doumpos2002multicriteria, greco2010multiple}. For example, in the field of credit rating, financial institutions predict the credit risk of a prospective debtor (e.g., an individual, a company, or a government) and assign a grade (e.g., from AAA to B- in Standard \& Poor's) to each debtor, where grades are intended to represent probability of default. Another example comes from medical diagnostics, where doctors evaluate physical conditions of patients and classify them into groups representing different disease grades according to the observed symptoms.
	
	The practical significance of sorting problems has motivated researchers to develop multiple streams of methods for addressing such problems, including (a) value-driven methods (e.g., \cite{doumpos2002multicriteria,greco2010multiple,kadzinski2015modeling,kadzinski2020contingent}), (b) outranking-based methods (e.g., \cite{almeida2010electre, corrente2016hierarchy, dias2002aggregation, pelissari2019flowsort, tervonen2009stochastic}), and (c) rule induction-oriented models (e.g., \cite{greco2001drsa, kadzinski2016robustness}). In this paper, we focus on the value-driven sorting procedure, which employs a value function as the preference model and assigns a~numerical score to each alternative by aggregating its performances on different criteria. The value function model is widely used and highly appreciated by the Multiple Criteria Decision Aiding (MCDA) community due to its relatively easy computation and intuitive interpretation \cite{kadzinski2015modeling}. In the sorting context, the assignment of an alternative is usually determined by comparing its value to thresholds that explicitly delimit consecutive classes (threshold-based sorting procedure, see \cite{doumpos2002multicriteria}) or reference alternatives that implicitly characterize each class (example-based sorting procedure, see \cite{greco2010multiple}).
	
	Under the assumption on the preferential independence of criteria, the value of an alternative can be expressed as the sum of marginal value functions on each criterion \cite{keeney1976decisions, sobrie2018uta}, and such a preference model is called \emph{additive value function}. There are various types of additive value models for characterizing the preferences over alternatives on the individual criteria. A basic form of an additive-based value model is composed of linear marginal value functions, where all marginal values are defined as linear functions on the performance ranges of individual criteria. Such a value function can be seen as a simple weighted average aggregation model and is relatively easy to explain to a non-experienced Decision Maker (DM). However, the ability of such a value function for addressing complex decision structure is rather limited due to incorporating an assumption on the linear form. Another way of modeling marginal values is to use piecewise-linear marginal value functions, which have been used in the UTADIS family~\cite{doumpos2002multicriteria}. The advantage of using piecewise-linear marginal value function consists in the possibility of reflecting various decision policies such as risk aversion or risk seeking attitude. This capability enhances its appropriateness and practical usefulness for a wide range of applications \cite{liu2019preference}. Nevertheless, such a type of value function is criticized for its lack of smoothness, which may cause a~sudden change in slope at breakpoints and hence limits their interpretability in some contexts \cite{sobrie2018uta}. To overcome the shortcoming of piecewise-linear marginal value functions, \cite{sobrie2018uta} proposed to use cubic splines for constructing marginal value functions. A cubic spline is a set of polynomials of degree three, which is continuous and has continuous first- and second-order derivatives at breakpoints \cite{hastie2009elements,sobrie2018uta}. The continuity character of splined marginal value functions makes them advantageous in terms of interpreting human preferences. Another appreciated model is the general monotone value function, which is defined by marginal values at characteristic points corresponding to all unique performance levels. Such a preference model makes only the monotonicity assumption on general shape of marginal value functions, and therefore proves to be the most flexible value function model to represent human preferences \cite{kadzinski2017expressiveness}.
	
	This paper introduces a new analytical framework for multiple criteria sorting problems. We consider linear, piecewise-linear, splined, and general monotone value functions under a~unified framework, in which the DM is allowed to refer to a desired type of value function. We aim to learn an additive value model from a given set of holistic decision examples (also called training samples) composed of a set of reference alternatives and their desired assignments. The latter ones could come from past decisions provided by the DM, such as historical credit rating reports or past patient classification records. Differently from the existing sorting methods that consider crisp assignments, where each reference alternative is definitely assigned to a unique class, our framework allows for taking into account valued decision examples, in which each reference alternative can be classified into multiple classes with respective credibility degrees~\cite{cailloux2012}. Such an imprecise assignment has many potential applications in business and management (e.g., funds granting, credit approval, medical diagnostics), when the DM is unconfident about the desired assignments of alternatives or the collected information is not fully credible. To learn a value function model from valued decision examples, we investigate the preference relation for any pair of reference alternatives by considering each alternative's possible assignment outcome, and then propose an original optimization model to simultaneously account for different objectives with respective credibility degrees concerning each possible preference relation for a pair of alternatives. The targets involved in the proposed optimization model definitely enhance value difference between a pair of reference alternatives such that one alternative is always assigned to a class better than the other, and/or equalize values of a pair of reference alternatives, with a certain credibility, such that they could be classified into the same class. In this way, the constructed preference model highlights the most certain part in the valued decision examples, and keeps a vague preference relation for a pair of alternatives among which one cannot indicate a better one.
	
	In this work, learning a value function model from valued decision examples is formulated within the regularization framework by considering both the model's fitting ability and its complexity simultaneously. In the context of valued assignment, the fitting ability of a value function model is measured by accounting for the credible difference between the comprehensive values for all pairs of reference alternatives. In other words, a~``best-fit'' value function model should be as credibly consistent with the preference relations between reference alternatives as possible. However, a value function model that ``best-fits'' the given decision examples can be very complex and may encounter the \emph{over-fitting} problem, that is, the constructed model fits the decision examples well but has poor generalization performance on new instances. For a comprehensive discussion on this issue, one can refer to \cite{liu2019preference}. To improve the predictive ability of a value function model, \cite{dembczynski2006additive,doumpos2007regularized,liu2019preference} introduce regularization terms for controlling the model's complexity and deriving a simple value function model while maintaining its fitting ability. From the viewpoint of the statistical learning theory \cite{vapnik1998statistical, murphy2012machine}, a proper complexity control contributes to avoiding the over-fitting problem and improving the model's generalization ability. In this paper, as the analytical framework is applicable to linear, piecewise-linear, splined, and general monotone value functions, we define the complexity measure for each type of value function model and formulate the learning problem in the unified regularization framework. In this way, the constructed value function model makes a trade-off between the fitting ability and the model's complexity, and improves its generalization ability on new instances. Apart from the methodological advance from the statistical learning theory, we also introduce a state-of-the-art algorithm named the \emph{alternating direction method of multipliers} (ADMM) \cite{boyd2011distributed} to address the learning problem and improve computational efficiency for dealing with large-scale datasets. Using ADMM, the learning problem is solved by decomposing the original problem into a series of small-size optimization problems, which can be easily addressed without extraordinary efforts. Moreover, the implementation of ADMM for the learning problem is well suited to distributed optimization, and has the advantage of parallel computation.
	
	Once a value function model is constructed from the given valued decision examples, we can use the constructed preference model to predict the assignment for a new alternative. In this paper, we provide two types of assignments in this context: crisp and valued. The crisp assignment specifies a class to which the alternative can be assigned so that the greatest credible consistency between this alternative and all reference alternatives would be obtained. The valued assignment associates a credibility degree with each class for the alternative which is derived by accounting for the credible consistency between this alternative and all reference alternatives when this alternative is put in each class. 
	
	In addition, we consider a complementary component in the analytical framework which allows to adjust classification performance across classes. The appeal of such a component stems from the fact that in may real-world applications the DM may want to prioritize the importance for classes. For example, in medical diagnostics where a doctor aims to classify patients into the ``healthy'' and ``unhealthy'' groups, the ``unhealthy'' group is prior to the ``healthy'' group although the latter is preferred to the former. This is due to that the doctor usually hopes to achieve as high classification performance as possible on the ``unhealthy'' group, because incorrect prediction will delay necessary treatment for patients. On the other hand, the classification performance on the ``healthy'' group is relatively less important, because classifying a healthy person as unhealthy only results in more medical examination for her/him. To implement flexible adjustment of classification performance across classes, we require the DM to specify a priority ranking of all classes, rather than precise values of priorities for each class, which therefore is less demanding in terms of the required cognitive effort. Then, we discuss a method for adjusting a~classification performance across classes according to the specified priority ranking. This method pays more attention to classes with greater priorities and enhances the credible consistency between reference alternatives that can be assigned to these classes and other reference alternatives so that the classification performance on these classes is improved.
	
	We validate the classification performance of four variants of the analytical framework using linear, piecewise-linear, splined, and general monotone value functions on a real-world dataset in terms of Top-$N$ accuracies and Kendall's tau coefficient. Specifically, we examine these measures achieved by the four variants on valued decision examples with different credibility distribution which are generated by simulated value functions with different complexity. Then, we investigate the ability of the proposed method for adjusting classification performance across classes according to the specified priority ranking of the classes.
	
	The remainder of this paper is organized in the following way. In Section \ref{sec-2}, we present the analytical framework for learning diverse types of value function models from valued decision examples and give the flexible method for adjusting classification performance across classes. In Section \ref{sec-3}, we apply the analytical framework to a real-world dataset. Section \ref{sec-4} concludes and discusses future work for this study.

	\section{The analytical framework for multiple criteria sorting problems}
	\label{sec-2}
	
	\subsection{Additive value functions composed of linear, piecewise-linear, splined and general marginal value functions}
	\label{sec-21}
	
	\noindent Let us consider a decision problem regarding $m$ alternatives $A = \left\{ {{a_1},...,{a_m}} \right\}$ evaluated in terms of $n$ criteria $G = \left\{ {{g_1},...,{g_n}} \right\}$. Each criterion $g_j \in G$ is used to assess an alternative $a \in A$ from a certain perspective, and the performance of $a$ on $g_j$ is denoted by $g_j(a)$. All criteria are assumed to be monotone (gain- or cost-type), i.e., for any alternative $a$, either the greater $g_j(a)$, the better is $a$ on $g_j$ (in case of gain-type criteria), or the less $g_j(a)$, the better is $a$ on $g_j$ (in case of cost-type criteria). For dealing with non-monotonic criteria, see \cite{ghaderi2017linear,liu2019preference,rezaei2018nonlinear}. For the sake of simplicity, but without loss of generality, we suppose that all criteria are of gain-type and that the performances on criteria have a monotone increasing direction of preferences. Let ${X_j} = \left[ {{\alpha _j},{\beta _j}} \right]$ be the bounded interval of the performances on criterion $g_j$, where $\alpha_j$ and $\beta_j$ are the worst and best performances, respectively. For any alternative $a \in A$, we shall use a value function in the following additive form as the preference model to aggregate the performances of $a$ on multiple criteria \cite{keeney1976decisions}:
	\begin{equation*}
	U\left( a \right) = \sum\limits_{j = 1}^n {{u_j}\left( {{g_j}\left( a \right)} \right)},
	\end{equation*}
	where ${u_j}\left( \cdot \right)$, $j=1,...,n$, are monotone non-decreasing marginal value functions. The value function assigns a~numerical score to each alternative, which is used to represent its comprehensive value and impose a preference relation on the set of alternatives. 
	
	The analytical framework introduced in this paper admits various types of marginal value functions including (a) linear, (b) piecewise-linear, (c) splined shaped, and (d) general monotone ones. In case marginal value functions are assumed to be linear, ${u_j}\left(  \cdot  \right)$, $j=1,...,n$, can be constructed as follows:
	\begin{equation*}
	{u_j}\left( x \right) = {w_j}\frac{{x - {\alpha _j}}}{{{\beta _j} - {\alpha _j}}},\;\;\;\;x \in \left[ {{\alpha _j},{\beta _j}} \right],
	\end{equation*}
	where $w_j = {u_j}\left( {{\beta _j}} \right)$ is the maximal share of each criterion $g_j$ in the comprehensive value and can be understood as a trade-off weight of $g_j$. To normalize the value function within the interval [0,1], we usually require
	\begin{equation*}
	\left. \begin{gathered}
	{w_j} \ge 0,\;\;j = 1,...,n, \hfill \\
	\sum\limits_{j = 1}^n {{w_j}}  = 1. \hfill 
	\end{gathered}  \right\}E_{{\text{BASE}}}^{{\text{LINEAR}}}
	\end{equation*}
	Note that, for linear marginal value functions ${u_j}\left(  \cdot  \right)$, $j=1,...,n$, the vector ${\bm{\theta }} = {\left( {{w_1},...,{w_n}} \right)^{\text{T}}}$ is the only parameter to be estimated. In particular, the comprehensive value of alternative $a$ can be formulated with respect to ${\bm{\theta }}$ as $U\left( a \right) = {{\bm{\theta }}^{\text{T}}}{\mathbf{V}}\left( a \right)$, where ${\mathbf{V}}\left( a \right) = {\left( {\frac{{{g_1}\left( a \right) - {\alpha _1}}}{{{\beta _1} - {\alpha _1}}},...,\frac{{{g_n}\left( a \right) - {\alpha _n}}}{{{\beta _n} - {\alpha _n}}}} \right)^{\text{T}}}$.
	
	In the piecewise-linear case, the performance scale $\left[ {{\alpha _j},{\beta _j}} \right]$ on each criterion $g_j$, $j=1,...,n$, is divided into a number of sub-intervals, and marginal value functions ${u_j}\left(  \cdot  \right)$ are assumed to be linear over each sub-interval. Suppose that the performance scale $\left[ {{\alpha _j},{\beta _j}} \right]$ on criterion $g_j$ is divided into ${\gamma _j}$ equal-length sub-intervals $\left[ {x_j^0,x_j^1} \right]$, $\left[ {x_j^1,x_j^2} \right]$, ..., $\left[ {x_j^{{\gamma _j} - 1},x_j^{{\gamma _j}}} \right]$, where each breakpoint is given by $x_j^k = {\alpha _j} + \frac{k}{{{\gamma _j}}}\left( {{\beta _j} - {\alpha _j}} \right)$, $k = 0,1,...,{\gamma _j}$. Such a~way of defining sub-intervals is easy to implement (for other techniques, see \cite{kadzinski2017expressiveness}). Then, the marginal value corresponding to the performance ${g_j}\left( a \right) \in \left[ {x_j^k,x_j^{k + 1}} \right]$, $k = 0,1,...,{\gamma _j} - 1$, is defined with linear interpolation:
	\begin{equation*}
	{u_j}\left( {{g_j}\left( a \right)} \right) = {u_j}\left( {x_j^k} \right) + \frac{{{g_j}\left( a \right) - x_j^k}}
	{{x_j^{k + 1} - x_j^k}}\left( {{u_j}\left( {x_j^{k + 1}} \right) - {u_j}\left( {x_j^k} \right)} \right).
	\end{equation*}
	Therefore, once the marginal values at breakpoints (i.e., ${u_j}\left( {x_j^0} \right) = {u_j}\left( {{\alpha _j}} \right)$, ${u_j}\left( {x_j^1} \right)$, ..., ${u_j}\left( {x_j^{{\gamma _j}}} \right) = {u_j}\left( {{\beta _j}} \right)$) are estimated, we can fully specify piecewise-linear marginal value functions ${u_j}\left(  \cdot  \right)$, $j=1,...,n$. Let $\Delta u_j^k = {u_j}\left( {x_j^k} \right) - {u_j}\left( {x_j^{k - 1}} \right)$, $k = 1,...,{\gamma _j}$, and then the marginal value corresponding to the performance ${g_j}\left( a \right) \in \left[ {x_j^k,x_j^{k + 1}} \right]$, $k = 0,1,...,{\gamma _j} - 1$, can be reformulated as ${u_j}\left( {{g_j}\left( a \right)} \right) = \sum\limits_{t = 1}^k {\Delta u_j^t}  + \frac{{{g_j}\left( a \right) - x_j^k}}
	{{x_j^{k + 1} - x_j^k}}\Delta u_j^{k + 1}$. To normalize the value function within the interval [0,1], one can consider the following linear constraints:
	\begin{equation*}
	\left. \begin{gathered}
	\Delta u_j^k \geqslant 0,\;\;k = 1,...,{\gamma _j},\;j = 1,...,n, \hfill \\
	\sum\limits_{j = 1}^n {\sum\limits_{k = 1}^{{\gamma _j}} {\Delta u_j^k} }  = 1, \hfill  
	\end{gathered}  \right\}E_{{\text{BASE}}}^{{\text{PIECEWISE - LINEAR}}}
	\end{equation*}
	where $\sum\limits_{k = 1}^{{\gamma _j}} {\Delta u_j^k}$ is the maximal share of marginal value $u_j \left( \cdot \right)$ in the comprehensive value, which can be interpreted as a trade-off weight of marginal value function $u_j \left( \cdot \right)$. Let ${\bm{\theta }} = {\left( {{\bm{\theta }}_1^{\text{T}},...,{\bm{\theta }}_n^{\text{T}}} \right)^{\text{T}}}$, ${{\bm{\theta }}_j} = {\left( {\Delta u_j^1,...,\Delta u_j^{{\gamma _j}}} \right)^{\text{T}}}$ for $j=1,...,n$, and ${\mathbf{V}}\left( a \right) = {\left( {{{\mathbf{V}}_1}{{\left( a \right)}^{\text{T}}},...,{{\mathbf{V}}_n}{{\left( a \right)}^{\text{T}}}} \right)^{\text{T}}}$, ${{\mathbf{V}}_j}\left( a \right) = {\left( {\underbrace {1,...,1,}_{{k_j}\left( a \right)}\frac{{{g_j}\left( a \right) - x_j^{{k_j}\left( a \right)}}}{{x_j^{{k_j}\left( a \right) + 1} - x_j^{{k_j}\left( a \right)}}},\underbrace {0,...,0}_{{\gamma _j} - {k_j}\left( a \right) - 1}} \right)^{\text{T}}}$ where ${k_j}\left( a \right) \in \left\{ {0,1,...,{\gamma _j} - 1} \right\}$ such that ${g_j}\left( a \right) \in \left[ {x_j^{{k_j}\left( a \right)},x_j^{{k_j}\left( a \right) + 1}} \right]$, for $j=1,...,n$. Then, in terms of ${\bm{\theta }}$ and ${\mathbf{V}}\left( a \right)$, the comprehensive value of $a$ can be formulated as $U\left( a \right) = {{\bm{\theta }}^{\text{T}}}{\mathbf{V}}\left( a \right)$. Note that piecewise-linear marginal value functions with a sufficiently large number of sub-intervals can approximate any non-linear value function \cite{liu2019preference}. As~we use piecewise-linear marginal value functions to approximate the actual value function, rather than making assumptions on its form, piecewise-linear marginal value functions can be seen as a non-parametric method for modeling preferences.
	
	One noticeable disadvantage of piecewise-linear marginal value functions consists in the lack of smoothness, which causes a~sudden change in slope at breakpoints and hence limits their interpretability in some contexts~\cite{sobrie2018uta}. An alternative way of constructing ``natural'' marginal value function is to use cubic smoothing spline, which is continuous and has continuous first- and second-order derivatives at breakpoints \cite{hastie2009elements,sobrie2018uta}. Let the performance scale $\left[ {{\alpha _j},{\beta _j}} \right]$ on each criterion $g_j$, $j=1,...,n$, be divided into $\gamma_j$ equal-length sub-intervals $\left[ {x_j^0,x_j^1} \right]$, $\left[ {x_j^1,x_j^2} \right]$, ..., $\left[ {x_j^{{\gamma _j} - 1},x_j^{{\gamma _j}}} \right]$, where $x_j^k = {\alpha _j} + \frac{k}{{{\gamma _j}}}\left( {{\beta _j} - {\alpha _j}} \right)$, $k = 0,1,...,{\gamma _j}$. A cubic smoothing splined marginal value function ${u_j}\left(  \cdot  \right)$ is a piecewise-polynomial of order three, where the $k$-th polynomial over the sub-interval $\left[ {x_j^{k - 1},x_j^k} \right]$, $k=1,...,{\gamma _j}$ has the following form:
	\begin{equation*}
	S_j^k\left( x \right) = s_j^{k,0} + s_j^{k,1}x + s_j^{k,2}{x^2} + s_j^{k,3}{x^3},\;\;x \in \left[ {x_j^{k - 1},x_j^k} \right],
	\end{equation*}
	where $s_j^{k,0}$, $s_j^{k,1}$, $s_j^{k,2}$ and $s_j^{k,3}$ are parameters that need to be determined. Then, marginal value function ${u_j}\left(  \cdot  \right)$ can be formulated as:
	\begin{equation*}
	{u_j}\left( x \right) = \sum\limits_{k = 1}^{{\gamma _j}} {{\mathcal{I} }\left( {x \in \left[ {x_j^{k - 1},x_j^k} \right]} \right)S_j^k\left( x \right)} ,
	\end{equation*}
	where ${\mathcal{I}}\left( {x \in \left[ {x_j^{k - 1},x_j^k} \right]} \right)$ is an indicator function defined as follows:
	\begin{equation*}
	{\mathcal{I}}\left( {x \in \left[ {x_j^{k - 1},x_j^k} \right]} \right) = \left\{ {\begin{array}{*{20}{c}}
		{1,} & {x \in \left[ {x_j^{k - 1},x_j^k} \right],}  \\
		{0,} & {x \notin \left[ {x_j^{k - 1},x_j^k} \right].}  \\
		\end{array} } \right.
	\end{equation*}
	To ensure the continuity up to the second-order derivative and the monotonicity and normalization of cubic smoothing splined marginal value functions ${u_j}\left(  \cdot  \right)$, $j=1,...,n$, we can consider the following linear constraints:
	\begin{equation*}
	\left. \begin{gathered}
	\left( {{\text{LC1}}} \right)\;\;S_j^k\left( {x_j^k} \right) = S_j^{k + 1}\left( {x_j^k} \right),\;\;k = 1,...,{\gamma _j} - 1,\;\;j = 1,...,n, \hfill \\
	\left( {{\text{LC2}}} \right)\;\;{\left. {\frac{{{\text{d}}S_j^k\left( x \right)}}
			{{{\text{d}}x}}} \right|_{x = x_j^k}} = {\left. {\frac{{{\text{d}}S_j^{k + 1}\left( x \right)}}
			{{{\text{d}}x}}} \right|_{x = x_j^k}},\;\;k = 1,...,{\gamma _j} - 1,\;\;j = 1,...,n, \hfill \\
	\left( {{\text{LC3}}} \right)\;\;{\left. {\frac{{{{\text{d}}^2}S_j^k\left( x \right)}}
			{{{\text{d}}{x^2}}}} \right|_{x = x_j^k}} = {\left. {\frac{{{{\text{d}}^2}S_j^{k + 1}\left( x \right)}}
			{{{\text{d}}{x^2}}}} \right|_{x = x_j^k}},\;\;k = 1,...,{\gamma _j} - 1,\;\;j = 1,...,n, \hfill \\
	\left( {{\text{LC4}}} \right)\;\;S_j^k\left( {x_j^k} \right) \geqslant 0,\;\;k = 0,1,...,{\gamma _j},\;\;j = 1,...,n, \hfill \\
	\left( {{\text{LC5}}} \right)\;\;{\left. {\frac{{{\text{d}}S_j^k\left( x \right)}}
			{{{\text{d}}x}}} \right|_{x = x_j^k}} \geqslant 0,\;\;k = 0,1,...,{\gamma _j},\;\;j = 1,...,n, \hfill \\
	\left( {{\text{LC6}}} \right)\;\;S_j^1\left( {{\alpha _j}} \right) = 0,\;\;j = 1,...,n, \hfill \\
	\left( {{\text{LC7}}} \right)\;\;\sum\limits_{j = 1}^n {S_j^{{\gamma _j}}\left( {{\beta _j}} \right)}  = 1, \hfill 
	\end{gathered}  \right\}E_{{\text{BASE}}}^{{\text{SPLINE}}}
	\end{equation*}
	where constraints (LC1), (LC2), (LC3) guarantee the continuity of ${u_j}\left(  \cdot  \right)$ and their first- and second-order derivatives at breakpoints, respectively. Constraints (LC4) and (LC5) ensure the non-negativity of piecewise-polynomials $S_j^k\left(  \cdot  \right)$ and their first-order derivatives at breakpoints, respectively, which are used to make ${u_j}\left(  \cdot  \right)$ non-negative and monotone non-decreasing at breakpoints. Note that constraints (LC4) and (LC5) are not sufficient conditions for deriving non-negative monotone non-decreasing marginal value functions over the whole performance scales, since they only work for breakpoints. However, in case the non-negativity or monotone non-decreasing properties do not hold, we can divide the performance scales into more refined sub-intervals and incorporate more constraints until deriving desired marginal value functions. This method is easy to implement without more dedicated techniques. Another possible way to generate non-negative polynomials is to use semidefinite programming models (refer to \cite{sobrie2018uta} for more details). Constraints (LC6) and (LC7) normalize marginal value functions, where $S_j^{{\gamma _j}}\left( \beta_j  \right)$ can be understood as the trade-off weight of marginal value function $u_j \left( \cdot \right)$ in the comprehensive value. Analogously to piecewise-linear marginal value function, cubic smoothing splined marginal value function is also a non-parametric model for modeling preferences. Let ${\bm{\theta }} = {\left( {{{\bm{\theta }}_1^\text{T}},...,{{\bm{\theta }}_n^\text{T}}} \right)^{\text{T}}}$, ${{\bm{\theta }}_j} = {\left( {{\bm{\theta }}_j^{1\text{T}},...,{\bm{\theta }}_j^{{\gamma_j} {\text{T}}}} \right)^{\text{T}}}$ for $j=1,...,n$, ${\bm{\theta }}_j^k = {\left( {s_j^{k,0},s_j^{k,1},s_j^{k,2},s_j^{k,3}} \right)^{\text{T}}}$ for $k = 1,...,{\gamma _j}$, and ${\mathbf{V}}\left( a \right) = {\left( {{{\mathbf{V}}_1}\left( a \right)^\text{T},...,{{\mathbf{V}}_n}\left( a \right)^\text{T}} \right)^{\text{T}}}$, ${{\mathbf{V}}_j}\left( a \right) = {\left( {{\mathbf{V}}_j^1\left( a \right)^\text{T},...,{\mathbf{V}}_j^{{\gamma _j}}\left( a \right)^\text{T}} \right)^{\text{T}}}$ for $j = 1,...,n$, ${\mathbf{V}}_j^k\left( a \right) = \left\{ {\begin{array}{*{20}{c}} {{{\left( {1,{g_j}\left( a \right),{{\left( {{g_j}\left( a \right)} \right)}^2},{{\left( {{g_j}\left( a \right)} \right)}^3}} \right)}^{\text{T}}},} & {{\text{if}}\;{g_j}\left( a \right) \in \left[ {x_j^{k - 1},x_j^k} \right],}  \\ {{{\left( {0,0,0,0} \right)}^{\text{T}}},} & {{\text{if}}\;{g_j}\left( a \right) \notin \left[ {x_j^{k - 1},x_j^k} \right],}  \\ \end{array} } \right.$ for $k = 1,...,{\gamma _j}$, and then the comprehensive value of $a$ can be formulated as $U\left( a \right) = {{\bm{\theta }}^{\text{T}}}{\mathbf{V}}\left( a \right)$.
	
	When it comes to general monotone value function, it is a very flexible preference model as it considers all monotone non-decreasing marginal value functions (rather than linear, piecewise-linear, or splined shaped marginal value functions) and does not involve any arbitrary or restrictive parametrization \cite{corrente2013robust, greco2010multiple}. Let ${\chi _j} = \left\{ {x \in \mathbb{R}\;|\;\exists a \in A\;{\text{such}}\;{\text{that}}\;{g_j}\left( a \right) = x} \right\}$ and $x_j^0 = {\alpha _j},x_j^1,...,x_j^{{m_j}} = {\beta _j}$ be ordered performance values of ${\chi _j}$, $x_j^k < x_j^{k + 1}$, $k=0,1,...,m_j-1$, ${m_j} = \left| {{\chi _j}} \right| \leqslant m$. In defining general monotone marginal value functions ${u_j}\left(  \cdot  \right)$, $j=1,...,n$, all marginal values corresponding to characteristic points $x_j^k \in {\chi _j}$, $k=0,1,...,m_j$, $j=1,...,n$, are parameters to be determined by considering the following linear constraints which are used to ensure monotonicity and normalization:
	\begin{equation*}
	\left. \begin{gathered}
	{u_j}\left( {x_j^k} \right) \leqslant {u_j}\left( {x_j^{k + 1}} \right),\;\;k = 0,1,...,{m_j} - 1,\;\;j = 1,...,n, \hfill \\
	{u_j}\left( {{\alpha _j}} \right) = 0,\;\;j = 1,...,n, \hfill \\
	\sum\limits_{j = 1}^n {{u_j}\left( {{\beta _j}} \right)}  = 1, \hfill
	\end{gathered}  \right\}E_{{\text{BASE}}}^{{\text{GENERAL}}}
	\end{equation*}
	where ${u_j}\left( {{\beta _j}} \right)$ represents the trade-off weight of marginal value function $u_j \left( \cdot \right)$ in the comprehensive value. Since we only make the monotonicity assumption on general shape of marginal value functions, it can be deemed as a non-parametric preference model \cite{spliet2014preference}. Let ${\bm{\theta }} = {\left( {{\bm{\theta }}_1^{\text{T}},...,{\bm{\theta }}_n^{\text{T}}} \right)^{\text{T}}}$, ${{\bm{\theta }}_j} = {\left( {{u_j}\left( {x_j^0} \right),...,{u_j}\left( {x_j^{{m_j}}} \right)} \right)^{\text{T}}}$ for $j=1,...,n$, and ${\mathbf{V}}\left( a \right) = {\left( {{{\mathbf{V}}_1}{{\left( a \right)}^{\text{T}}},...,{{\mathbf{V}}_n}{{\left( a \right)}^{\text{T}}}} \right)^{\text{T}}}$, ${{\mathbf{V}}_j}\left( a \right) = {\left( {v_j^0\left( a \right),...,v_j^{{m_j}}\left( a \right)} \right)^{\text{T}}}$ for $j=1,...,n$, $v_j^k\left( a \right) = \left\{ {\begin{array}{*{20}{c}}
		{1,} \hfill & {{\text{if}}\;{g_j}\left( a \right) = x_j^k,} \hfill  \\
		{0,} \hfill & {{\text{if}}\;{g_j}\left( a \right) \ne x_j^k,} \hfill  \\
		\end{array} } \right.$ for $k=0,1,...,m_j$. Then, the comprehensive value of $a$ can be formulated as $U\left( a \right) = {{\bm{\theta }}^{\text{T}}}{\mathbf{V}}\left( a \right)$.
	
	To sum up, for any type of the considered value functions in the above, the comprehensive value of $a$ can be written in a linear form $U\left( a \right) = {{\bm{\theta }}^{\text{T}}}{\mathbf{V}}\left( a \right)$, where $\bm{\theta} $ is the intrinsic character of the employed value function and irrelevant for an alternative, while ${\mathbf{V}}\left( a \right)$ depends on the performances of the corresponding alternative $a$ on multiple criteria. In this perspective, we use a linear model to approximate preferences, although the actual value function model could be non-linear.
	
	\subsection{Constructing additive value function model from valued decision examples}
	\label{sec-22}
	
	\noindent We aim to construct an additive value function model from a given set of decision examples. In the unified analytical framework, the constructed additive value model can be composed of any type of linear, piecewise-linear, splined, or general marginal value functions introduced in Section \ref{sec-21}. In contrast to traditional sorting problems, where each reference alternative is assigned precisely to only one decision class, the sorting problem considered in this study involves a set of valued decision examples, each of which assigns a reference alternative to more than one class with respective credibility degrees. Let us use the following notation to describe the considered sorting problem: $CL = \{{Cl_{1}, Cl_{2}, ..., Cl_{q}}\}$ is a set of predefined and preference-ordered decision classes, such that $Cl_{s+1}$ is preferred to $Cl_{s}$ (denoted by $Cl_{s+1} \succ Cl_{s}$), $s=1,...,q-1$. Suppose that the set of alternatives $A$ can be divided into two subsets -- the reference one $A^R$ and the non-reference one $A^T$. For any reference alternative $a \in A^R$, it could be assigned to multiple classes, and we use a vector ${\bm{\sigma }}\left( a \right) = {\left( {{{\mathbf{\sigma }}_1}\left( a \right),...,{{\mathbf{\sigma }}_q}\left( a \right)} \right)^{\text{T}}}$ to represent the credibility degrees for each possible assignment, i.e., $a$ is assigned to class $Cl_s$ with a credibility degree ${\sigma _s}\left( a \right)$, $s=1,...,q$. Note that for normalization we require $\sum\limits_{s = 1}^q {{\sigma _s}\left( a \right)}  = 1$ and ${\sigma _s}\left( a \right) \geqslant 0$ for $s=1,...,q$. Moreover, a crisp decision example, which is considered in traditional sorting problems, is a particular case of a~valued decision example, where there exists $s \in \left\{ {1,...,q} \right\}$ such that ${\sigma _s}\left( a \right) = 1$, and ${\sigma _{s'}}\left( a \right) = 0$ for $s' \in \left\{ {1,...,q} \right\}$, $s' \ne s$. The~assignment for each non-reference alternative $a \in A^T$ needs to be determined using the constructed preference model.
	
	\subsubsection{Dealing with valued assignment examples}
	\noindent For a general sorting problem, we often refer to a sorting rule called example-based sorting procedure given by \cite{greco2010multiple}, which is described as follows.
	\begin{description}
		\item[Definition 1.] For any pair of alternatives $a_i$ and $a_j$, a value function $U\left(  \cdot  \right)$ is said to be consistent with the assignments of $a_i$ and $a_j$ iff:
		\begin{equation}\label{eq-1}
		U\left( {a_i} \right) \geqslant U\left( {a_j} \right) \Rightarrow Cl\left( {a_i} \right) \succsim Cl\left( {a_j} \right),
		\end{equation}
		where $Cl\left( {{a_i}} \right),Cl\left( {{a_j}} \right) \in CL$ are the assignments of $a_i$ and $a_j$, respectively, and $\succsim$ means ``at least as good as''.  Observe that implication (\ref{eq-1}) is equivalent to:
		\begin{equation}\label{eq-2}
		Cl\left( {a_i} \right) \succ Cl\left( {a_j} \right) \Rightarrow U\left( {a_i} \right) > U\left( {a_j} \right).
		\end{equation}
	\end{description}
	Definition 1 says that ``if alternative $a_i$ has a value which is not less than for alternative $a_j$, the assignment of $a_i$ should be at least as good as the assignment of $a_j$'', or equivalently, ``if the assignment of $a_i$ is better than the assignment of $a_j$, the value of $a_i$ should be greater than the value of $a_j$''. Thus, for any pair of reference alternatives $a_i, a_j \in A^R$ such that $a_i$ is assigned to a class better than $a_j$, we can infer a value function by maximizing the difference between $U\left( {{a_i}} \right)$ and $U\left( {{a_j}} \right)$ (i.e., $U\left( {{a_i}} \right) - U\left( {{a_j}} \right)$). The aim of doing so is two-fold: on the one hand, when there exists at least one value function $U\left(  \cdot  \right)$ compatible with the assignments of $a_i$ and $a_j$ (i.e., $U\left( {{a_i}} \right) - U\left( {{a_j}} \right) > 0$), maximizing $U\left( {{a_i}} \right) - U\left( {{a_j}} \right)$ highlights the difference between $U\left( {{a_i}} \right)$ and $U\left( {{a_j}} \right)$; on the other hand, when no such compatible value function exists (i.e., $U\left( {{a_i}} \right) - U\left( {{a_j}} \right) \leqslant 0$ for all $U\left(  \cdot  \right)$), maximizing $U\left( {{a_i}} \right) - U\left( {{a_j}} \right)$ amounts to minimizing the inconsistency level between $U\left( {{a_i}} \right)$ and $U\left( {{a_j}} \right)$.
	
	In addition to the above requirements for pairs of reference alternatives that come from distinct classes, another target to be accounted for when inferring a value function model consists in that, the values of reference alternatives from the same class should be as concentrated as possible, so that consecutive classes could be clearly delimited. This goal can be implemented by minimizing the absolute difference between $U\left( {{a_i}} \right)$ and $U\left( {{a_j}} \right)$ (i.e., $\left| {U\left( {{a_i}} \right) - U\left( {{a_j}} \right)} \right|$) for pairs of reference alternatives ${a_i},{a_j}$ in the same class. This target is in line with the idea of maximizing the distances of the correctly classified alternatives from the class thresholds in a~threshold-based sorting procedure (e.g., the UTADIS II method \cite{doumpos2002multicriteria}).
	
	In the context of valued decision examples, as a reference alternative could be assigned to multiple classes with different credibility degrees, we can account for the above two targets for any pair of reference alternatives by investigating all their desired assignments and attaching each of them with a certain credibility. 
	\begin{description}
		\item[Definition 2.] For any pair of reference alternatives ${a_i},{a_j} \in {A^R}$, the credibility degrees for the facts that ``$a_i$ is assigned to a better class than $a_j$'', ``both $a_i$ and $a_j$ are assigned to same class'', and ``$a_i$ is assigned to a~worse class than $a_j$'' are defined as follows, respectively:
		\begin{align*}
		& {D_ \succ }\left( {{a_i},{a_j}} \right) = \sum\limits_{s = 2}^q {\sum\limits_{r = 1}^{s - 1} {{\sigma _s}\left( {{a_i}} \right){\sigma _r}\left( {{a_j}} \right)} } , \\
		& {D_ = }\left( {{a_i},{a_j}} \right) = \sum\limits_{s = 1}^q {{\sigma _s}\left( {{a_i}} \right){\sigma _s}\left( {{a_j}} \right)} , \\
		& {D_ \prec }\left( {{a_i},{a_j}} \right) = \sum\limits_{s = 1}^{q - 1} {\sum\limits_{r = s + 1}^q {{\sigma _s}\left( {{a_i}} \right){\sigma _r}\left( {{a_j}} \right)} } .
		\end{align*}
	\end{description}
	Coefficients ${D_ \succ }\left( {{a_i},{a_j}} \right)$, ${D_ = }\left( {{a_i},{a_j}} \right)$, and ${D_ \prec }\left( {{a_i},{a_j}} \right)$ aggregate the credibility degrees ${{\sigma _s}\left( {{a_i}} \right)}$ and ${{\sigma _r}\left( {{a_j}} \right)}$ for all possible cases $C{l_s} \succ C{l_r}$, $C{l_s} = C{l_r}$, and $C{l_s} \prec C{l_r}$, respectively. Note that, for ${D_ \succ }\left( {{a_i},{a_j}} \right)$, ${D_ = }\left( {{a_i},{a_j}} \right)$, and ${D_ \prec }\left( {{a_i},{a_j}} \right)$, the multiplicative form ${\sigma _s}\left( {{a_i}} \right){\sigma _r}\left( {{a_j}} \right)$ is applied, because the two events ``alternative $a_i$ is assigned to class $Cl_s$ with a credibility degree ${\sigma _s}\left( {{a_i}} \right)$'' and ``alternative $a_j$ is assigned to class $Cl_r$ with a credibility degree ${\sigma _r}\left( {{a_j}} \right)$'' are independent. With the credibility degrees for the comparison between the possible assignments of $a_i$ and $a_j$, we can consider to minimize the following objective for inferring a value function model:
	\begin{equation*}
	\xi \left( {{a_i},{a_j}} \right) =  - {D_ \succ }\left( {{a_i},{a_j}} \right)\left( {U\left( {{a_i}} \right) - U\left( {{a_j}} \right)} \right) + {D_ = }\left( {{a_i},{a_j}} \right)\left| {U\left( {{a_i}} \right) - U\left( {{a_j}} \right)} \right| + {D_ \prec }\left( {{a_i},{a_j}} \right)\left( {U\left( {{a_i}} \right) - U\left( {{a_j}} \right)} \right).
	\end{equation*}
	Specifically, minimizing $\xi \left( {{a_i},{a_j}} \right)$ aims to maximize ${U\left( {{a_i}} \right) - U\left( {{a_j}} \right)}$ with the credibility degree ${D_ \succ }\left( {{a_i},{a_j}} \right)$ and minimize $\left| {U\left( {{a_i}} \right) - U\left( {{a_j}} \right)} \right|$ with the credibility degree ${D_ = }\left( {{a_i},{a_j}} \right)$ as well as ${U\left( {{a_i}} \right) - U\left( {{a_j}} \right)}$ with the credibility degree ${D_ \prec }\left( {{a_i},{a_j}} \right)$. Note that we minimize $\left| {U\left( {{a_i}} \right) - U\left( {{a_j}} \right)} \right|$ rather than $\left({U\left( {{a_i}} \right) - U\left( {{a_j}} \right)}\right)^2$, for the case that both $a_i$ and $a_j$ are assigned to the same class, as in this way we keep all considered targets in the same magnitude. For any pair ${a_i},{a_j} \in {A^R}$, it is easy to verify that $\xi \left( {{a_i},{a_j}} \right) = \xi \left( {{a_j},{a_i}} \right)$. Moreover, $\xi \left( {{a_i},{a_j}} \right)$ has the following two properties.
	\begin{description}
		\item[Property 1.] For any pair of reference alternatives ${a_i},{a_j} \in {A^R}$, if there exists $s,s' \in \{ 1,...,q \}$ such that $s \geqslant s'$, and ${\sigma _r}\left( {{a_i}} \right) > 0$ for $r = s,...,q$, and ${\sigma _r}\left( {{a_i}} \right) = 0$ for $r = 1,...,s - 1$, and ${\sigma _r}\left( {{a_j}} \right) = 0$ for $r = s',...,q$, and ${\sigma _r}\left( {{a_j}} \right) > 0$ for $r = 1,...,s' - 1$, then minimizing $\xi \left( {{a_i},{a_j}} \right)$ amounts to maximizing ${U\left( {{a_i}} \right) - U\left( {{a_j}} \right)}$ definitely as ${D_ \succ }\left( {{a_i},{a_j}} \right) = 1$, ${D_ = }\left( {{a_i},{a_j}} \right) = 0$ and ${D_ \prec }\left( {{a_i},{a_j}} \right) = 0$.
		\item[Property 2.] For any pair of reference alternatives ${a_i},{a_j} \in {A^R}$, if ${\bm{\sigma }}\left( {{a_i}} \right) = {\bm{\sigma }}\left( {{a_j}} \right)$, minimizing $\xi \left( {{a_i},{a_j}} \right)$ includes minimizing $\left| {U\left( {{a_i}} \right) - U\left( {{a_j}} \right)} \right|$ with a certain credibility degree ${D_ = }\left( {{a_i},{a_j}} \right) = \sum\limits_{s = 1}^q {{\sigma _s}{{\left( {{a_i}} \right)}^2}} $. In particular, if there exists $s,s' \in \{ 1,...,q \}$ such that $s \leqslant s'$, and ${\sigma _r}\left( {{a_i}} \right) = {1 \mathord{\left/ {\vphantom {1 {\left( {s' - s} \right)}}} \right. \kern-\nulldelimiterspace} {\left( {s' - s + 1} \right)}}$ for $r = s,...,s'$, and ${\sigma _r}\left( {{a_i}} \right) = 0$ for $r = 1,...,s - 1,s'+1,...,q$, then the credibility degree ${D_ = }\left( {{a_i},{a_j}} \right) $ for minimizing $\left| {U\left( {{a_i}} \right) - U\left( {{a_j}} \right)} \right|$ becomes ${1 \mathord{\left/ {\vphantom {1 {\left( {s' - s} \right)}}} \right. \kern-\nulldelimiterspace} {\left( {s' - s + 1} \right)}}$ and such a value increases as $s'-s$ decreases. In an extreme case where $s = s'$ (i.e., both $a_i$ and $a_j$ are assigned to a unique class), minimizing $\xi \left( {{a_i},{a_j}} \right)$ amounts to minimizing $\left| {U\left( {{a_i}} \right) - U\left( {{a_j}} \right)} \right|$ with the credibility degree ${D_ = }\left( {{a_i},{a_j}} \right) = 1$.
	\end{description}
	Property 1 states that, if the assignment of alternative $a_i$ is unanimously better than the assignment of alternative $a_j$ (without overlap of non-zero credibility degrees), minimizing $\xi \left( {{a_i},{a_j}} \right)$ is equal to maximizing ${U\left( {{a_i}} \right) - U\left( {{a_j}} \right)}$ completely credibly. Property 2 reveals that, for a pair of alternatives $a_i$ and $a_j$ that have the same distribution of credibility degrees for each class, the more concentrated the distribution is, the more credible it is to minimize $\left| {U\left( {{a_i}} \right) - U\left( {{a_j}} \right)} \right|$. Particularly, if both $a_i$ and $a_j$ are assigned to a unique class definitely, minimizing $\xi \left( {{a_i},{a_j}} \right)$ is equal to minimizing $\left| {U\left( {{a_i}} \right) - U\left( {{a_j}} \right)} \right|$ completely credibly. Therefore, minimizing $\xi \left( {{a_i},{a_j}} \right)$ accounts for not only differentiating the values of $a_i$ and $a_j$ for the case that $a_i$ is always assigned to a better class than $a_j$, but also equalizing the values of $a_i$ and $a_j$ for the case that both $a_i$ and $a_j$ are assigned to a unique class definitely. Such an observation derived from Property 1 and 2 confirms the appropriateness of minimizing $\xi \left( {{a_i},{a_j}} \right)$ for inferring a value function model.
	
	\subsubsection{Regularization}
	\noindent Minimizing $\xi \left( {{a_i},{a_j}} \right)$ for all pairs of reference alternatives ${a_i},{a_j} \in {A^R}$ can be seen as constructing a preference model that can fit the valued decision examples as confidently as possible. Besides the consideration of the model's fitting ability, we also need to account for the complexity of the preference model. As suggested by the statistical learning theory \cite{doumpos2007regularized, liu2019preference}, a proper complexity control contributes to avoiding the over-fitting problem in which the constructed preference model fits the decision examples well but has poor generalization performance on non-reference alternatives. Thus, in this study, we also incorporate the regularization techniques into the preference learning procedure to address the trade-off between the preference model's fitting performance and complexity control so that the constructed preference model would have good generalization performance and be robust to the noise in the decision examples. The basic idea is to construct a value function model that is as ``simple'' as possible while maintaining its fitting performance on decision examples. As we consider four types of marginal value functions in the framework, we will discuss how to define the complexity measure for them separately.
	
	For the case of linear marginal value function, defining the complexity measure is relatively ``simpler'' than for the piecewise-linear, splined, and general monotone marginal value functions, since each linear marginal value function has only one parameter $w_j$ to estimate. In the MCDA context, since all criteria contained in a consistent family are relevant to the decision problem, we do not hope any criterion has an overwhelming weight than others \cite{liu2019preference}. Therefore, the complexity control of an additive value function can be implemented by minimizing the sum of squares of $w_j$, i.e.: 
	\begin{equation*}
	{\Omega ^{{\text{LINEAR}}}}\left( U \right) = C \sum\limits_{j = 1}^n {w_j^2},
	\end{equation*}
	where $C > 0$ is a constant to establish the trade-off between the complexity control and the fitting ability. Minimizing ${\Omega ^{{\text{LINEAR}}}}\left( U \right)$ can be seen as penalizing the square of the $L_2$ norm of the weight vector. According to the statistical learning theory \cite{hastie2009elements, murphy2012machine}, the $L_2$ norm can regularize the weights to be smooth across criteria, thus avoiding some criterion having overwhelming weights.
	
	When it comes to piecewise-linear marginal value function, the complexity measure has been defined by \cite{liu2019preference} as the smoothness of this function, which can be quantified as the variations of slope at breakpoints as follows:
	\begin{equation*}
	{\Omega ^{{\text{PIECEWISE - LINEAR}}}}\left( U \right) = {C_1}\sum\limits_{j = 1}^n {{{\left( {\sum\limits_{t = 1}^{{\gamma _j}} {\Delta u_j^t} } \right)}^2}}  + {C_2}{\sum\limits_{j = 1}^n {\sum\limits_{t = 1}^{{\gamma _j} - 1} {\left( {\frac{{{\gamma _j}\left( {\Delta u_j^{t + 1} - \Delta u_j^t} \right)}}
					{{{\beta _j} - {\alpha _j}}}} \right)} } ^2},
	\end{equation*}
	where $C_1,C_2 > 0$ are two constants to make trade-off between the complexity control and the fitting ability, and ${\sum\limits_{t = 1}^{{\gamma _j}} {\Delta u_j^t} }$ is the trade-off weight of marginal value function $u_j \left( \cdot \right)$ in the comprehensive value, and ${\frac{{{\gamma _j}\left( {\Delta u_j^{t + 1} - \Delta u_j^t} \right)}}{{{\beta _j} - {\alpha _j}}}}$ measures the variation of slope of marginal value function $u_j \left( \cdot \right)$ at breakpoint $x_j^t$. In this way, we not only avoid generating some marginal value functions that have overwhelming weights, but also pursue marginal value functions that are as linear as possible.
	
	When using splined marginal value function, the basic idea to control its complexity is to add a smoothing term that penalizes functions that are ``too wiggly''. This can be performed by penalizing the curvature in the function \cite{hastie2009elements}, i.e.:
	\begin{align*}
	\begin{gathered}
	{\Omega ^{{\text{SPLINE}}}}\left( U \right) = {C_1}\sum\limits_{j = 1}^n {{{\left( {S_j^{{\gamma _j}}\left( {{\beta _j}} \right)} \right)}^2}}  + {C_2}\sum\limits_{j = 1}^n {\int_{{\alpha _j}}^{{\beta _j}} {{{\left( {\frac{{{{\text{d}}^2}{u_j}\left( x \right)}}
						{{{\text{d}}{x^2}}}} \right)}^2}} {\text{d}}x}  \hfill \\
	\;\;\;\;\;\; = {C_1}\sum\limits_{j = 1}^n {{{\left( {S_j^{{\gamma _j}}\left( {{\beta _j}} \right)} \right)}^2}}  + {C_2}\sum\limits_{j = 1}^n {\sum\limits_{k = 1}^{{\gamma _j}} {\int_{x_j^{k - 1}}^{x_j^k} {{{\left( {\frac{{{{\text{d}}^2}S_j^k}}
							{{{\text{d}}{x^2}}}} \right)}^2}} {\text{d}}x} },  \hfill \\ 
	\end{gathered}
	\end{align*}
	where $C_1,C_2 > 0$ are two constants to take into account the trade-off between the complexity control and the fitting ability, and $S_j^{{\gamma _j}}\left( \beta_j  \right)$ is the trade-off weight of marginal value function $u_j \left( \cdot \right)$ in the comprehensive value, and ${\sum\limits_{j = 1}^n {\sum\limits_{k = 1}^{{\gamma _j}} {\int_{x_j^{k - 1}}^{x_j^k} {\left( {\frac{{{{\text{d}}^2}S_j^k}}{{{\text{d}}{x^2}}}} \right)} } } ^2}{\text{d}}x$ is used to avoid generating marginal value functions that are too wiggly over the performance scales.

	As for general monotone marginal value function, we prefer functions that increase stably over the performance scales and thus a proper measure of its complexity can be the variation of growth rates of marginal values over consecutive sub-intervals as follows:
	\begin{equation*}
	{\Omega ^{{\text{GENERAL}}}}\left( U \right) = {C_1}\sum\limits_{j = 1}^n {{{\left( {{u_j}\left( {{\beta _j}} \right)} \right)}^2}}  + {C_2}\sum\limits_{j = 1}^n {\sum\limits_{k = 1}^{{m_j} - 1} {{{\left( {\frac{{{u_j}\left( {x_j^{k + 1}} \right) - {u_j}\left( {x_j^k} \right)}}
						{{x_j^{k + 1} - x_j^k}} - \frac{{{u_j}\left( {x_j^k} \right) - {u_j}\left( {x_j^{k - 1}} \right)}}
						{{x_j^k - x_j^{k - 1}}}} \right)}^2}} },
	\end{equation*}
	where $C_1,C_2 > 0$ are two constants to make a~trade-off between the complexity control and the fitting ability, and ${u_j}\left( {{\beta _j}} \right)$ is the trade-off weight of marginal value function $u_j \left( \cdot \right)$ in the comprehensive value, and $\frac{{{u_j}\left( {x_j^{k + 1}} \right) - {u_j}\left( {x_j^k} \right)}}{{x_j^{k + 1} - x_j^k}} - \frac{{{u_j}\left( {x_j^k} \right) - {u_j}\left( {x_j^{k - 1}} \right)}}{{x_j^k - x_j^{k - 1}}}$ measures the difference of growth rates of marginal values $u_j \left( \cdot \right)$ over the consecutive sub-intervals $\left[ {x_j^{k - 1},x_j^k} \right]$ and $\left[ {x_j^k,x_j^{k + 1}} \right]$. By considering the above complexity measure, we avoid deriving general monotone marginal value functions that have very disparate growth rates over consecutive sub-intervals.
	
	In a joint consideration of the fitting ability and the complexity control, we propose the following optimization model for constructing a value function model from the valued decision examples:
	\begin{equation*}
	\begin{gathered}
	\left( {{\text{P1}}} \right):\;Minimize\;F\left({\bm{\theta}}\right) = \sum\limits_{{a_i},{a_j} \in {A^R}:\;i < j} {\xi \left( {{a_i},{a_j}} \right)}  + {\Omega ^M}\left( U \right) \hfill \\
	= \sum\limits_{{a_i},{a_j} \in {A^R}:\;i < j} {\left( { - {D_ \succ }\left( {{a_i},{a_j}} \right)\left( {U\left( {{a_i}} \right) - U\left( {{a_j}} \right)} \right) + {D_ = }\left( {{a_i},{a_j}} \right)\left| {U\left( {{a_i}} \right) - U\left( {{a_j}} \right)} \right| + {D_ \prec }\left( {{a_i},{a_j}} \right)\left( {U\left( {{a_i}} \right) - U\left( {{a_j}} \right)} \right)} \right)}  + {\Omega ^M}\left( U \right) \hfill \\
	= \sum\limits_{{a_i},{a_j} \in {A^R}:\;i < j} {\left( {\left( {{D_ \prec }\left( {{a_i},{a_j}} \right) - {D_ \succ }\left( {{a_i},{a_j}} \right)} \right){{\bm{\theta }}^{\text{T}}}\left( {{\mathbf{V}}\left( a_i \right) - {\mathbf{V}}\left( a_j \right)} \right) + {D_ = }\left( {{a_i},{a_j}} \right)\left| {{{\bm{\theta }}^{\text{T}}}\left( {{\mathbf{V}}\left( a_i \right) - {\mathbf{V}}\left( a_j \right)} \right)} \right|} \right)}  + {\Omega ^M}\left( U \right), \hfill \\
	\;\;\;\;\;\;\;\;{\text{s}}{\text{.t}}{\text{.}}\;E_{{\text{BASE}}}^M. \hfill \\ 
	\end{gathered}
	\end{equation*}
	where $M \in \left\{ {{\text{LINEAR,}}\;{\text{PIECEWISE - LINEAR,}}\;{\text{SPLINE,}}\;{\text{GENERAL}}} \right\}$ so that the above model applies to different types of value functions. Note that the hyper-parameters $C$, $C_1$ and $C_2$ in the complexity control ${\Omega ^M}\left( U \right)$ are used to make a~trade-off between the the fitting ability and the complexity control. Therefore, choosing the hyper-parameters $C$, $C_1$ and $C_2$ signifies to select between models with different degrees of complexity, which is known as \emph{model selection} \cite{murphy2012machine}. A widely used method for solving this problem is \emph{$k$-fold cross-validation}, where $k$~is specified by a user, usually 5 or 10. Cross-validation for selecting $C$, $C_1$ and $C_2$ can be performed as follows: reference set $A^R$ is first randomly partitioned into $k$ subsets of (approximately) equal size, called folds. Next, for certain $C$, $C_1$ and $C_2$, $k-1$ folds serve as the training samples to construct an additive value function model and the remaining fold is used to test the constructed model. This process is repeated using different combinations of $k-1$ folds and thus generates $k$ possible results. Finally, the $k$ results are averaged to evaluate the performance of the constructed models corresponding to certain $C$, $C_1$ and $C_2$. We choose the values of $C$, $C_1$ and $C_2$ corresponding to the best performance as the optimal setting for these hyper-parameters.
	
	\subsubsection{Optimization model}
	\noindent It is easy to verify that both ${\left| {{{\bm{\theta }}^{\text{T}}}\left( {{\mathbf{V}}\left( a \right) - {\mathbf{V}}\left( b \right)} \right)} \right|}$ and ${\Omega ^*}\left( U \right)$ are convex in terms of $\bm{\theta}$ and particularly, ${\Omega ^*}\left( U \right)$ is in a quadratic form. Therefore, Model P1 is a constrained convex quadratic optimization problem. To solve such a problem, a common method is to introduce an auxiliary variable $\tau \left( {{a_i},{a_j}} \right)$ for any pair of reference alternatives ${a_i},{a_j} \in {A^R}$ and transform P1 to the following form:
	\begin{equation*}
	\footnotesize
	\begin{gathered}
	\left( {{\text{P1}}} \right)':\;Minimize\;F\left({\bm{\theta}}\right) = \sum\limits_{{a_i},{a_j} \in {A^R}:\;i < j} {\left( {\left( {{D_ \prec }\left( {{a_i},{a_j}} \right) - {D_ \succ }\left( {{a_i},{a_j}} \right)} \right){{\bm{\theta }}^{\text{T}}}\left( {{\mathbf{V}}\left( a_i \right) - {\mathbf{V}}\left( a_j \right)} \right) + {D_ = }\left( {{a_i},{a_j}} \right)\tau \left( {{a_i},{a_j}} \right)} \right)}  + {\Omega ^M}\left( U \right), \hfill \\
	\;\;\;\;\;\;\;\;\;\;\;\;{\text{s}}{\text{.t}}{\text{.}}\;{{\bm{\theta }}^{\text{T}}}\left( {{\mathbf{V}}\left( a_i \right) - {\mathbf{V}}\left( a_j \right)} \right) \leqslant \tau \left( {{a_i},{a_j}} \right),\;\;{\text{for}}\;{a_i},{a_j} \in {A^R},\;\;i < j, \hfill \\
	\;\;\;\;\;\;\;\;\;\;\;\;\;\;\;\;\;- {{\bm{\theta }}^{\text{T}}}\left( {{\mathbf{V}}\left( a_i \right) - {\mathbf{V}}\left( a_j \right)} \right) \leqslant \tau \left( {{a_i},{a_j}} \right),\;\;{\text{for}}\;{a_i},{a_j} \in {A^R},\;\;i < j, \hfill \\
	\;\;\;\;\;\;\;\;\;\;\;\;\;\;\;\;\;E_{{\text{BASE}}}^M. \hfill \\ 
	\end{gathered}
	\end{equation*}
	Then, some popular optimization packages, such as Lingo, Cplex, or Matlab can be used to address the above problem. However, when the number of reference alternatives is reasonably large, the number of pairs ${a_i},{a_j} \in {A^R}$ such that $i < j$ is in a huge amount and this may exceed the processing ability of most solvers.
	
	In this paper, to enhance the practical ability for addressing large-scale problems, we introduce computational advances in the convex optimization field and use the \emph{alternating direction method of multipliers} (ADMM) to address Model P1. This method is well suited to large-scale problems and has a potential for distributed implementation. It has been widely used in statistics, machine learning, and related areas \cite{boyd2011distributed}. Specifically, to~apply ADMM for solving Model P1, we need to transform P1 to the following form:
	\begin{equation*}
	\begin{gathered}
	({\text{P1}})'':\;Minimize\;f\left( {\bm{\theta }} \right) + h\left( {\mathbf{z}} \right), \hfill \\
	\;\;\;\;\;\;\;\;\;\;\;\;{\text{s}}{\text{.t}}{\text{.}}\;{{\mathbf{Y}}^{\text{T}}}{\bm{\theta }} = {\mathbf{z}}, \hfill \\ 
	\end{gathered}
	\end{equation*}
	where the vector ${\mathbf{z}}$ is an auxiliary variable, and the functions $f\left( {\bm{\theta }} \right)$ and $h\left( {\mathbf{z}} \right)$ are formulated as:
	\begin{equation*}
	f\left( {\bm{\theta }} \right) = \left\{ {\begin{array}{*{20}{c}}
		{\sum\limits_{{a_i},{a_j} \in {A^R}:\;i < j} {\left( {\left( {{D_ \prec }\left( {{a_i},{a_j}} \right) - {D_ \succ }\left( {{a_i},{a_j}} \right)} \right){{\left( {{\mathbf{V}}\left( {{a_i}} \right) - {\mathbf{V}}\left( {{a_j}} \right)} \right)}^{\text{T}}}} \right)} {\bm{\theta }} + {\Omega ^M}\left( U \right),} \hfill & {{\text{if}}\;{\bm{\theta }}\;{\text{satisfies}}\;E_{{\text{BASE}}}^M,} \hfill  \\
		{ + \infty } \hfill & {{\text{otherwise}}.} \hfill  \\
		\end{array} } \right.
	\end{equation*}
	and
	\begin{equation*}
	h\left( {\mathbf{z}} \right) = {\left\| {\mathbf{z}} \right\|_1},
	\end{equation*}
	respectively, and ${\mathbf{Y}}$ is a matrix defined as ${\mathbf{Y}} = \left[ {...,{{\mathbf{y}}_{ij}},...} \right]$, and each column ${{\mathbf{y}}_{ij}}$ is given as ${{\mathbf{y}}_{ij}} = {D_ = }\left( {{a_i},{a_j}} \right)\left( {{\mathbf{V}}\left( {{a_i}} \right) - {\mathbf{V}}\left( {{a_j}} \right)} \right)$ for any pair of reference alternatives ${a_i},{a_j} \in {A^R}$ such that $i < j$. Note that the objective $f\left( {\bm{\theta }} \right) + h\left( {\mathbf{z}} \right)$ is equivalent to the objective $F$ in Model P1 by incorporating the constraints ${E_{{\text{BASE}}}^M}$. Moreover, we connect $\bm{\theta}$ to the auxiliary variable $\mathbf{z}$ through the equation ${{\mathbf{Y}}^{\text{T}}}{\bm{\theta }} = {\mathbf{z}}$. In this way, the terms ${\left| {{{\bm{\theta }}^{\text{T}}}\left( {{\mathbf{V}}\left( {{a_i}} \right) - {\mathbf{V}}\left( {{a_j}} \right)} \right)} \right|}$ for any pair of reference alternatives ${a_i},{a_j} \in {A^R}$ such that $i < j$ is converted to the $L_1$ norm of $\mathbf{z}$ (i.e., ${\left\| {\mathbf{z}} \right\|_1}$).
	
	\noindent To solve model $({\text{P1}})''$, ADMM consists of the following iterations \cite{boyd2011distributed}:
	\begin{align*}
	& \text{step 1:}\;\;\;\;\;\;{{\bm{\theta }}^{k + 1}}: = \mathop {\arg \min }\limits_{\bm{\theta }} \left( {f\left( {\bm{\theta }} \right) + \left( {{\rho  \mathord{\left/
					{\vphantom {\rho  2}} \right.
					\kern-\nulldelimiterspace} 2}} \right)\left\| {{{\mathbf{Y}}^{\text{T}}}{\bm{\theta }} - {{\mathbf{z}}^k} + {{\mathbf{u}}^k}} \right\|_2^2} \right), \\
	& \text{step 2:}\;\;\;\;\;\;{{\mathbf{z}}^{k + 1}}: = \mathop {\arg \min }\limits_{\mathbf{z}} \left( {h\left( {\mathbf{z}} \right) + \left( {{\rho  \mathord{\left/
					{\vphantom {\rho  2}} \right.
					\kern-\nulldelimiterspace} 2}} \right)\left\| {{{\mathbf{Y}}^{\text{T}}}{{\bm{\theta }}^{k + 1}} - {\mathbf{z}} + {{\mathbf{u}}^k}} \right\|_2^2} \right)\\
	& \;\;\;\;\;\;\;\;\;\;\;\;\;\;\;\;\;\;\;\;\;\;\;\;\; = \mathop {\arg \min }\limits_{\mathbf{z}} \left( {{{\left\| {\mathbf{z}} \right\|}_1} + \left( {{\rho  \mathord{\left/
					{\vphantom {\rho  2}} \right.
					\kern-\nulldelimiterspace} 2}} \right)\left\| {{{\mathbf{Y}}^{\text{T}}}{{\bm{\theta }}^{k + 1}} - {\mathbf{z}} + {{\mathbf{u}}^k}} \right\|_2^2} \right), \\
	& \text{step 3:}\;\;\;\;\;\;{{\mathbf{u}}^{k + 1}}: = {{\mathbf{u}}^k} + {{\mathbf{Y}}^{\text{T}}}{{\bm{\theta }}^{k + 1}} - {{\mathbf{z}}^{k + 1}},
	\end{align*}
	where $\rho > 0$ is a constant, and the superscript $k$ represents iteration $k$, and $\mathbf{u}$ is an auxiliary variable. The~advantage of using ADMM for addressing large-scale problems derives from the following pair of observations. First, in step 1, we address a convex quadratic problem, in which $\bm{\theta}$ is the only variable, and thus the complexity of solving such a problem only relies on the dimension of $\bm{\theta}$, irrelevant from the number of reference alternatives. Particularly, the coefficient ${\sum\limits_{{a_i},{a_j} \in {A^R}:\;i < j} {\left( {\left( {{D_ \prec }\left( {{a_i},{a_j}} \right) - {D_ \succ }\left( {{a_i},{a_j}} \right)} \right){{\left( {{\mathbf{V}}\left( {{a_i}} \right) - {\mathbf{V}}\left( {{a_j}} \right)} \right)}^{\text{T}}}} \right)} }$ involved in this problem can be calculated and stored in advance since it keeps the same during the whole process. Second, when addressing the optimization problem in step 2, although the term ${{{\left\| {\mathbf{z}} \right\|}_1}}$ is not differentiable, it has been proved that a~simple closed-form solution to this problem exists as follows \cite{donoho1995noising,boyd2011distributed}:
	\begin{equation*}
	\begin{gathered}
	{\left[ {{{\mathbf{z}}^{k + 1}}} \right]_i}: = {S_{1/\rho }}\left( {{{\left[ {{{\mathbf{Y}}^{\text{T}}}{{\bm{\theta }}^{k + 1}} + {{\mathbf{u}}^k}} \right]}_i}} \right) \hfill \\
	= \left\{ {\begin{array}{*{20}{c}}
		{{{\left[ {{{\mathbf{Y}}^{\text{T}}}{{\bm{\theta }}^{k + 1}} + {{\mathbf{u}}^k}} \right]}_i} - 1/\rho ,} \hfill & {{\text{if}}\;{{\left[ {{{\mathbf{Y}}^{\text{T}}}{{\bm{\theta }}^{k + 1}} + {{\mathbf{u}}^k}} \right]}_i} > 1/\rho ,} \hfill  \\
		{0,} \hfill & {{\text{if}}\;\left| {{{\left[ {{{\mathbf{Y}}^{\text{T}}}{{\bm{\theta }}^{k + 1}} + {{\mathbf{u}}^k}} \right]}_i}} \right| \leqslant 1/\rho ,} \hfill  \\
		{{{\left[ {{{\mathbf{Y}}^{\text{T}}}{{\bm{\theta }}^{k + 1}} + {{\mathbf{u}}^k}} \right]}_i} + 1/\rho ,} \hfill & {{\text{if}}\;{{\left[ {{{\mathbf{Y}}^{\text{T}}}{{\bm{\theta }}^{k + 1}} + {{\mathbf{u}}^k}} \right]}_i} <  - 1/\rho .} \hfill  \\
		\end{array} } \right. \hfill \\ 
	\end{gathered}
	\end{equation*}
	where ${\left[ {{{\mathbf{z}}^{k + 1}}} \right]_i}$ and ${{{\left[ {{{\mathbf{Y}}^{\text{T}}}{{\bm{\theta }}^{k + 1}} + {{\mathbf{u}}^k}} \right]}_i}}$ stands for the $i$-th entries of the vectors ${{{\mathbf{z}}^{k + 1}}}$ and ${{{\mathbf{Y}}^{\text{T}}}{{\bm{\theta }}^{k + 1}} + {{\mathbf{u}}^k}}$, respectively, and ${S_{1/\rho }}\left(  \cdot  \right)$ is called the \emph{soft thresholding operator} \cite{donoho1995noising}. One can observe that the updating for $\mathbf{z}$ proceeds sequentially for its each dimension. Even if the number of reference alternatives is very large, this updating can be finished in a short time. In particular, calculating coefficient ${\sum\limits_{{a_i},{a_j} \in {A^R}:\;i < j} {\left( {\left( {{D_ \prec }\left( {{a_i},{a_j}} \right) - {D_ \succ }\left( {{a_i},{a_j}} \right)} \right){{\left( {{\mathbf{V}}\left( {{a_i}} \right) - {\mathbf{V}}\left( {{a_j}} \right)} \right)}^{\text{T}}}} \right)} }$ in step 1 and updating $\mathbf{z}$ in step 2 can be implemented in a parallel manner, such as using the MapReduce framework \cite{miner2012mapreduce}. According to the above observation, ADMM enhances the processing ability of Model P1 for large-scale problems and its practical usefulness for real-world applications.

	\subsection{Determining assignments for non-reference alternatives}
	\label{sec-23}
	
	\noindent Once the optimal solution of $\bm{\theta}$ is obtained by solving Model P1, we can calculate the comprehensive values of both reference alternatives $a \in A^R$ and non-reference alternatives $b \in A^T$ using the employed value function model. Then, we determine the assignments of non-reference alternatives $b \in A^T$ based on the valued decision examples. Because each reference alternative $a \in A^R$ is assigned to multiple classes with respective credibility degrees, we cannot determine a crisp assignment for any non-reference alternatives $b \in A^T$. Instead, we propose to calculate a valued assignment for each $b \in A^T$ with a credibility vector ${\bm{\sigma }}\left( b \right) = {\left( {{\sigma _1}\left( b \right),...,{\sigma _q}\left( b \right)} \right)^{\text{T}}}$ such that ${\sigma _r}\left( b \right) \geqslant 0$, $r=1,...,q$, and $\sum\limits_{r = 1}^q {{\sigma _r}\left( b \right)}  = 1$. In line with the procedure for inferring a preference model from valued assignment example, we wish the valued assignment of $b$ should be consistent with the valued decision examples as credibly as possible. Therefore, a linear programming model for deriving ${\bm{\sigma }}\left( b \right) = {\left( {{\sigma _1}\left( b \right),...,{\sigma _q}\left( b \right)} \right)^{\text{T}}}$ is developed as follows:
	\begin{equation*}
	\begin{gathered}
	\left( {{\text{P2}}} \right):\;Minimize\;h\left( {\bm{\sigma }} \right) = \sum\limits_{a \in {A^R}} {\left( {\sum\limits_{s = 1}^{q - 1} {\sum\limits_{r = s + 1}^q {{\sigma _s}\left( a \right){\sigma _r}\left( b \right)\left( {U\left( a \right) - U\left( b \right)} \right)} } } \right.}  \hfill \\
	\;\;\;\;\;\;\;\;\;\;\;\;\;\;\;\;\;\;\;\;\;\;\;\;\;\;\;\;\;\;\;\;\;\;\;\;\;\;\;\;\;\;\left. { + \sum\limits_{s = 1}^q {{\sigma _s}\left( a \right){\sigma _s}\left( b \right)\left| {U\left( a \right) - U\left( b \right)} \right|}  - \sum\limits_{s = 2}^q {\sum\limits_{r = 1}^{s - 1} {{\sigma _s}\left( a \right){\sigma _r}\left( b \right)\left( {U\left( a \right) - U\left( b \right)} \right)} } } \right), \hfill \\
	\;\;\;\;\;\;\;\;\;\;\;{\text{s}}{\text{.t}}{\text{.}}\;\sum\limits_{r = 1}^q {{\sigma _r}\left( b \right)}  = 1, \hfill \\
	\;\;\;\;\;\;\;\;\;\;\;\;\;\;\;\;\;\;{\sigma _r}\left( b \right) \geqslant 0,\;\;r = 1,...,q. \hfill \\ 
	\end{gathered}
	\end{equation*}
	Model P2 aims to determine a credibility vector ${\bm{\sigma }}\left( b \right) = {\left( {{\sigma _1}\left( b \right),...,{\sigma _q}\left( b \right)} \right)^{\text{T}}}$ such that the difference between $U(b)$ and $U(a)$ for each $a \in A^R$ is optimized as credibly as possible. About Model P2, there are two useful propositions:
	\begin{description}
		\item[Proposition 1.] For each non-reference alternative $b \in A^T$ and each class $Cl_r$, $r=1,...,q$, let us define:
		\begin{equation*}
		{\Gamma _r}\left( b \right) = \sum\limits_{a \in {A^R}} {\left( {\sum\limits_{s = 1}^{r - 1} {{\sigma _s}\left( a \right)\left( {U\left( a \right) - U\left( b \right)} \right)}  + {\sigma _r}\left( a \right)\left| {U\left( a \right) - U\left( b \right)} \right| - \sum\limits_{s = r + 1}^q {{\sigma _s}\left( a \right)\left( {U\left( a \right) - U\left( b \right)} \right)} } \right)}.
		\end{equation*}
		(a) if there exists $r \in \left\{ {1,...,q} \right\}$ such that ${\Gamma _r}\left( b \right) < {\Gamma _{r'}}\left( b \right)$ for any $r' = 1,...,r - 1,r + 1,...,q$, the optimal solution of Model P2 is ${\sigma _r}\left( b \right) = 1$ and ${\sigma _{r'}}\left( b \right) = 0$ for any $r' = 1,...,r - 1,r + 1,...,q$, which says that $b$ should be assigned to class $Cl_r$ definitely.
		
		(b) if there exists a subset $\Lambda  \subseteq \left\{ {1,...,q} \right\}$ such that ${\Gamma _r}\left( b \right) = {\Gamma _{r'}}\left( b \right)$ for any $r,r' \in \Lambda$ and ${\Gamma _r}\left( b \right) < {\Gamma _{r''}}\left( b \right)$ for $r \in \Lambda $ and $r'' \in \left\{ {1,...,q} \right\}\backslash \Lambda $, Model P2 has infinitely many solutions satisfying $\sum\limits_{r \in \Lambda } {{\sigma _r}\left( b \right)}  = 1$ and ${\sigma _{r''}}\left( b \right) = 0$ for any $r'' \in \left\{ {1,...,q} \right\}\backslash \Lambda$. In this case, we can set ${\sigma _r}\left( b \right) = {1 \mathord{\left/
				{\vphantom {1 {\left| \Lambda  \right|}}} \right.
				\kern-\nulldelimiterspace} {\left| \Lambda  \right|}}$ for any $r \in \Lambda $, where $\left| \Lambda  \right|$ is the number of elements in $\Lambda $, which means that $b$ can be assigned to any class $C{l_r}$, $r \in \Lambda$, with the same credibility degree ${\sigma _r}\left( b \right) = {1 \mathord{\left/
				{\vphantom {1 {\left| \Lambda  \right|}}} \right.
				\kern-\nulldelimiterspace} {\left| \Lambda  \right|}}$.
	\end{description}
	Proof. See e-Appendix A (supplementary material available on-line). \qed
	
	Proposition 1 indicates that the assignment of any non-reference alternative $b \in A^T$ can be determined by examining each ${\Gamma _r}\left( b \right)$, $r=1,...,q$, and the classes $Cl_r$ with the least ${\Gamma _r}\left( b \right)$ are the most credible assignments. Actually, ${\Gamma _r}\left( b \right)$ is equal to the value of the objective of Model P2 when $\sigma_r \left( b \right) = 1$ and $\sigma_{r'} \left( b \right) = 0$ for $r' \ne r$.
	
	\begin{description}
		\item[Proposition 2.] For any non-reference alternatives $b,b' \in {A^T}$, suppose that $U\left( b \right) \geqslant U\left( {b'} \right)$, and Model P2 determines crisp assignments $Cl_r$ and $Cl_{r'}$ for $b$ and $b'$, respectively. Then, it must be that $r \geqslant r'$, that is, the assignment of $b$ is at least as good as the assignment of $b'$.
	\end{description}
	Proof. See e-Appendix B (supplementary material available on-line). \qed
	
	Note that, for any non-reference alternatives $b,b' \in {A^T}$, if Model P2 assigns them to multiple classes, we can also derive that the assignment of $b$ is at least as good as the assignment of $b'$, which can be analyzed in an analogous way. Proposition 2 proves that the assignments of non-reference alternatives determined by Model P2 are consistent with the sorting rule in the example-based sorting procedure.
	
	In real-world applications, the assignment of a non-reference alternative $b \in A^T$ determined by Model P2 is often unique because we rarely encounter a situation where more than one class $Cl_r$ have the same ${\Gamma _r}\left( b \right)$. However, for some problems, we may hope to obtain such results that each non-reference alternative $b \in A^T$ is assigned to multiple classes with non-zero credibility degrees, rather than a crisp assignment. Therefore, we propose to use the \emph{softmax} function to derive a credibility vector ${\bm{\sigma }}\left( b \right) = {\left( {{\sigma _1}\left( b \right),...,{\sigma _q}\left( b \right)} \right)^{\text{T}}}$ with each $\sigma _ r \left( b \right) > 0 $, $r=1,...,q$. The softmax function is often used to transform a vector of real numbers to a multinoulli probability distribution proportional to the exponentials of each input number, which has been widely used in multi-class logistic regression, linear discriminant analysis, artificial neural network (particularly, deep learning)~\cite{goodfellow2016deep, murphy2012machine} and discrete choice model~\cite{train2009discrete}. In our context, the credibility vector ${\bm{\sigma }}\left( b \right) = {\left( {{\sigma _1}\left( b \right),...,{\sigma _q}\left( b \right)} \right)^{\text{T}}}$ for the valued assignment of each non-reference alternative $b \in A^T$ can be obtained using the softmax function as follows:
	\begin{equation*}
	{\sigma _r}\left( b \right) = \frac{{\exp \left( - {{\Gamma _r}\left( b \right)} \right)}}
	{{\sum\limits_{s = 1}^q {\exp \left( - {{\Gamma _s}\left( b \right)} \right)} }},\;\;r = 1,...,q,
	\end{equation*}
	where $\exp \left(  \cdot  \right)$ is the exponential function with respect to the mathematical constant $e = 2.71828...$. The less ${{\Gamma _r}\left( b \right)}$ is, the greater ${\sigma _r}\left( b \right)$. Note that although $-{{\Gamma _r}\left( b \right)}$ could be less than zero, ${\sigma _r}\left( b \right)$ derived from the softmax function ensures ${\sigma _r}\left( b \right) > 0$, $r=1,...,q$, and $\sum\limits_{r = 1}^q {{\sigma _r}\left( b \right)}  = 1$. Obviously, class $Cl_r$ with the least ${\Gamma _r}\left( b \right)$ has the greatest ${\sigma _r}\left( b \right)$, which means $b$ can be assigned to class $Cl_r$ with the greatest credibility. Such an observation is consistent with the assignment determined by Model P2. In this way, we derive a ``soft'' valued assignment for each non-reference alternative in contrast to the ``hard'' assignment determined by Model P2.
	
	\subsection{Adjusting classification performance across classes according to class priorities}
	\label{sec-24}
	
	\noindent In constructing a preference model from valued assignment examples introduced in Section \ref{sec-22}, an important assumption is the equal priorities for all classes, such that the classification performance for each class is addressed in a fair way. However, this is not always the case in many real-world applications, such as credit rating, medical diagnostics, etc., where we need to pay more attention to some particular classes. In this case, the DM may allocate priorities to respective classes and requires to obtain different classification performance according to the specified class priorities. In this section, we propose a method for adjusting classification performance across classes based on the initial preference model constructed by the optimization Model P1. Such a method can be seen as a complementary component of the analytical framework and the DM can decide whether to launch it.
	
	At the beginning of this method, the DM is required to review the classification performance of the initial preference model suggested by Model P1 on the reference set (i.e., the fitting ability on valued decision examples). Specifically, for each reference alternative $a \in A^R$, we regard it as a fictitious non-reference alternative, and use the method discussed in Section \ref{sec-23} to predict its valued assignment (denoted by ${\bm{\sigma }}'\left( a \right) = {\left( {{\sigma _1}'\left( a \right),...,{\sigma _q}'\left( a \right)} \right)^{\text{T}}}$), and then compare the predicted valued assignment to its actual one ${\bm{\sigma }}\left( a \right) = {\left( {{\sigma _1}\left( a \right),...,{\sigma _q}\left( a \right)} \right)^{\text{T}}}$. Then, according to ${\bm{\sigma }}\left( a \right) $ and ${\bm{\sigma }}'\left( a \right) $, we can derive two ranking lists of classes for $a$, denoted by $C{l_{{\phi _1}\left( a \right)}},...,C{l_{{\phi _q}\left( a \right)}}$ and $C{l_{{\phi _1}'\left( a \right)}},...,C{l_{{\phi _q}'\left( a \right)}}$, respectively, such that ${\sigma _{{\phi _i}\left( a \right)}} > {\sigma _{{\phi _j}\left( a \right)}}$ (or ${\sigma _{{\phi _i}'\left( a \right)}}' > {\sigma _{{\phi _j}'\left( a \right)}}'$), for $i<j$, where all classes rank in a descending order of credibility degrees. Then, for each class $Cl_r$, $r=1,...,q$, we can define the following cardinal and ordinal classification performance measures, respectively, as follows:
	\begin{equation*}
	{\text{CardP}}{{\text{f}}_r} = \frac{1}
	{{\left| {{A^R}} \right|}}\sum\limits_{a \in {A^R}} {\left| {{\sigma _r}\left( a \right) - {\sigma _r}\left( a \right)'} \right|} ,
	\end{equation*}
	\begin{equation*}
	{\text{OrdP}}{{\text{f}}_r} = \frac{1}
	{{\left( {q - 1} \right)\left| {{A^R}} \right|}}\sum\limits_{a \in {A^R}} {\left| {{\text{po}}{{\text{s}}_\phi }\left( {a,r} \right) - {\text{po}}{{\text{s}}_{\phi '}}\left( {a,r} \right)} \right|},
	\end{equation*}
	where ${{\text{po}}{{\text{s}}_\phi }\left( {a,r} \right)}$ and ${{\text{po}}{{\text{s}}_{\phi '}}\left( {a,r} \right)}$ represent the positions of class $Cl_r$ in the ranking lists $C{l_{{\phi _1}\left( a \right)}},...,C{l_{{\phi _q}\left( a \right)}}$ and $C{l_{{\phi _1}'\left( a \right)}},...,C{l_{{\phi _q}'\left( a \right)}}$, respectively. ${\text{CardP}}{{\text{f}}_r}$ quantifies the average difference between the credibility degrees of class $Cl_r$ in the actual and predicted credibility distributions for all reference alternatives $a \in A^R$, while ${\text{OrdP}}{{\text{f}}_r}$ measures the average distance between the positions of class $Cl_r$ in the actual and predicted ranking lists for all reference alternatives $a \in A^R$. Note that both ${\text{CardP}}{{\text{f}}_r}$ and ${\text{OrdP}}{{\text{f}}_r}$ are normalized within the interval [0, 1]. Obviously, the less ${\text{CardP}}{{\text{f}}_r}$ and ${\text{OrdP}}{{\text{f}}_r}$, the better performance is achieved on class $Cl_r$.
	
	The measures ${\text{CardP}}{{\text{f}}_r}$ and ${\text{OrdP}}{{\text{f}}_r}$ for each class $Cl_r$, $r=1,...,q$, are submitted to the DM, and (s)he can review such results and then decides whether to adjust the classification performance. If the DM thinks the performance on some class $Cl_s$ is relatively low, (s)he may require to improve the performance on this class. 
	However, acquiring the precise values of the priorities for each class from the DM is a difficult task. Instead, we can require the DM to specify a priority ranking of all $q$ classes in the following form
	\begin{equation*}
	C{l_{\tau \left( 1 \right)}},\;\;C{l_{\tau \left( 2 \right)}},\;\;...,\;\;C{l_{\tau \left( q \right)}},
	\end{equation*}
	where $\tau \left(  \cdot  \right) \in \{1,...,q\}$ is the permutation on the set of indices of classes according to the specified priorities, such that class $C{l_{\tau \left( s \right)}}$ is prior to class $C{l_{\tau \left( s+1 \right)}}$, $s=1,...,q-1$, which means that once the performance of class $C{l_{\tau \left( s+1 \right)}}$ is improved, that of class $C{l_{\tau \left( s \right)}}$ should also be improved. The DM wishes that the performance of each class improves according to this priority order.
	
	To implement flexible adjustment of classification performance across classes, our method pays more attention to classes with higher priorities and aims to improve the credible consistency between the reference alternatives that can be assigned to these classes with certain credibility degrees and other reference alternatives that are assigned to other classes. Specifically, the credible consistency for each class $Cl_s$, $s=1,...,q$, can be quantified as follows:
	\begin{equation*}
	{O_s} = \sum\limits_{r = 1}^{s - 1} {\sum\limits_{{a_i},{a_j} \in {A^R}} {{\sigma _s}\left( {{a_i}} \right){\sigma _r}\left( {{a_j}} \right)\left( {U\left( {{a_i}} \right) - U\left( {{a_j}} \right)} \right)} }  + \sum\limits_{r = s + 1}^q {\sum\limits_{{a_i},{a_j} \in {A^R}} {{\sigma _s}\left( {{a_i}} \right){\sigma _r}\left( {{a_j}} \right)\left( {U\left( {{a_j}} \right) - U\left( {{a_i}} \right)} \right)} },
	\end{equation*}
	where the left part concerns the value difference between the reference alternatives $a_i$ that can be assigned to class $Cl_s$ with certain credibility degrees and other $a_j$ that are classified into a worse class, while the right part measures the value difference between these reference alternatives $a_i$ and those $a_j$ that come from a better class. According to the rule for the example-based sorting procedure, the greater $O_s$ is, the more likely it is to achieve an~improved performance on class $Cl_s$. Therefore, we can consider to increase $O_s$ for classes $Cl_s$, $s=1,...,q$, according to the specified priority order.
	
	\begin{description}
		\item[Definition 3.] \cite{nocedal1999numerical} Let ${{\bm{\theta }}}$ be a parameter vector of the employed preference model. Regarding $O_s$, $s=1,...,q$, a vector ${\mathbf{d}} \ne {\mathbf{0}}$ is said to be an ascent direction of $O_s$ at ${{\bm{\theta }}}$, if there exits a positive number $\delta$ such that $O_s \left( {{\bm{\theta }} + \lambda {\mathbf{d}}} \right) > O_s \left( {{\bm{\theta }}} \right)$ for any scalar $\lambda  \in \left( {0,\delta } \right)$.
	\end{description}
	
	\begin{description}
		\item[Proposition 3.] \cite{nocedal1999numerical} Let ${{\bm{\theta }}}$ be a parameter vector of the employed preference model. Regarding $O_s$, $s=1,...,q$, if a vector ${\mathbf{d}} \ne {\mathbf{0}}$ satisfies $\nabla {O_s}^{\text{T}}{\mathbf{d}} > 0$ where $\nabla {O_s}$ stands for the gradient of $O_s$, there exists a positive number $\delta$ such that $O_s \left( {{\bm{\theta }} + \lambda {\mathbf{d}}} \right) > O_s \left( {{\bm{\theta }}} \right)$ for any scalar $\lambda  \in \left( {0,\delta } \right)$, that is, ${\mathbf{d}}$ is an ascent direction of $O_s$ at~${{\bm{\theta }}}$.
	\end{description}
	
	\noindent According to the above definition and proposition, increasing $O_s$, $s=1,...,q$, can be done by finding an ascent direction $\mathbf{d}$ for $O_s$. For this purpose, let us consider the following mixed-integer programming model:
	\begin{equation*}
	\begin{gathered}
	\left( {{\text{P3}}} \right):\;Maximize\;\sum\limits_{s = 1}^q {{v_s}} , \hfill \\
	\;\;\;\;\;{\text{s}}{\text{.t}}{\text{.}}\;\left( {{\text{LC1}}} \right)\;\nabla {O_s {|_{{\bm{\theta }} = \bm {\hat {\theta }}}}}^{\text{T}}{\mathbf{d}} + Q\left( {1 - {v_s}} \right) \geqslant 0,\;\;s = 1,...,q, \hfill \\
	\;\;\;\;\;\;\;\;\;\;\;\left( {{\text{LC2}}} \right)\;{v_{\tau \left( s \right)}} \geqslant {v_{\tau \left( {s + 1} \right)}},\;\;s = 1,...,q - 1, \hfill \\
	\;\;\;\;\;\;\;\;\;\;\;\left( {{\text{LC3}}} \right)\;{v_s} \in \left\{ {0,1} \right\},\;\;s = 1,...,q, \hfill \\
	\;\;\;\;\;\;\;\;\;\;\;\left( {{\text{LC4}}} \right)\;{\left| {\left[ {\mathbf{d}} \right]_j} \right|} \leqslant 1,\;\;j = 1,...,\dim \left( {\mathbf{d}} \right), \hfill \\ 
	\;\;\;\;\;\;\;\;\;\;\;E_{{\text{BASE}}}^M, \hfill \\ 
	\end{gathered}
	\end{equation*}
	where $\bm{\hat \theta}$ represents the current value of the parameter vector $\bm{\theta}$, $Q$ is an auxiliary constant equal to a sufficiently large positive value such that $Q \geqslant {\left\| {\nabla {O_s}{|_{{\bm{\theta }} = {\bm{\hat \theta }}}}} \right\|_1}$, $v_s$ for $s=1,...,q$ are binary variables, ${{\left[ {\mathbf{d}} \right]_j}}$ is the $j$-th entry of $\mathbf{d}$, $\dim \left( {\mathbf{d}} \right)$ represents the dimension of $\mathbf{d}$, and $M \in \left\{ {{\text{LINEAR,}}\;{\text{PIECEWISE - LINEAR,}}\;{\text{SPLINE,}}\;{\text{GENERAL}}} \right\}$ so that the above model applies to different types of value functions. If $v_s = 1$, constraint (LC1) amounts to $\nabla {O_s{|_{{\bm{\theta }} = \bm {\hat {\theta }}}}}^{\text{T}}{\mathbf{d}} \geqslant 0$, which requires to find a direction $\mathbf{d}$ along which $O_s$ increases. Constraint (LC2) ensures that $O_s$ for all classes $Cl_s$, $s=1,...,q$, increase according to the specified priority order, such that once $O_{s+1}$ for class $Cl_{s+1}$ with a lower priority increases, $O_{s}$ for class $Cl_{s}$ with a higher priority should also increase. Constraint (LC4) guarantees to derive a bounded $\mathbf{d}$. Model P3 aims to maximize the number of classes whose $O_s$ can be increased according to the specified priority order. Let $\mathbf{d} ^ *$ and $v_s^*$ be the values of $\mathbf{d}$ and $v_s$ at the optimum, respectively, which indicate that $O_s$ for classes $Cl_s$ with $v_s^* = 1$ increase along the direction $\mathbf{d} ^ *$ to improve the credible consistency for these classes. Then, we can adjust the current preference model by solving the following linear programming model:
	\begin{equation*}
	\begin{gathered}
	\left( {{\text{P4}}} \right):\;Maximize\;\sum\limits_{s = 1,...,q:\;{v_s^*} = 1} {{O_s}{|_{{\bm{\theta }} = \bm {\hat {\theta }} + \lambda {\mathbf{d}}}}} , \hfill \\
	\;\;\;\;\;{\text{s}}{\text{.t}}{\text{.}}\;\lambda  \geqslant 0, \hfill \\
	\;\;\;\;\;\;\;\;\;\;\;E_{{\text{BASE}}}^M{|_{{\bm{\theta }} = \bm {\hat{\theta }}   + \lambda {\mathbf{d}}}}, \hfill \\ 
	\end{gathered}
	\end{equation*}
	where the variable $\lambda$ represents a step. Model P4 aims to maximize $O_s$ for classes $Cl_s$ such that $v_s^* = 1$ by adjusting $\bm{\hat \theta}$ along the ascent direction $\mathbf{d} ^ *$, so that the credible consistency for these classes is improved. Once the value of $\lambda$ at the optimum (denoted by $\lambda ^ *$) is achieved, we can derive a new value of the parameter vector $\bm{\theta}$ according to the following formula
	\begin{equation*}
	\bm{\hat \theta}' = \bm{\hat \theta} + {\lambda ^ *} {\mathbf{d} ^ *},
	\end{equation*}
	where $\bm{\hat \theta}'$ stands for the adjusted value of the parameter vector $\bm{\theta}$.
	
	The procedure for adjusting $O_s$ according to the specified priority order is an iterative process, which can be organized as Algorithm 1. Threshold $\zeta$ is a positive value used to control the complexity of the adjusted preference model $U'$, and a stopping criterion of Algorithm 1 consists in that the complexity measure $\Omega \left( {U'} \right)$ exceeds threshold $\left( {1 + \zeta } \right)\Omega \left( U \right)$. Without threshold $\zeta$, we may obtain an adjusted preference model that over-fits the decision examples and has poor generalization performance on new alternatives. Note that $\zeta$ can be specified in advance (e.g., 10\%, 20\%, etc.), or adjusted during the process by checking the classification performance on a~subset of reference alternatives for validation (i.e., cross-validation).
	
	\begin{algorithm}	
		\caption{Method for adjusting classification performance across classes according to specified priority order.}
		\begin{algorithmic}[1]
			\REQUIRE ~~\\
			Solution ${\bm{\hat \theta }}$ output by Model P1, complexity measure $\Omega \left( U \right)$ of preference model $U$ corresponding to $\bm{\hat \theta }$, priority ranking of classes $C{l_{\tau \left( 1 \right)}},\;C{l_{\tau \left( 2 \right)}},\;...,\;C{l_{\tau \left( q \right)}}$, threshold $\zeta$.
			\STATE Solve Model P3 to derive ascent direction $\mathbf{d} ^ *$ and identify classes $Cl_s$ such that $v_s^* = 1$.
			\IF{$\mathbf{d} = \mathbf{0}$}
			\STATE Terminate.
			\ENDIF
			\STATE Solve Model P4 to obtain step $\lambda ^ *$.
			\IF{$\lambda ^ * = 0$}
			\STATE Terminate.
			\ELSE 
			\STATE $\bm{\hat \theta}' \leftarrow \bm{\hat \theta} + {\lambda ^ *} {\mathbf{d} ^ *}$.
			\STATE Derive new preference model $U'$ according to ${\bm{\hat \theta }}'$ and then measure its complexity $\Omega \left( {U'} \right)$.
			\IF{$\Omega \left( {U'} \right) > \left( {1 + \zeta } \right)\Omega \left( U \right)$}
			\STATE Terminate.
			\ELSE
			\STATE $\bm{\hat \theta} \leftarrow \bm{\hat \theta}'$, $U \leftarrow U'$.
			\STATE Go to step 1.
			\ENDIF		
			\ENDIF
			\ENSURE ~~\\
			Adjusted solution ${\bm{\hat \theta }}$ and corresponding adjusted preference model $U$.\\
		\end{algorithmic}	
	\end{algorithm}

	\section{Experimental analysis}
	\label{sec-3}
	
\begin{figure}[!htbp]
		\centering
		\scalebox{1} {\includegraphics[width=1\textwidth]{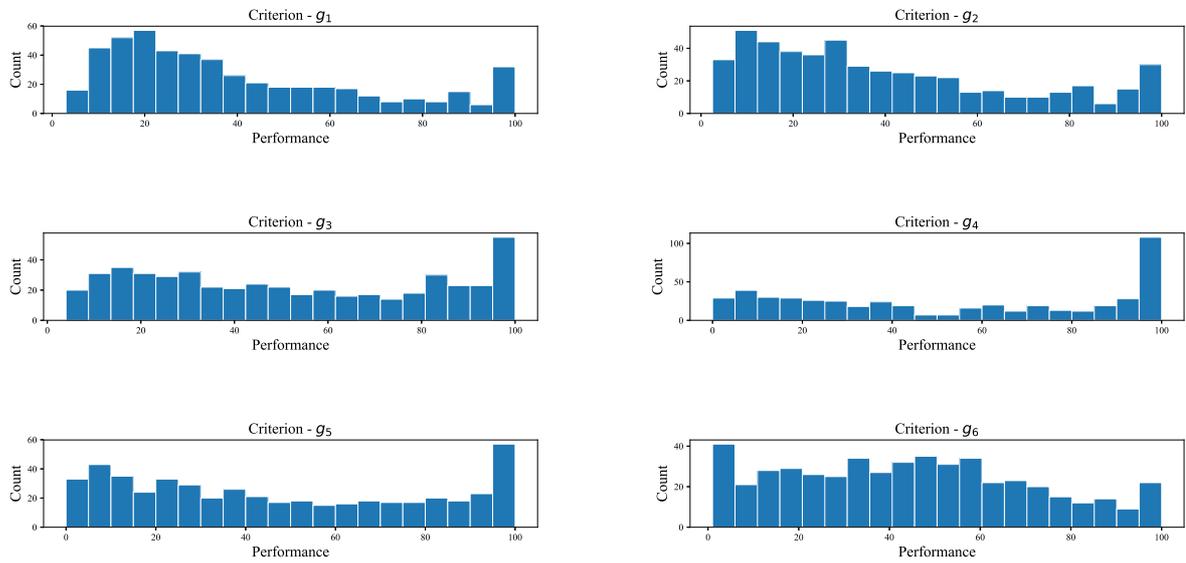}}
		\caption{\label{fig-1}Distribution of performance on each criterion.}
	\end{figure}

	\noindent In this section, we validate the practical performance of the proposed framework on a real-word dataset, which is collected from the QS World University Ranking\footnote{https://www.topuniversities.com/university-rankings/world-university-rankings/2020}. This dataset provides an overall ranking of 500 universities from all over the world according to six evaluation criteria, including ($g_1$) citation per faculty, ($g_2$) international students, ($g_3$) international faculty, ($g_4$) faculty student, ($g_5$) employer reputation, and ($g_6$) academic reputation. The ranking of 500 universities is derived by aggregating the performances of each university on all criteria using the criteria weights $w_1=0.4$, $w_2=0.1$, $w_3=0.2$, $w_4=0.05$, $w_5=0.05$ and $w_6=0.2$, where $w_j$ is the weight of criterion $g_j$, $j=1,...,6$. All criteria are of gain-type. A descriptive summary of the dataset is provided in Table~\ref{tab-1} and the distribution of performance values on each criterion is depicted in Figure~\ref{fig-1}. One can observe that the distribution of performance values on most criteria is not uniform: the distribution of performance levels on $g_1$ and $g_2$ is skewed to the left, whereas a significant proportion of performance values are located in the interval [95.0, 100.0] on $g_3$, $g_4$ and $g_5$. This observation incurs a careful setting for the experimental analysis, which will be discussed later.
	
	\begin{table}[!htbp] \caption{\label{tab-1}Descriptive summary of QS World University Ranking dataset.}
		\centering
		\footnotesize
		\begin{tabular}{cccc}
			\hline
			Criterion & \multicolumn{1}{l}{Min. Performance} & \multicolumn{1}{l}{Max. Performance} & \multicolumn{1}{l}{Avg. Performance} \\
			\hline
			$g_1$ & 3.0   & 100.0  & 39.7  \\
			$g_2$ & 2.5   & 100.0  & 40.1  \\
			$g_3$ & 4.0   & 100.0  & 51.9  \\
			$g_4$ & 0.0   & 100.0  & 53.4  \\
			$g_5$ & 0.0   & 100.0  & 47.2  \\
			$g_6$ & 1.0   & 100.0  & 44.2  \\
			\hline
		\end{tabular}
	\end{table}
	
	As our framework can be equipped with linear, piecewise-linear, splined, or general monotone marginal value functions, we implement the four variants in the experimental study. Furthermore, the UTADIS method as well as its three variants (UTADIS I, II, \& III) \cite{doumpos2002multicriteria} and the method proposed by \cite{liu2019preference} are investigated to compare with the four variants of the proposed framework. Both the UTADIS family and the method proposed by \cite{liu2019preference} employ an additive value function model composed of piecewise-linear marginal value functions as the preference model. The UTADIS method constructs a preference model by minimizing the sum of misclassification errors for all reference alternatives, and then other or near optimal solutions are explored and averaged to derive a final preference model in the post optimality analysis stage. In addition to the classification error,
	UTADIS I maximizes the distances of the correctly classified alternatives from the class thresholds, so that a sharp discrimination is achieved. UTADIS II uses a mixed-integer programming model to minimize the number of misclassified alternatives, rather than their magnitude. UTADIS III combines UTADIS I and II. The model proposed by \cite{liu2019preference} is based on the regularization framework and aims to construct a preference model composed of marginal value functions that are as ``smooth'' as possible while minimizing the sum of inconsistency levels for pairs of reference alternatives coming from distinct classes.
	
	In the experimental setting, for the methods that require dividing performance scales into a number of equal-length sub-intervals (including the piecewise-linear and splined variants of the proposed framework, the UTADIS method and its three variants, and the method proposed by \cite{liu2019preference}), the number of sub-intervals is perceived as a hyper-parameter, which is determined using cross-validation by examining the following values $\gamma_j \in \{1, 2, 3, 4, 5, 6, 7, 8, 9, 10\}$. In making trade-off between the preference model's fitting ability and its complexity control, we also use the cross-validation method to determine the hyper-parameters $C$, $C_1$ and $C_2$ from the following candidates \{${10^{ - 8}}$, $5 \times {10^{ - 8}}$, ${10^{ - 7}}$, $5 \times {10^{ - 7}}$,..., ${10^7}$, $5 \times {10^7}$, ${10^8}$, $5 \times {10^8}$\}. To construct a sorting problem, we assign all universities to five preference-ordered classes $CL = \left\{ {C{l_1},C{l_2},C{l_3},C{l_4}},C{l_5} \right\}$ according to their overall ranking: $Cl_5$, $Cl_4$, $Cl_3$, $Cl_2$ and $Cl_1$ consist of universities that are ranked in the intervals [1, 100], [101, 200], [201, 300], [301, 400], and [401, 500], respectively. In evaluating the performance of the proposed framework, the original dataset is randomly split into two parts, 70\% (referred to as $A^R$) for constructing the preference model and 30\% (referred to as $A^T$) for testing the predictive accuracy of the constructed model. Note that the distribution of universities from respective classes in the training and test sets are ensured to be (approximately) the same with that in the original dataset. This procedure is repeated 100 times, and the results are finally averaged. Note that, in each run, we use different sorting methods to construct a preference model and then test its performance on the same $A^R$ and $A^T$, so that the final averaged results on 100 runs can be used to compare different methods. All considered methods are implemented using Python and the involved optimization models are solved with the CVXPY optimization package\footnote{https://www.cvxpy.org/}.
	
	To evaluate the performance of a sorting method, we can consider the following measures including Top-$N$ accuracies and Kendall's tau coefficient. Specifically, for each non-reference alternative $b \in A^T$, let ${\bm{\sigma }}\left( b \right) = {\left( {{\sigma _1}\left( b \right),...,{\sigma _q}\left( b \right)} \right)^{\text{T}}}$ and ${\bm{\sigma }}'\left( b \right) = {\left( {{\sigma _1}'\left( b \right),...,{\sigma _q}'\left( b \right)} \right)^{\text{T}}}$ denote its actual and predicted credibility degrees for valued assignment. According to ${\bm{\sigma }}\left( b \right)$ and ${\bm{\sigma }}'\left( b \right)$, we can derive two ranking lists of classes for $b$, denoted by $C{l_{{\phi _1}\left( b \right)}},...,C{l_{{\phi _q}\left( b \right)}}$ and $C{l_{{\phi _1}'\left( b \right)}},...,C{l_{{\phi _q}'\left( b \right)}}$, respectively, such that ${\sigma _{{\phi _i}\left( b \right)}} > {\sigma _{{\phi _j}\left( b \right)}}$ (or ${\sigma _{{\phi _i}'\left( b \right)}}' > {\sigma _{{\phi _j}'\left( b \right)}}'$), for $i<j$. Then, for any $N=1,...,q$, let ${\Theta ^N}\left( b \right)$ and $\Theta {'^N}\left( b \right)$ denote the top $N$ classes with the greatest credibility degrees according to the above two ranking lists, respectively. Moreover, let $n_c$ and $n_d$ represent the numbers of concordant and discordant pairs of classes in the above two ranking lists, respectively. Then, with the use of this notation, Top-$N$ accuracies and Kendall's tau for $b$ are calculated as follows:
	\begin{equation*}
	{\text{Accuracy@}}N\left( b \right) = \frac{{\left| {{\Theta ^N}\left( b \right) \cap \Theta {'^N}\left( b \right)} \right|}}{N},
	\end{equation*}
	\begin{equation*}
	{\text{Kendall's}}\;{\text{tau}}\left( b \right) = \frac{{2\left( {{n_c} - {n_d}} \right)}}
	{{q\left( {q - 1} \right)}}.
	\end{equation*}
	${\text{Accuracy@}}N\left( b \right)$ reflects, in the top $N$ recommendations for alternative $b$, how many classes actually have the greatest $N$ credibility degrees, and ${\text{Kendall's}}\;{\text{tau}}\left( b \right)$ refers to the difference between the proportions of concordant and discordant pairs of classes in the above two ranking lists. Obviously, the greater ${\text{Accuracy@}}N\left( b \right)$ and ${\text{Kendall's}}\;{\text{tau}}\left( b \right)$, the better the classification performance is achieved on alternative $b$. Note that $N$ could be specified by the DM, since it concerns the most credible $N$ assignments for each alternative. Particularly, when $N = 1$, we care about the most credible one for each alternative. Finally, we can average the above measures for all non-reference alternatives $b \in A^T$ and obtain comprehensive performance evaluation for a sorting method.
	
	\subsection{Experiments with crisp assignment examples}
\begin{figure}[!htbp]
		\centering
		\scalebox{1} {\includegraphics[width=1\textwidth]{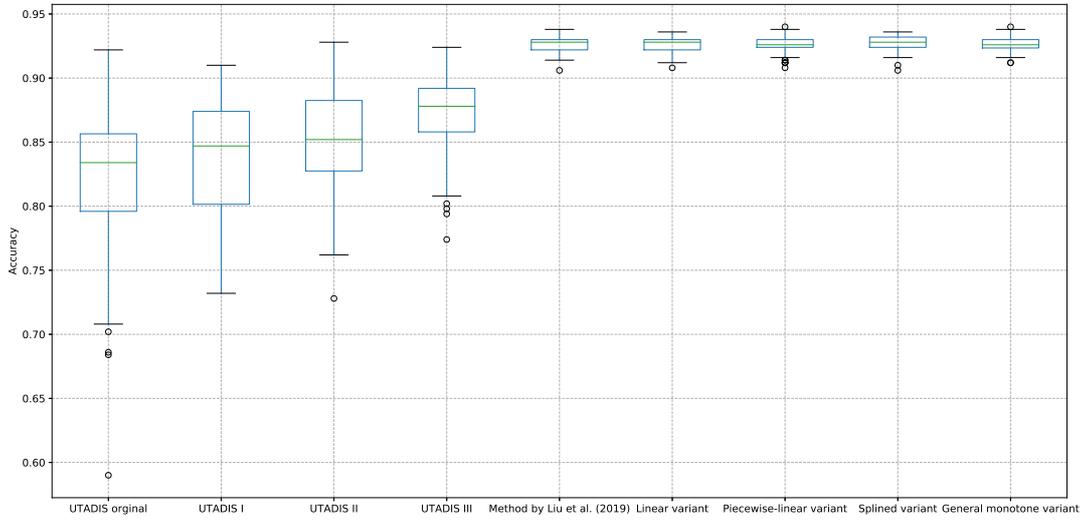}}
		\caption{\label{figure-1}Distribution of classification accuracy of each method for crisp decision examples.}
	\end{figure}

	\noindent We first report the outcomes of applying all above sorting methods to the original crisp decision examples. Since each reference alternative is assigned to only one class precisely, we only measure the classification accuracy (i.e., Accuracy@1) for each sorting method to reflect how many alternatives are correctly classified by each method. Besides, we also report the trade-off weight of marginal value function on each criterion in terms of mean and standard deviation to measure the ability of each method in constructing a preference model that is close to the actual one. The experimental results are summarized in Table \ref{table-1} and the distribution of classification of each method is depicted in Figure \ref{figure-1}. It is apparent that the four variants of the proposed framework and the method proposed by \cite{liu2019preference} achieve higher classification accuracies than the UTADIS family. Although UTADIS I, II, and III improve the original UTADIS method in some aspects, the performance of the three variants is unstable and the classification accuracies could be rather low for some sets of decision examples. Both the four variants of the proposed framework and the method proposed by \cite{liu2019preference} incorporate the advance of regularization techniques and tend to deriving as linear marginal value functions as possible, which are close to the actual preference model (i.e., linear value functions). When referring to the trade-off weight of marginal value function on each criterion, we observe that the averaged results from all sorting methods are close to the actual ones, but the outcomes from the four variants of the proposed framework and the method proposed by \cite{liu2019preference} are more stable than those suggested by the UTADIS family. Furthermore, the difference among the performance from the four variants of the proposed framework and the method proposed by \cite{liu2019preference} is marginal, since the actual preference model is simple and the decision examples over the experimental runs are consistent.
	
	\begin{table}[!htbp] \caption{\label{table-1}Classification performance and trade-off weight of marginal value function derived from respective method in terms of mean and standard deviation for crisp decision examples.}
		\centering
		\tiny
		\begin{tabular}{rccccccc}
			\hline
			\multirow{2}[2]{*}{Method} & \multirow{2}[2]{*}{Accuracy} & \multicolumn{6}{c}{Trade-off weight of marginal value function} \\
			\cmidrule{3-8}      &       & $g_1$ & $g_2$ & $g_3$ & $g_4$ & $g_5$ & $g_6$ \\
			\hline
			UTADIS original & 0.8206 $\pm$ 0.0570 & 0.3998 $\pm$ 0.0689 & 0.1030 $\pm$ 0.0245 & 0.2045 $\pm$ 0.0476 & 0.0508 $\pm$ 0.0138 & 0.0514 $\pm$ 0.0130 & 0.1932 $\pm$ 0.0416 \\
			UTADIS I & 0.8370 $\pm$ 0.0457 & 0.4013 $\pm$ 0.0558 & 0.0993 $\pm$ 0.0197 & 0.2014 $\pm$ 0.0380 & 0.0497 $\pm$ 0.0107 & 0.0497 $\pm$ 0.0107 & 0.1984 $\pm$ 0.0386 \\
			UTADIS II & 0.8519 $\pm$ 0.0393 & 0.3842 $\pm$ 0.0408 & 0.1011 $\pm$ 0.0186 & 0.2052 $\pm$ 0.0339 & 0.0507 $\pm$ 0.0095 & 0.0508 $\pm$ 0.0091 & 0.2077 $\pm$ 0.0282 \\
			UTADIS III & 0.8736 $\pm$ 0.0298 & 0.3930 $\pm$ 0.0327 & 0.1020 $\pm$ 0.0135 & 0.2012 $\pm$ 0.0265 & 0.0505 $\pm$ 0.0080 & 0.0491 $\pm$ 0.0076 & 0.2039 $\pm$ 0.0234 \\
			Method by \cite{liu2019preference} & 0.9267 $\pm$ 0.0057 & 0.4005 $\pm$ 0.0080 & 0.0999 $\pm$ 0.0030 & 0.1996 $\pm$ 0.0053 & 0.0498 $\pm$ 0.0016 & 0.0498 $\pm$ 0.0014 & 0.2001 $\pm$ 0.0054 \\
			Linear variant & 0.9258 $\pm$ 0.0057 & 0.4013 $\pm$ 0.0081 & 0.1000 $\pm$ 0.0028 & 0.1989 $\pm$ 0.0060 & 0.0499 $\pm$ 0.0015 & 0.0499 $\pm$ 0.0016 & 0.1997 $\pm$ 0.0051 \\
			Piecewise-linear variant & 0.9264 $\pm$ 0.0062 & 0.4005 $\pm$ 0.0073 & 0.0997 $\pm$ 0.0029 & 0.2009 $\pm$ 0.0048 & 0.0498 $\pm$ 0.0016 & 0.0497 $\pm$ 0.0015 & 0.1991 $\pm$ 0.0053 \\
			Splined variant & 0.9269 $\pm$ 0.0058 & 0.4009 $\pm$ 0.0080 & 0.1001 $\pm$ 0.0028 & 0.1996 $\pm$ 0.0050 & 0.0498 $\pm$ 0.0013 & 0.0499 $\pm$ 0.0017 & 0.1994 $\pm$ 0.0056 \\
			General monotone variant & 0.9263 $\pm$ 0.0054 & 0.3994 $\pm$ 0.0074 & 0.1004 $\pm$ 0.0029 & 0.1997 $\pm$ 0.0053 & 0.0502 $\pm$ 0.0015 & 0.0500 $\pm$ 0.0015 & 0.2001 $\pm$ 0.0053 \\
			\hline
		\end{tabular}
	\end{table}
	
\begin{figure}[!htbp] 
		\centering
		\subfloat[Linear marginal value function]{\begin{minipage}{9cm}
				\includegraphics[width=1\textwidth]{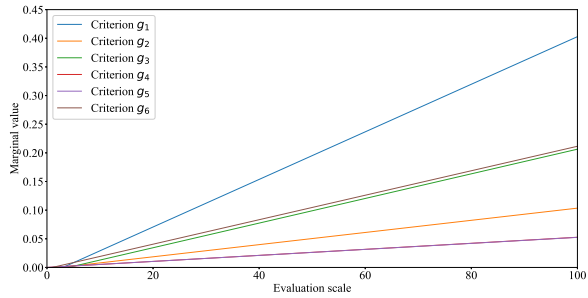}
		\end{minipage}}
		\subfloat[Piecewise-linear marginal value function]{\begin{minipage}{9cm}
				\includegraphics[width=1\textwidth]{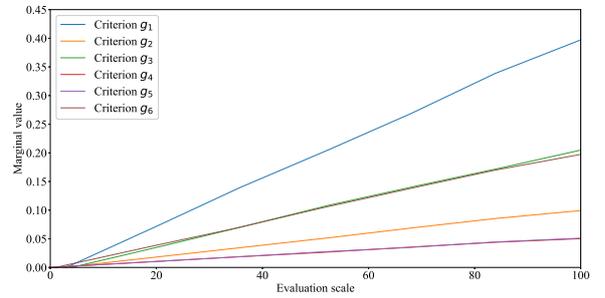}
		\end{minipage}}	
		
		\subfloat[Splined marginal value function]{\begin{minipage}{9cm}
				\includegraphics[width=1\textwidth]{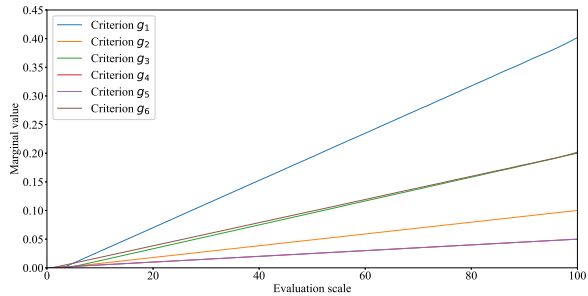}
		\end{minipage}}		
		\subfloat[General monotone marginal value function]{\begin{minipage}{9cm}
				\includegraphics[width=1\textwidth]{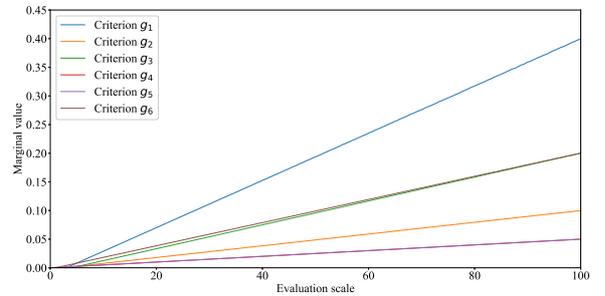}
		\end{minipage}}	
		\caption{\label{figure-2}An example of marginal value functions derived from four variants of proposed framework. (For interpretation of the references to color in this figure, the reader is referred to the web version of this article.)}
	\end{figure}

	For illustrative purpose, let us present the examples of the preference models constructed by the four variants of the proposed framework. Figure \ref{figure-2} illustrates the four types of marginal value functions of the constructed preference models corresponding to the greatest predictive accuracy. The derived linear marginal value functions are completely linearly increasing with respect to the performance values. As for the marginal value functions derived from the piecewise-linear, splined, and general monotone variants, they look almost linear since the regularization term is in favor of functions that are as linear as possible. As expected, the piecewise-linear functions have a~sudden change in slope at breakpoints, although it is very slight in the presented example, whereas the splined functions are completely smooth over the whole performance scales. When it comes to general monotone marginal value functions, they exhibit slight ``zig-zag'' behavior, since we allow each distinct performance value observed over the performance scales to be breakpoints and the constructed marginal value functions are difficult to be completely smooth.
	
	\subsection{Experiments with valued assignment examples}
	\noindent Let us now test the classification performance of the four variants of the proposed framework on valued decision examples. Note that the family of UTADIS and the method proposed by \cite{liu2019preference} cannot address this sorting task, since they only apply to crisp decision examples. To construct valued decision examples, we modify the original dataset in the following way: for any alternative, a majority proportion (including 90\%, 80\%, 70\%, and 60\%) of credibility degree is assigned to its actual assignment, and the remaining (including 10\%, 20\%, 30\%, and 40\%) is allocated to the two classes adjacent to its actual assignment. For example, when the majority proportion of credibility degree that is assigned to the actual assignment is 90\%, for an alternative $a$, of which the actual assignment is $Cl_3$, the constructed distribution of credibility degrees will be ${\bm{\sigma }}\left( a \right) = {\left( {0,\;0.05,\;0.9,\;0.05,\;0} \right)^{\text{T}}}$. Particularly, for an alternative $a'$ that is actually assigned to $Cl_1$ or $Cl_5$, the constructed ${\bm{\sigma }}\left( {a'} \right)$ will be set as ${\bm{\sigma }}\left( {a'} \right) = {\left( {0.9,\;0.1,\;0,\;0,\;0} \right)^{\text{T}}}$ or ${\bm{\sigma }}\left( {a'} \right) = {\left( {0,\;0,\;0,\;0.1,\;0.9} \right)^{\text{T}}}$. Then, we examine the classification performance of different sorting methods in terms of Accuracy@1, Accuracy@2, Accuracy@3, and Kendall's tau for 100 runs on randomly constructed $A^R$ and $A^T$. 
	
\begin{figure}[!htbp] 
		\centering
		\subfloat[Accuracy@1]{\begin{minipage}{9cm}
				\includegraphics[width=1\textwidth]{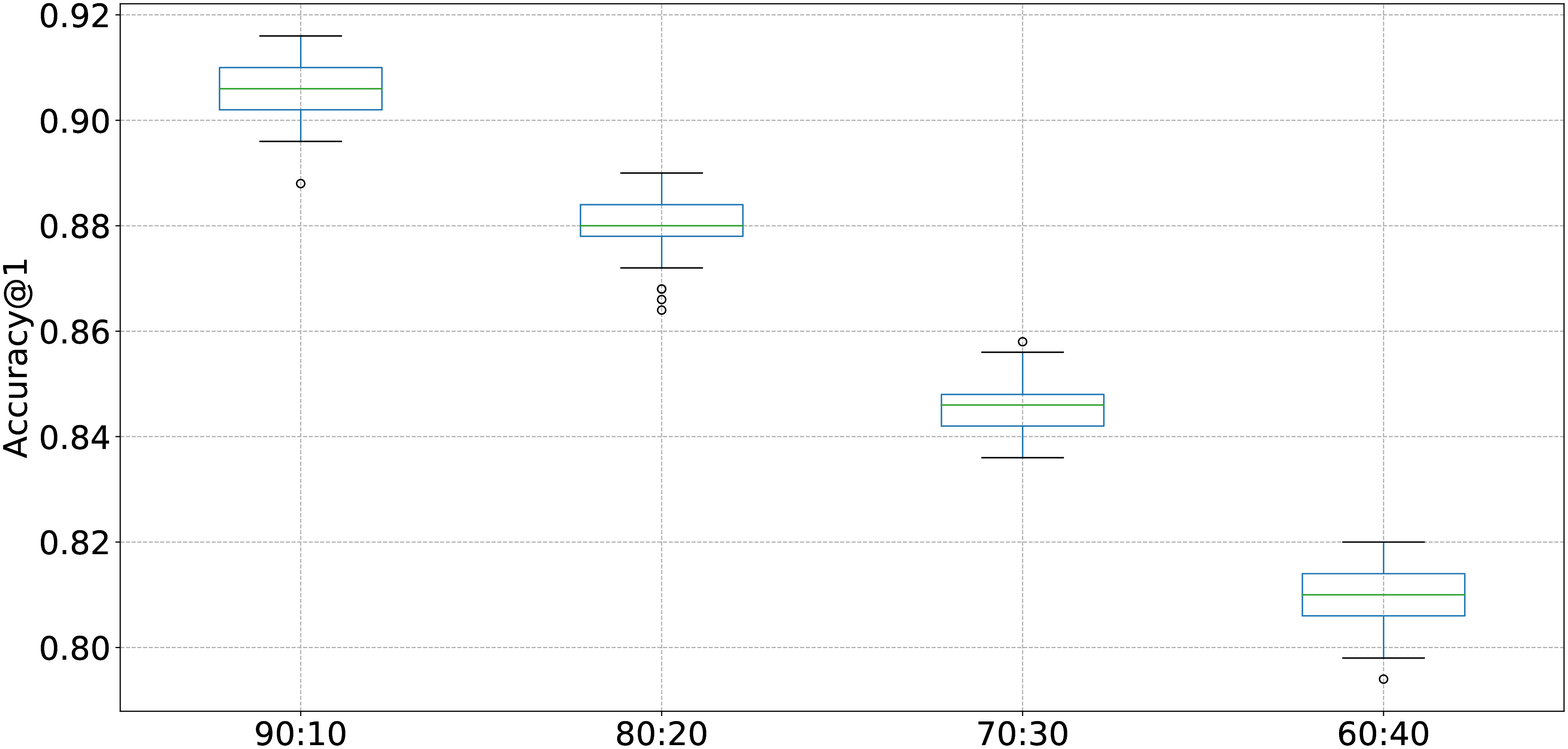}
		\end{minipage}}
		\subfloat[Accuracy@2]{\begin{minipage}{9cm}
				\includegraphics[width=1\textwidth]{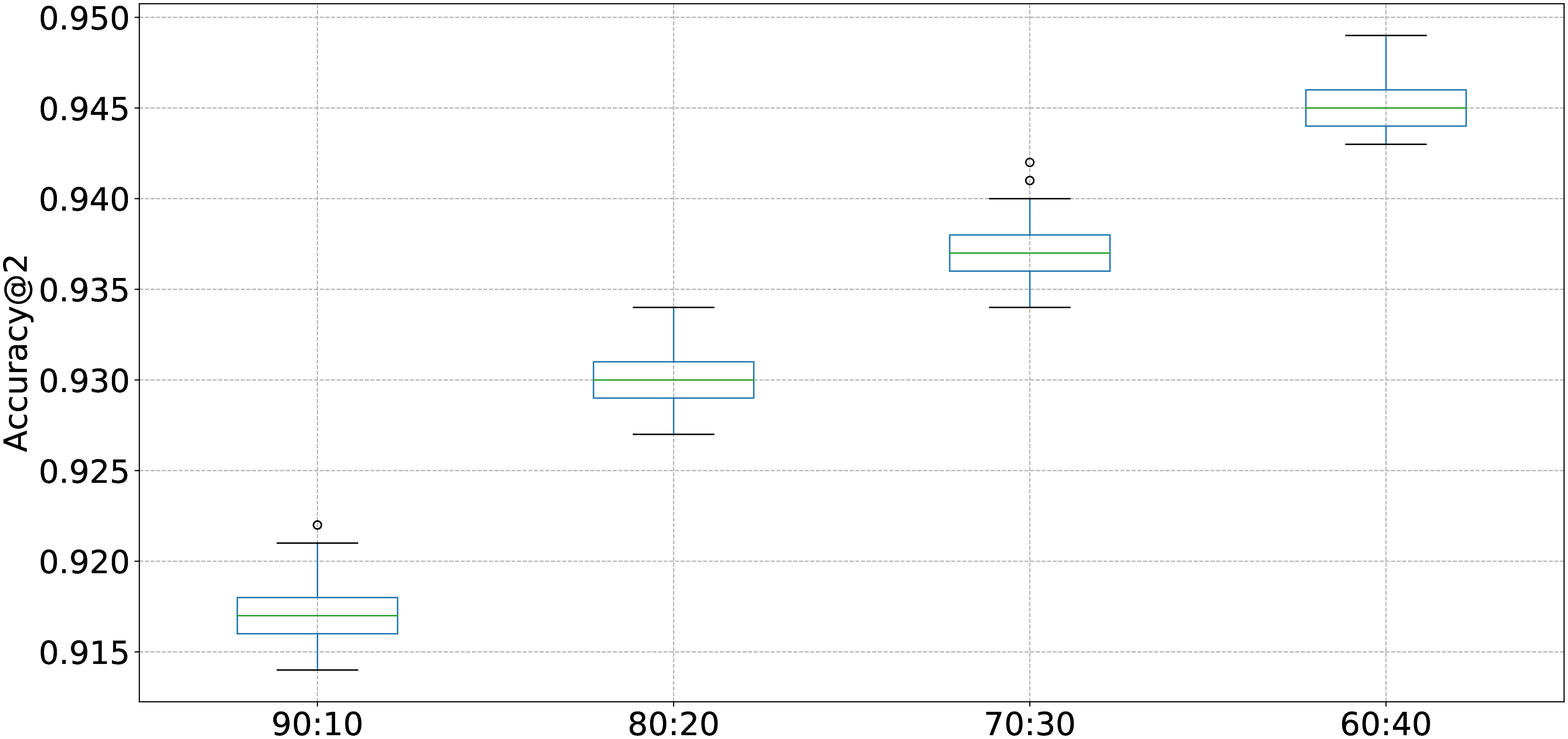}
		\end{minipage}}	
		
		\subfloat[Accuracy@3]{\begin{minipage}{9cm}
				\includegraphics[width=1\textwidth]{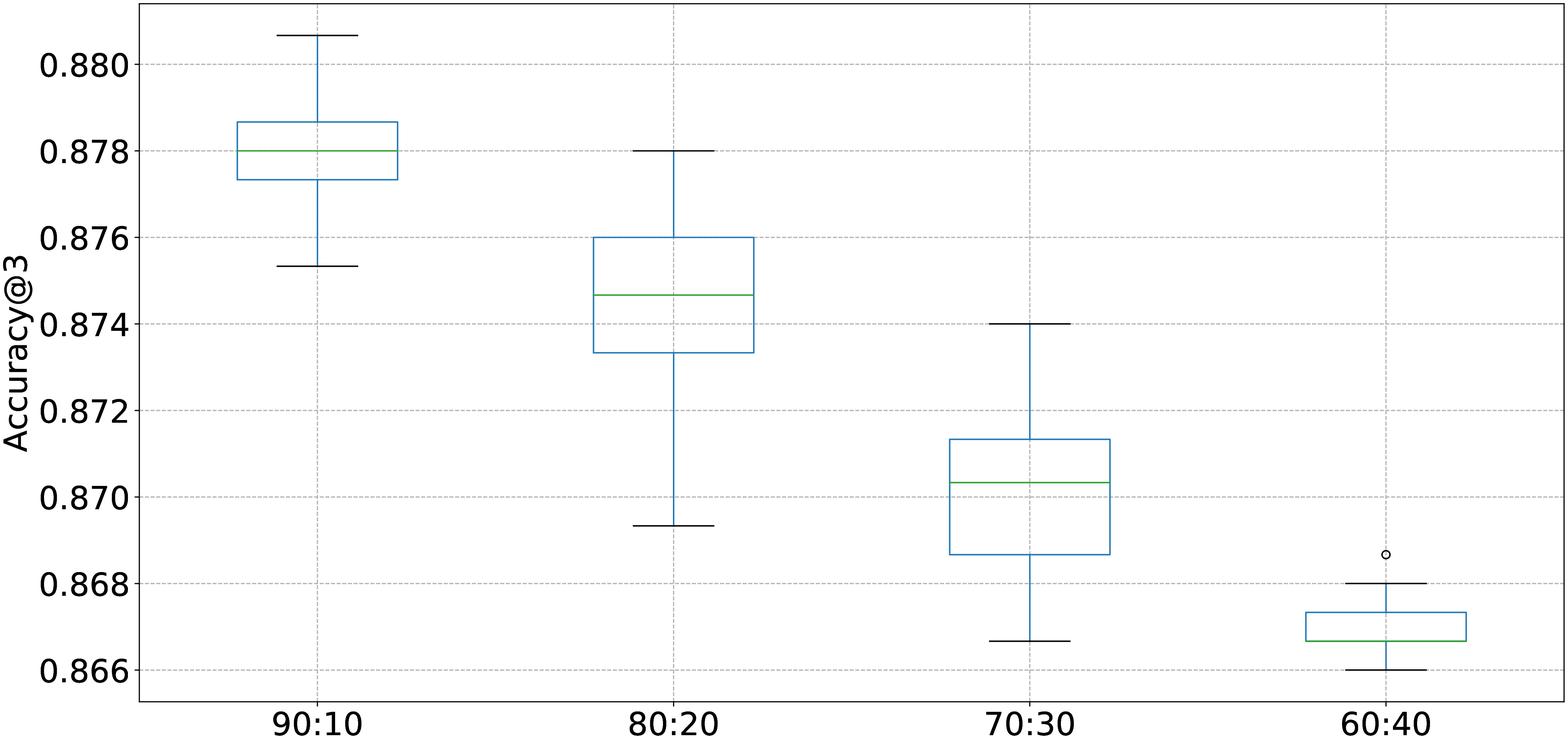}
		\end{minipage}}		
		\subfloat[Kendall's tau]{\begin{minipage}{9cm}
				\includegraphics[width=1\textwidth]{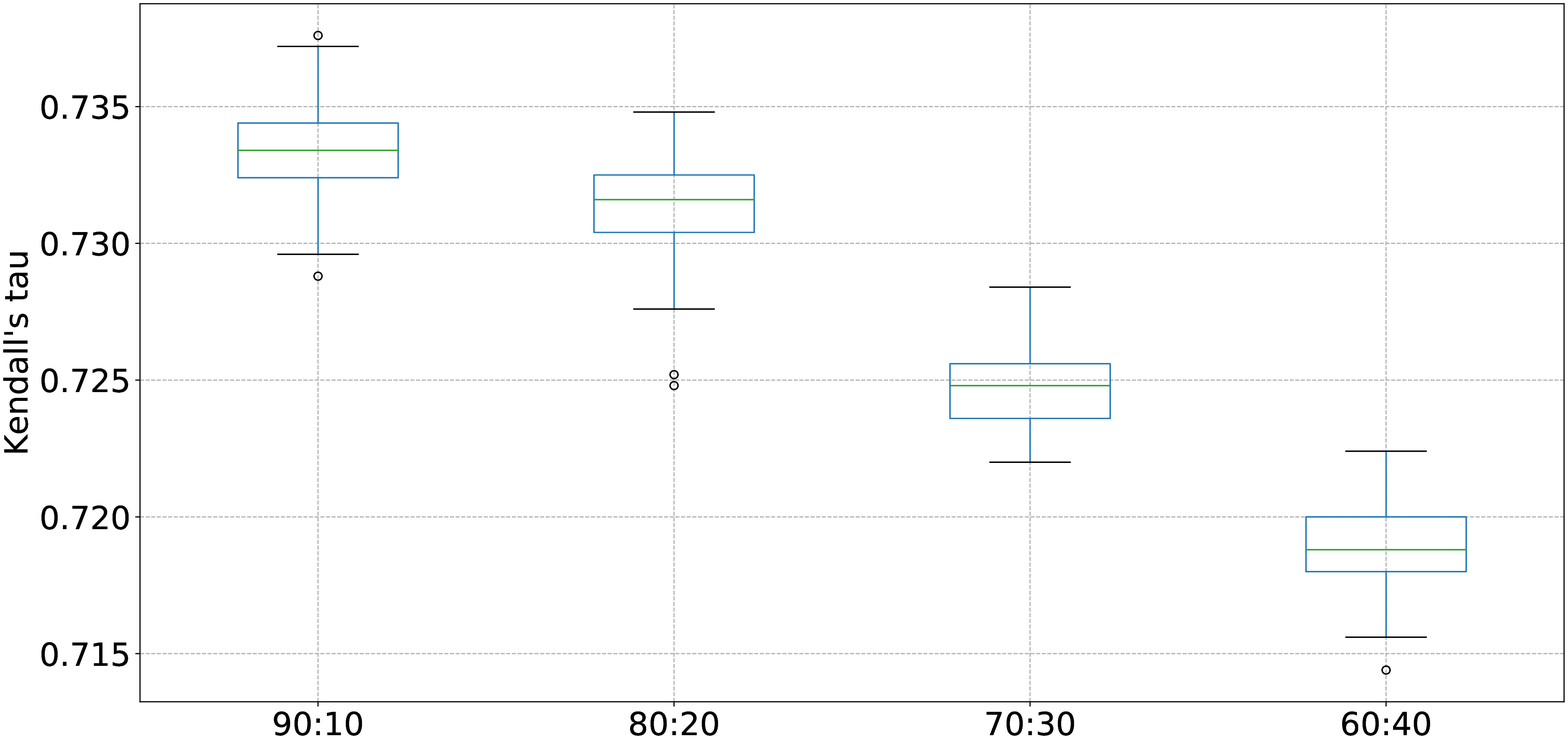}
		\end{minipage}}	
		\caption{\label{figure-3}Distribution of Top-$N$ accuracy and Kendall's tau of linear variant of proposed framework for valued decision examples.}
	\end{figure}

\begin{figure}[!htbp] 
		\centering
		\subfloat[Accuracy@1]{\begin{minipage}{9cm}
				\includegraphics[width=1\textwidth]{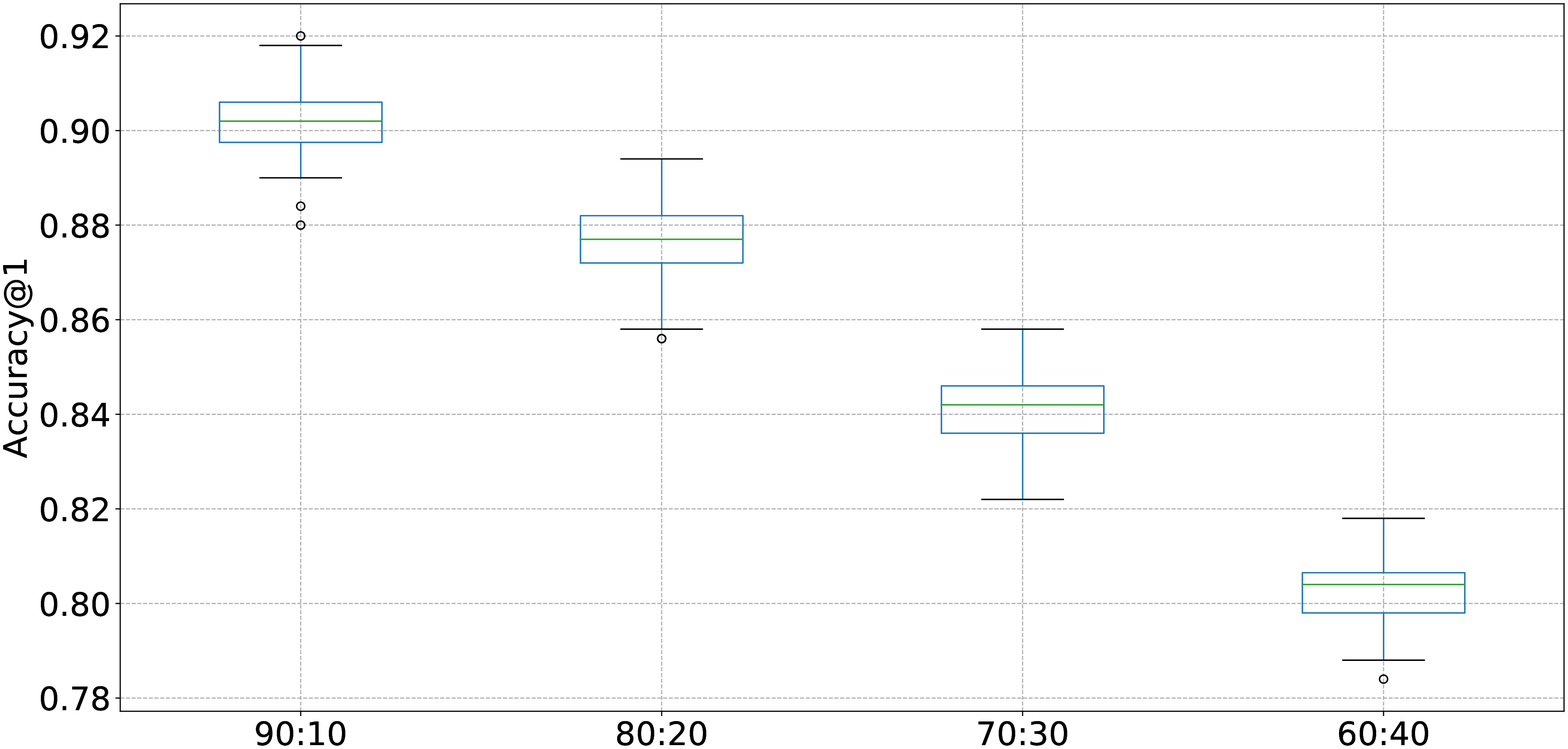}
		\end{minipage}}
		\subfloat[Accuracy@2]{\begin{minipage}{9cm}
				\includegraphics[width=1\textwidth]{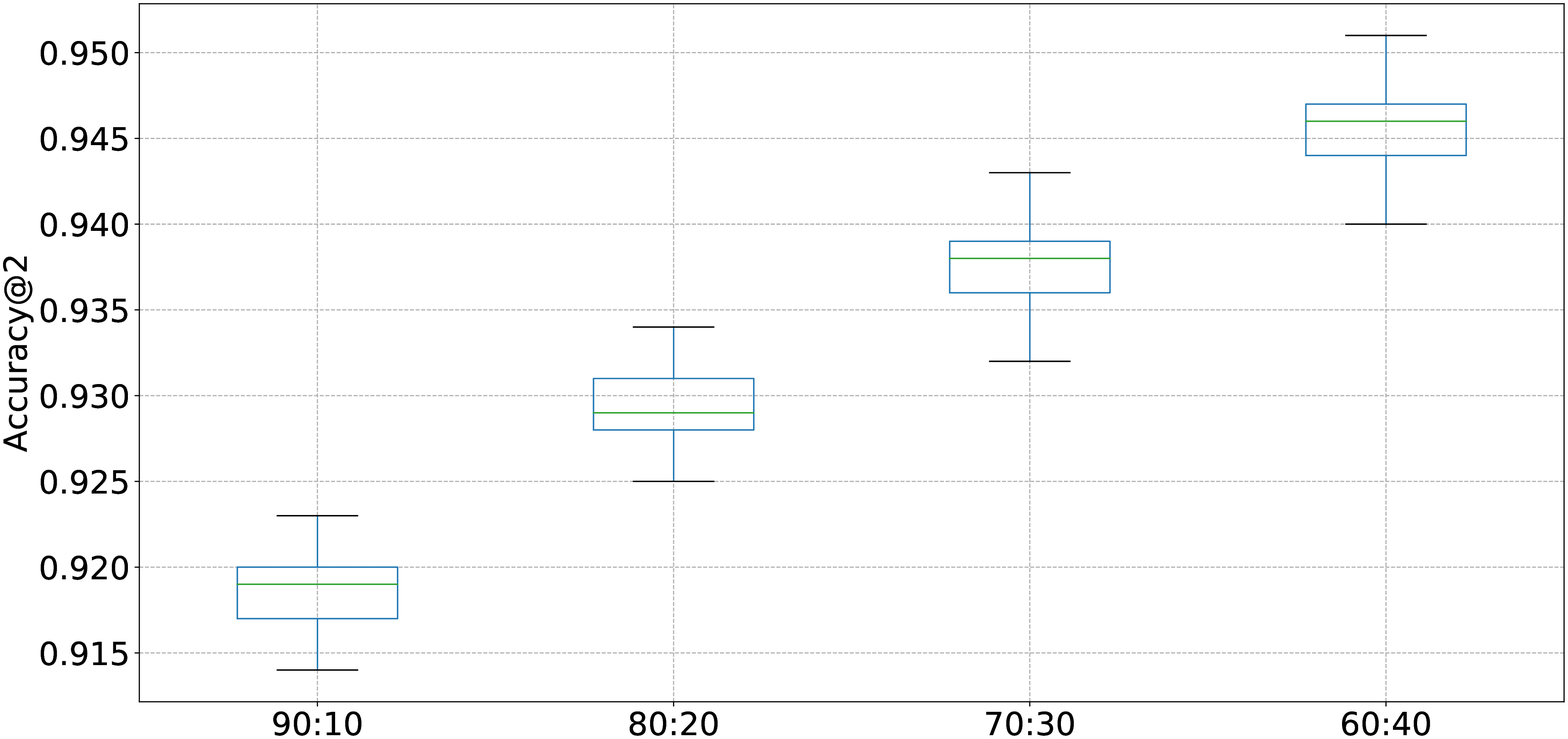}
		\end{minipage}}	
		
		\subfloat[Accuracy@3]{\begin{minipage}{9cm}
				\includegraphics[width=1\textwidth]{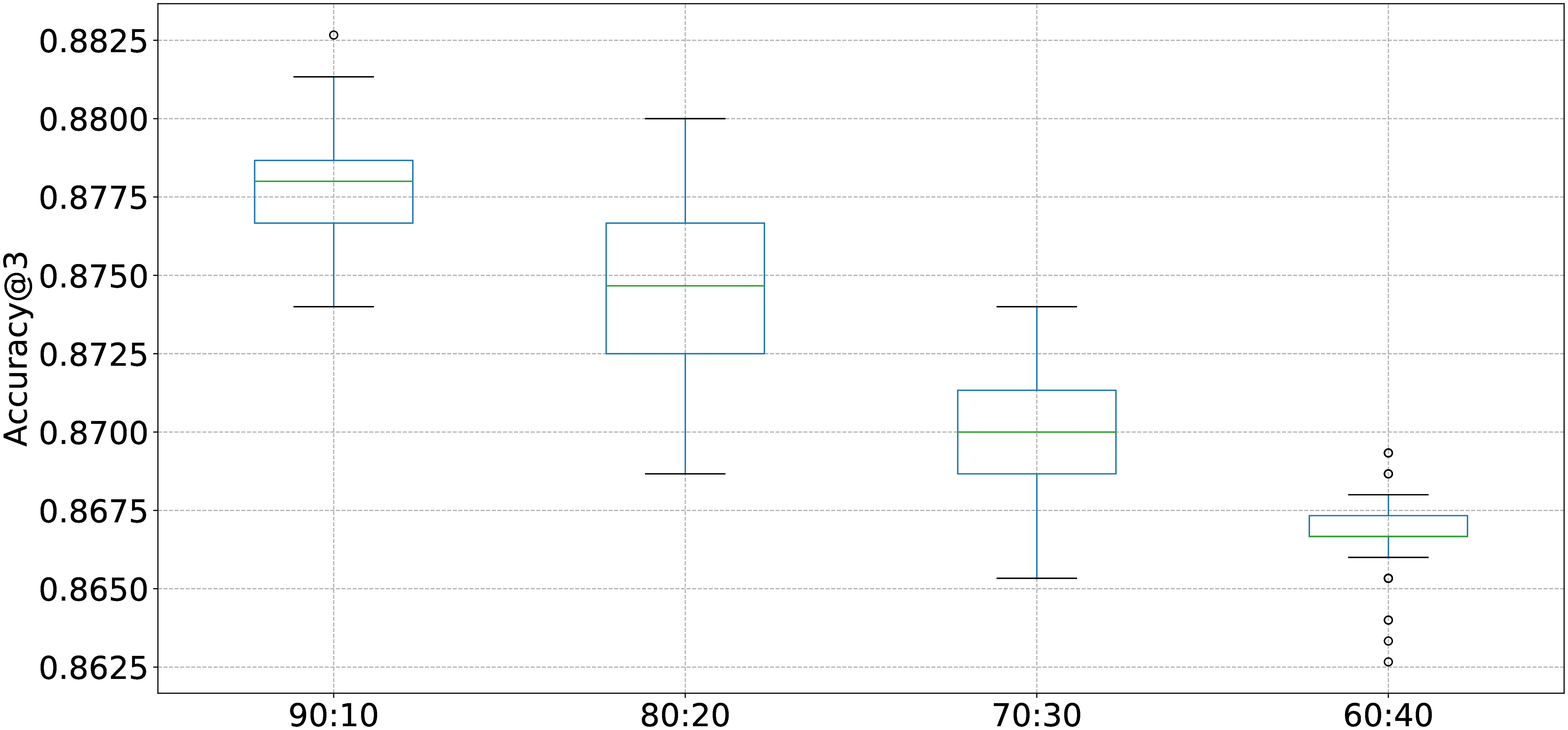}
		\end{minipage}}		
		\subfloat[Kendall's tau]{\begin{minipage}{9cm}
				\includegraphics[width=1\textwidth]{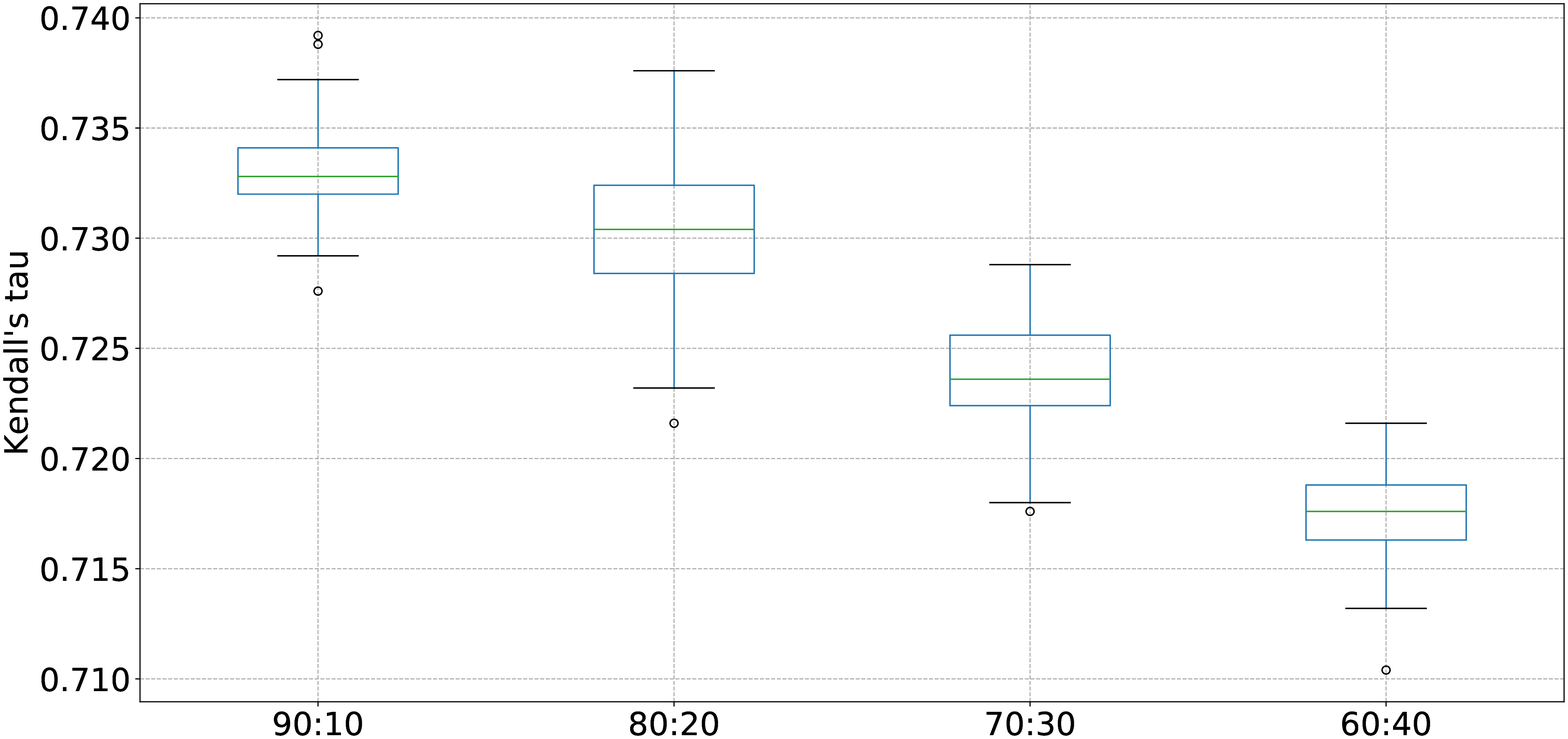}
		\end{minipage}}	
		\caption{\label{figure-4}Distribution of Top-$N$ accuracy and Kendall's tau of piecewise-linear variant of proposed framework for valued decision examples.}
	\end{figure}

\begin{figure}[!htbp] 
		\centering
		\subfloat[Accuracy@1]{\begin{minipage}{9cm}
				\includegraphics[width=1\textwidth]{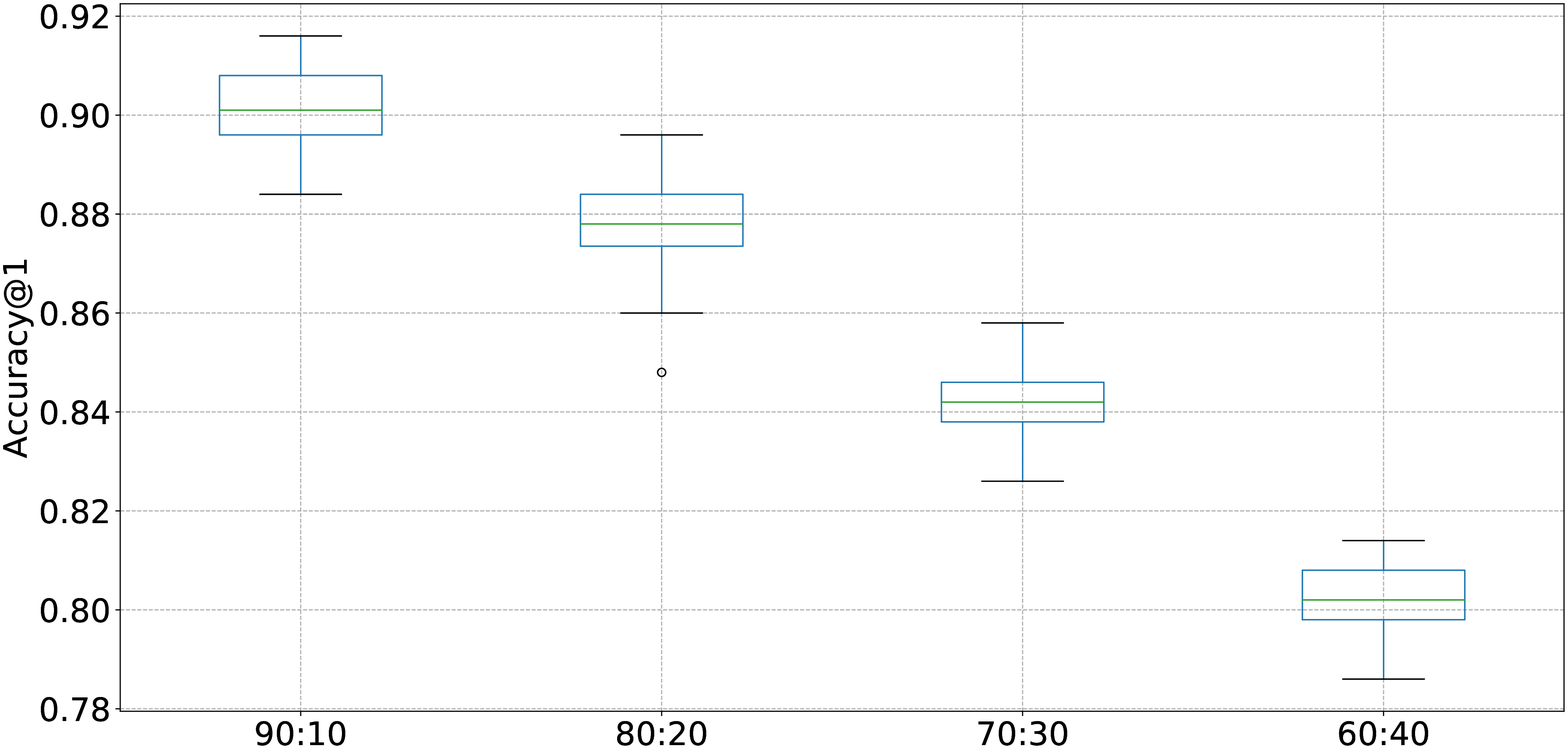}
		\end{minipage}}
		\subfloat[Accuracy@2]{\begin{minipage}{9cm}
				\includegraphics[width=1\textwidth]{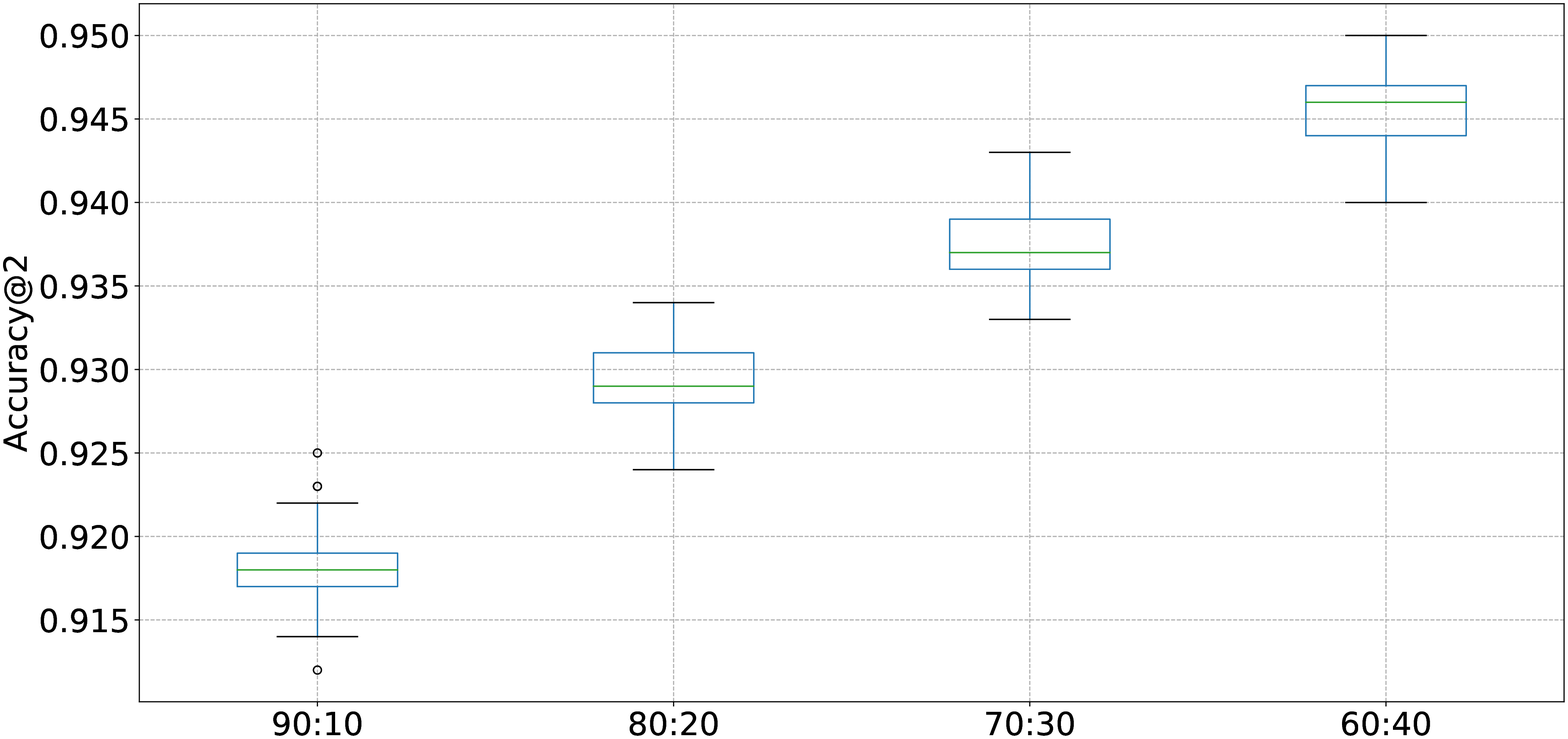}
		\end{minipage}}	
		
		\subfloat[Accuracy@3]{\begin{minipage}{9cm}
				\includegraphics[width=1\textwidth]{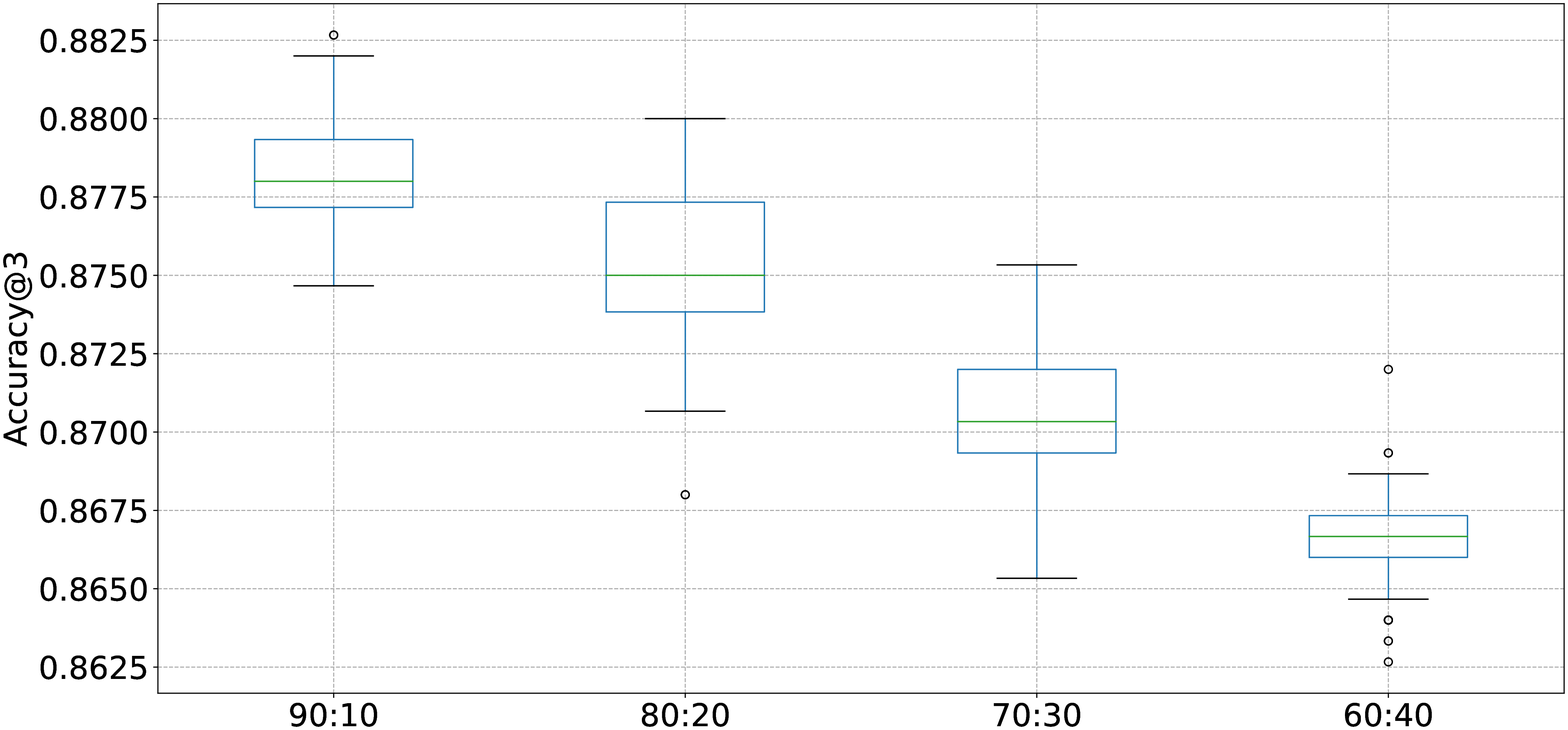}
		\end{minipage}}		
		\subfloat[Kendall's tau]{\begin{minipage}{9cm}
				\includegraphics[width=1\textwidth]{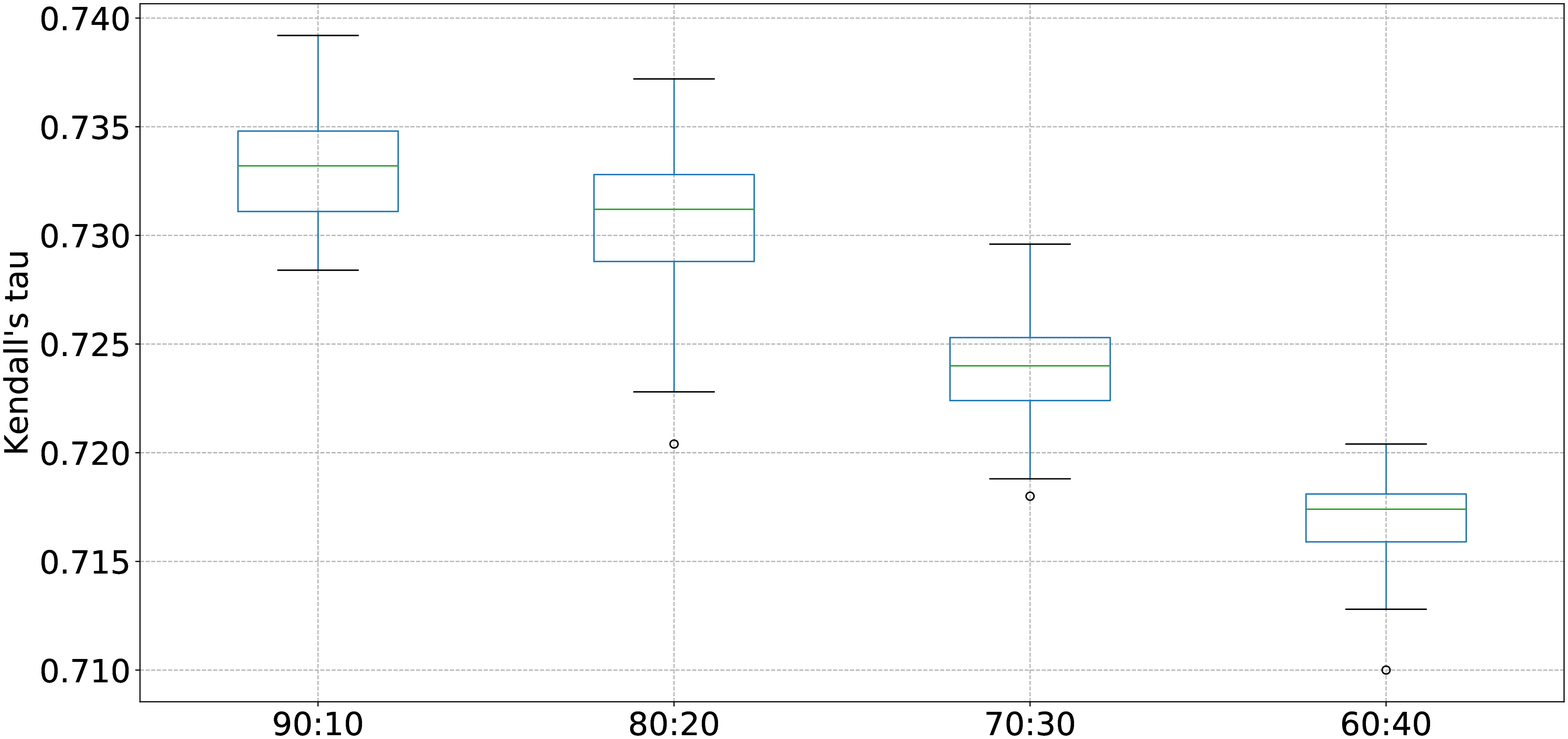}
		\end{minipage}}	
		\caption{\label{figure-5}Distribution of Top-$N$ accuracy and Kendall's tau of splined variant of proposed framework for valued decision examples.}
	\end{figure}

\begin{figure}[!htbp] 
		\centering
		\subfloat[Accuracy@1]{\begin{minipage}{9cm}
				\includegraphics[width=1\textwidth]{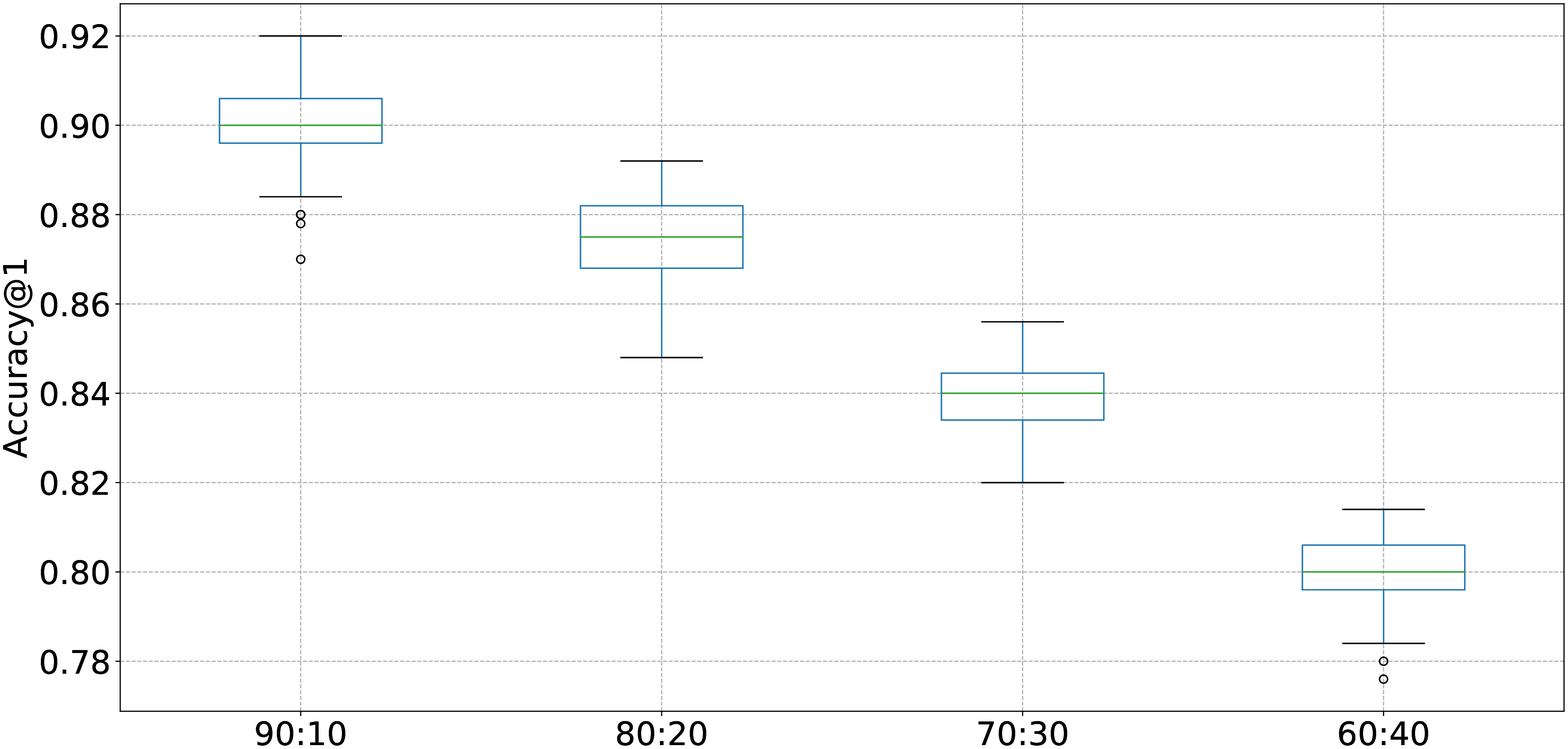}
		\end{minipage}}
		\subfloat[Accuracy@2]{\begin{minipage}{9cm}
				\includegraphics[width=1\textwidth]{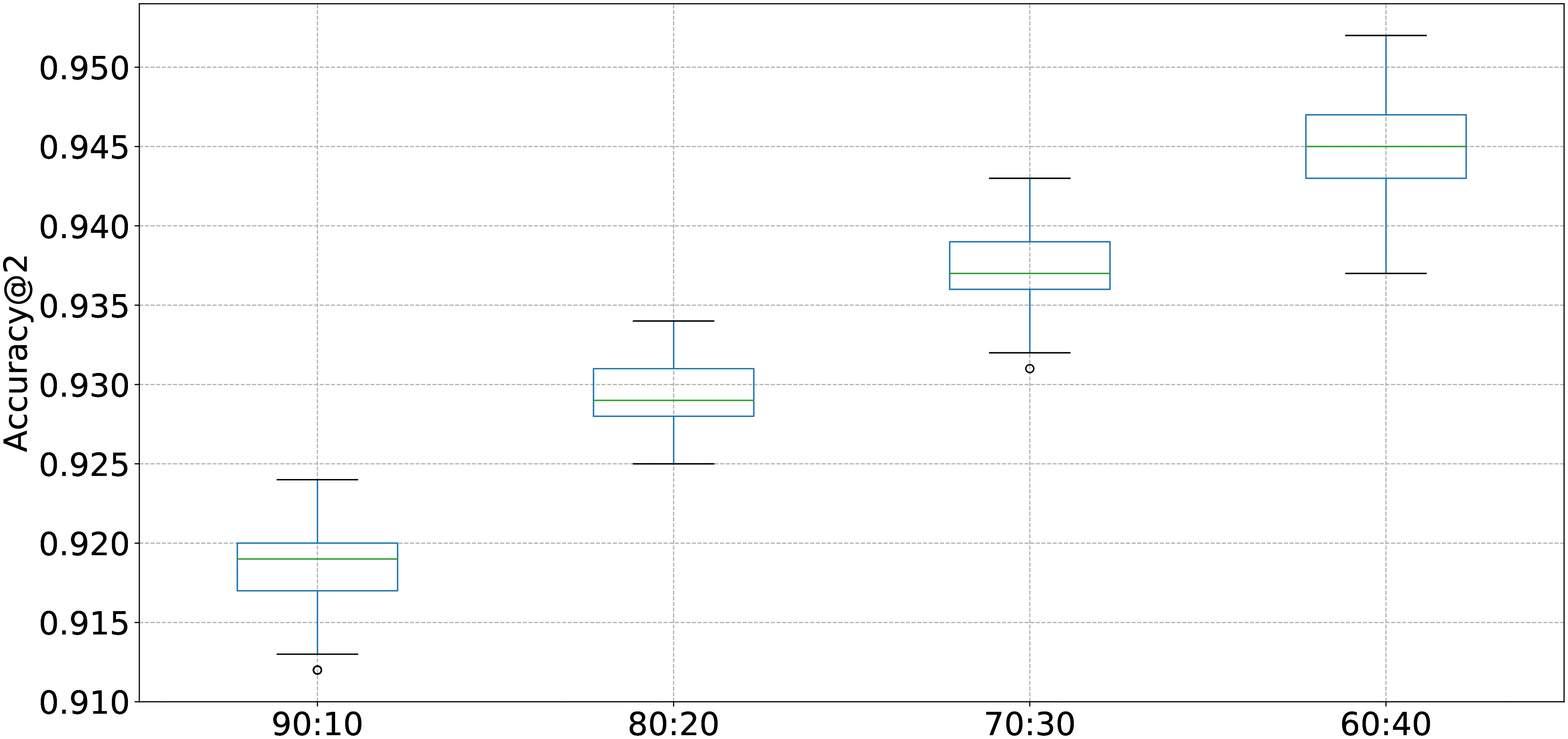}
		\end{minipage}}	
		
		\subfloat[Accuracy@3]{\begin{minipage}{9cm}
				\includegraphics[width=1\textwidth]{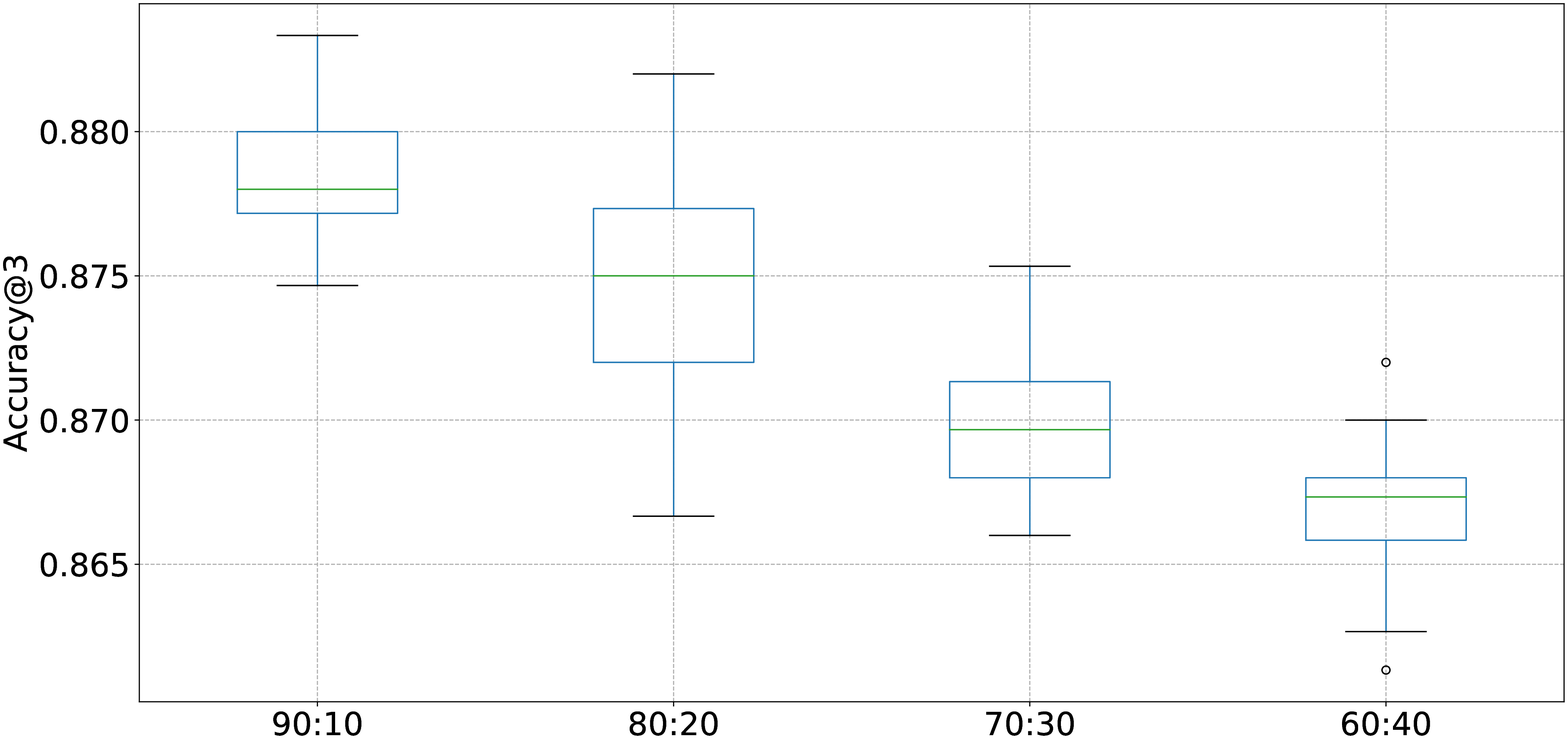}
		\end{minipage}}		
		\subfloat[Kendall's tau]{\begin{minipage}{9cm}
				\includegraphics[width=1\textwidth]{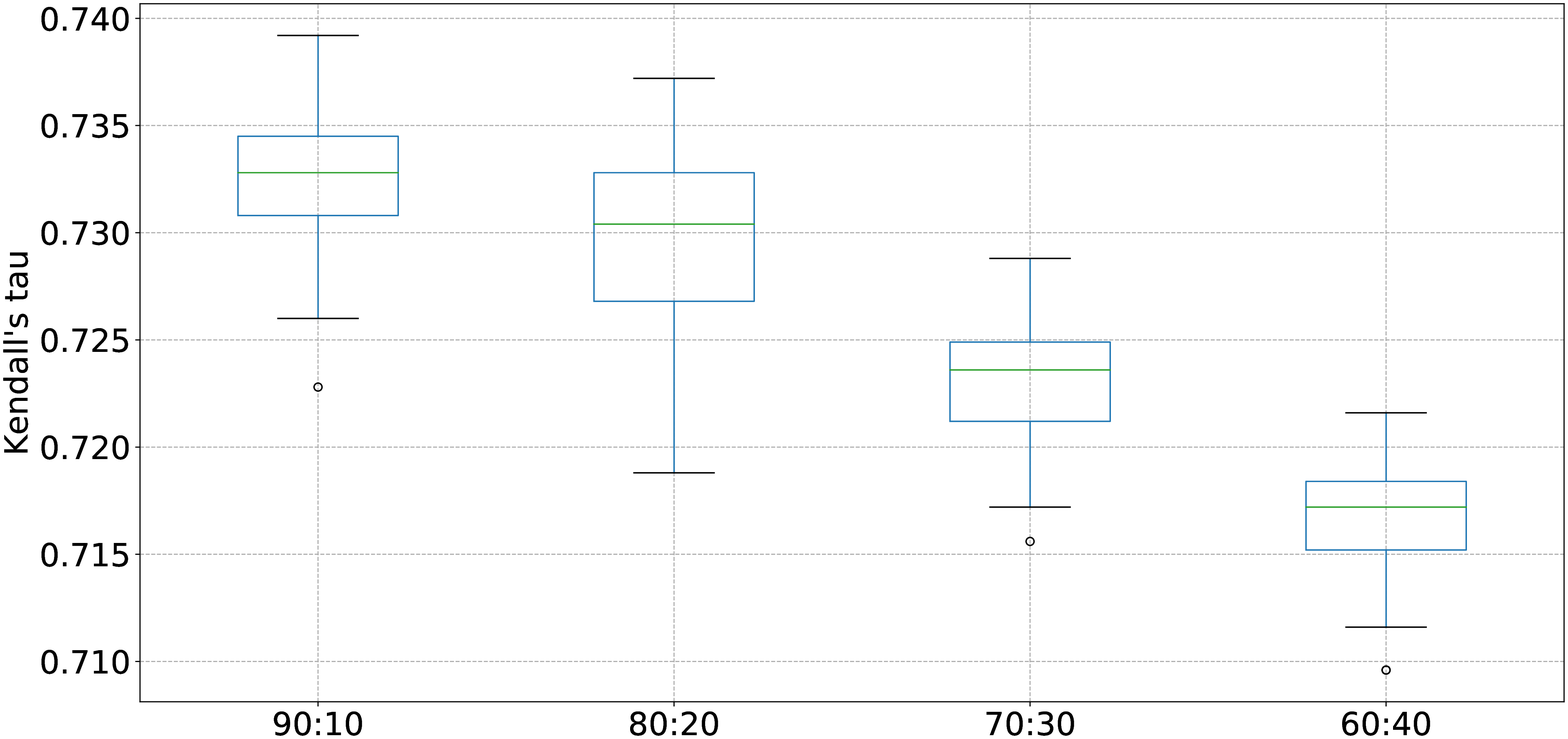}
		\end{minipage}}	
		\caption{\label{figure-6}Distribution of Top-$N$ accuracy and Kendall's tau of general monotone variant of proposed framework for valued decision examples.}
	\end{figure}

	The results are summarized in Table \ref{table-2} and depicted in Figures~\ref{figure-3} -- \ref{figure-6}. We observe that the Accuracy@1 measures of the four variants of the proposed framework decrease with the majority proportion of credibility degree that is assigned to the actual assignment. For example, the mean of Accuracy@1 of the linear variant decreases from 0.9258 for the crisp decision examples (see Table \ref{table-1}) to 0.8101 for the ``60\%-40\%'' valued decision examples. This is due to that, for a pair of reference alternatives $a$ and $a'$, where the actual assignments of $a$ and $a'$ are $Cl_{s+1}$ and $Cl_s$, $s=1,...,4$, in the case of crisp decision examples, it is very credible to maximize $U\left(a\right) - U\left(a'\right)$ since ${D_ \succ }\left( {a,a'} \right) = 1$; when the majority proportion of credibility degree that is assigned to the actual assignment decreases, ${D_ \succ }\left( {a,a'} \right)$ decreases while ${D_ \prec }\left( {a,a'} \right)$ and ${D_ = }\left( {a,a'} \right)$ increase, and in turn the credibility to maximize $U\left(a\right) - U\left(a'\right)$ decrease, which contradicts the actual preference relation between $a$ and $a'$. Moreover, it is interesting to observe that the Accuracy@2 measures of the four variants increase as the majority proportion of credibility degree that is assigned to the actual assignment decreases. Such an observation reflects that, although the decision examples are incredible when the majority proportion of credibility degree that is assigned to the actual assignment decreases, the four variants can work out top two recommendations that are credible enough for making correct prediction. In addition, the Accuracy@3 and Kendall's tau indicators decrease with the decline of the majority proportion of credibility degree that is assigned to the actual assignment. On the other hand, in the comparison of the four variants, the linear variant slightly outperforms others in terms of the Top-$N$ accuracy and Kendall's tau, since the actual value function model is linear, which makes the linear variant achieve better performance than the others.
	
	\begin{table}[!htbp] \caption{\label{table-2}Top-$N$ accuracy and Kendall's tau in terms of mean and standard deviation of four variants of proposed framework for valued decision examples.}
		\centering
		\scriptsize
		\begin{tabular}{rccccc}
			\hline
			Method & Credibility distribution & Accuracy@1 & Accuracy@2 & Accuracy@3 & Kendall's tau \\
			\hline
			\multirow{4}[0]{*}{Linear variant} & 90\%-10\% & 0.9058$\pm$0.0052 & 0.9172$\pm$0.0015 & 0.8778$\pm$0.0010 & 0.7333$\pm$0.0016 \\
			& 80\%-20\% & 0.8805$\pm$0.0051 & 0.9300$\pm$0.0014 & 0.8745$\pm$0.0017 & 0.7313$\pm$0.0018 \\
			& 70\%-30\% & 0.8452$\pm$0.0043 & 0.9374$\pm$0.0016 & 0.8702$\pm$0.0015 & 0.7247$\pm$0.0013 \\
			& 60\%-40\% & 0.8101$\pm$0.0047 & 0.9452$\pm$0.0013 & 0.8669$\pm$0.0006 & 0.7188$\pm$0.0014 \\
			\multicolumn{6}{l}{} \\
			\multirow{4}[0]{*}{Piecewise-linear variant} & 90\%-10\% & 0.9016$\pm$0.0071 & 0.9186$\pm$0.0021 & 0.8780$\pm$0.0015 & 0.7330$\pm$0.0021 \\
			& 80\%-20\% & 0.8769$\pm$0.0079 & 0.9293$\pm$0.0018 & 0.8746$\pm$0.0026 & 0.7303$\pm$0.0030 \\
			& 70\%-30\% & 0.8413$\pm$0.0074 & 0.9374$\pm$0.0022 & 0.8700$\pm$0.0019 & 0.7238$\pm$0.0024 \\
			& 60\%-40\% & 0.8027$\pm$0.0064 & 0.9455$\pm$0.0022 & 0.8669$\pm$0.0011 & 0.7173$\pm$0.0018 \\
			\multicolumn{6}{l}{} \\
			\multirow{4}[0]{*}{Splined variant} & 90\%-10\% & 0.9010$\pm$0.0079 & 0.9183$\pm$0.0021 & 0.8782$\pm$0.0017 & 0.7329$\pm$0.0026 \\
			& 80\%-20\% & 0.8772$\pm$0.0075 & 0.9293$\pm$0.0020 & 0.8750$\pm$0.0025 & 0.7307$\pm$0.0028 \\
			& 70\%-30\% & 0.8422$\pm$0.0066 & 0.9371$\pm$0.0022 & 0.8705$\pm$0.0020 & 0.7241$\pm$0.0023 \\
			& 60\%-40\% & 0.8024$\pm$0.0064 & 0.9454$\pm$0.0020 & 0.8666$\pm$0.0014 & 0.7170$\pm$0.0019 \\
			\multicolumn{6}{l}{} \\
			\multirow{4}[0]{*}{General monotone variant} & 90\%-10\% & 0.9000$\pm$0.0088 & 0.9186$\pm$0.0026 & 0.8782$\pm$0.0018 & 0.7327$\pm$0.0028 \\
			& 80\%-20\% & 0.8741$\pm$0.0102 & 0.9293$\pm$0.0020 & 0.8747$\pm$0.0031 & 0.7297$\pm$0.0040 \\
			& 70\%-30\% & 0.8398$\pm$0.0076 & 0.9373$\pm$0.0024 & 0.8698$\pm$0.0020 & 0.7233$\pm$0.0026 \\
			& 60\%-40\% & 0.8004$\pm$0.0069 & 0.9452$\pm$0.0025 & 0.8667$\pm$0.0018 & 0.7166$\pm$0.0024 \\
			\hline
		\end{tabular}
	\end{table}
	
	\subsection{Simulating decision policies with non-linear value functions}
	\noindent In the above experimental analysis, the actual preference model underlying the given decision examples is composed of linear marginal value functions. To test our methods on decision examples which are generated by other types of value functions, we conduct an additional experiment on simulating actual value functions to generate decision examples, on which the four variants of the proposed framework are applied to derive respective results for comparison. In this part, we assume actual value functions are in the general monotone form, that is, we make no assumptions about properties of actual value functions except monotonicity. The least assumption imposed on actual value functions makes the conclusion derived from the experimental outcomes would be general. A~widely used method for generating general monotone value functions is to assign marginal values to distinct performance values by sorting random values drawn from a uniform distribution \cite{kadzinski2013robust}. In a~recent study by \cite{dias2019generating}, this method for generating general monotone value functions is found to bias the obtained value functions towards a certain shape, especially when the distribution of performance values is not uniform. Recall that we observe the imbalanced distribution of performance values in Figure \ref{fig-1}. Hence, to avoid distorted conclusions, we use another method for generating general monotone value functions: (a) first, for each criterion, assign zero and a positive random value to the worst two performance values $x_j^0, x_j^1$, respectively, as the corresponding marginal values $u_j(x_j^0)$ and $u_j(x_j^1)$; (b) then, for the remaining performance values $x_j^2,...,x_j^{m_j}$, sequentially assign a random value $\varsigma$ satisfying the following equation to $x_j^k$, $k=2,...,m_j$, as the corresponding marginal value ${u_j}\left( {x_j^k} \right)$:
	\begin{equation*}
	\frac{{\varsigma  - {u_j}\left( {x_j^{k - 1}} \right)}}
	{{x_j^k - x_j^{k - 1}}} = \left( {1 + \delta } \right)\frac{{{u_j}\left( {x_j^{k - 1}} \right) - {u_j}\left( {x_j^{k - 2}} \right)}}
	{{x_j^{k - 1} - x_j^{k - 2}}},
	\end{equation*}
	where $\delta$ is a random value drawn uniformly from the interval $\left[ { - \rho ,\rho } \right]$, and $\rho > 0$ is a specified parameter used to control the complexity of the general monotone value function. Actually, $\left[ {1 - \rho ,1 + \rho } \right]$ delimits the variation range of growth rates of marginal values over consecutive sub-intervals. In this study, we consider the levels 25\%, 50\%, 75\%, and 100\% for $\rho$, and obviously a greater $\rho$ allows a wide variation range of growth rates of marginal values over consecutive sub-intervals, which is more likely to generate complex marginal value functions. (c) normalize the marginal values on all criteria. For each level of $\rho$, we randomly generate 100 general monotone value functions and the corresponding assignment for each alternative. We transform the crisp assignment for each alternative to the valued one by allocating 20\% credibility degree to the classes adjacent to its actual assignment. Then, each generated dataset is randomly split into the training set $A^R$ and the test set $A^T$, and the four variants of the proposed framework are applied to construct a preference model from $A^R$ and then tested on $A^T$.
	
\begin{figure}[!htbp] 
		\centering
		\subfloat[Accuracy@1]{\begin{minipage}{9cm}
				\includegraphics[width=1\textwidth]{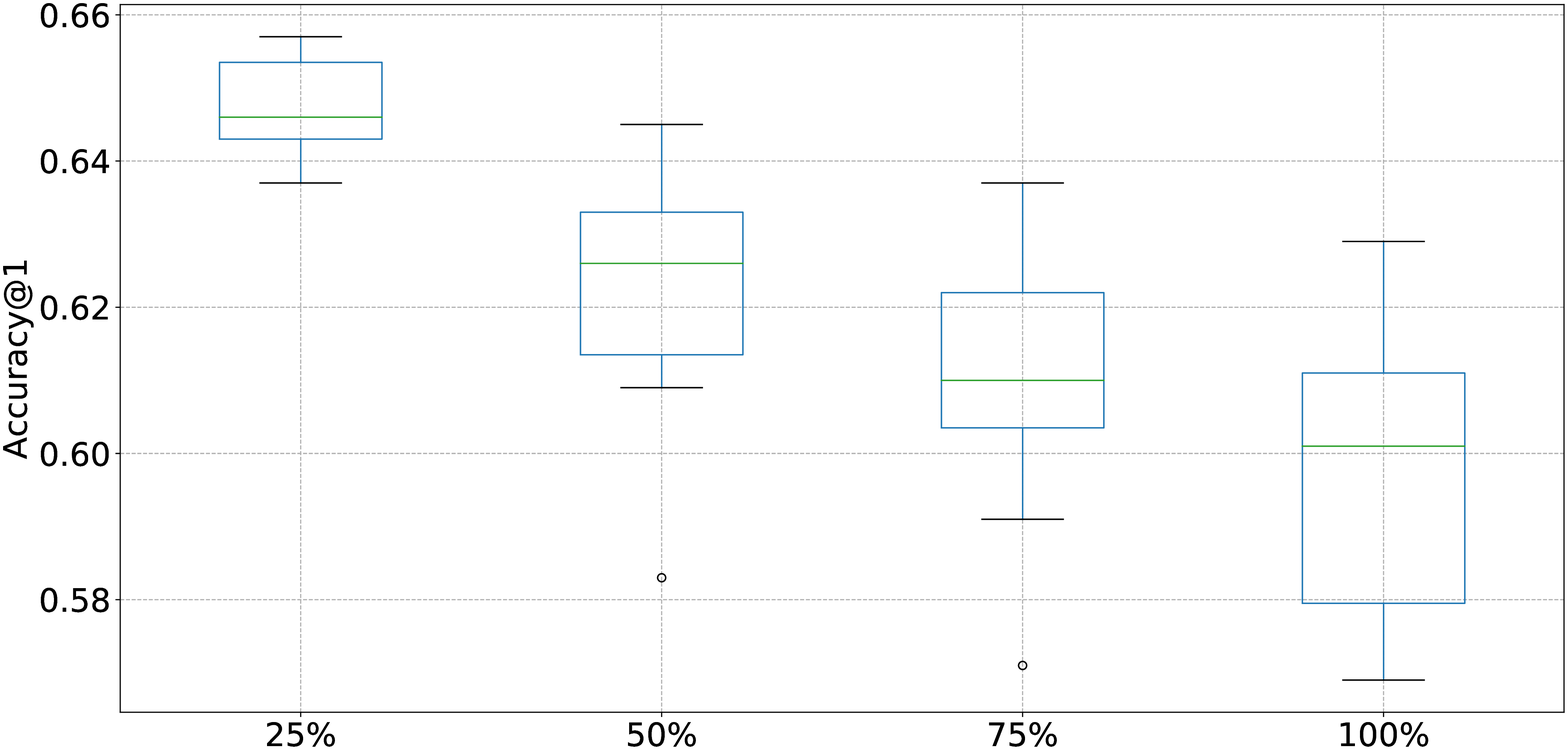}
		\end{minipage}}
		\subfloat[Accuracy@2]{\begin{minipage}{9cm}
				\includegraphics[width=1\textwidth]{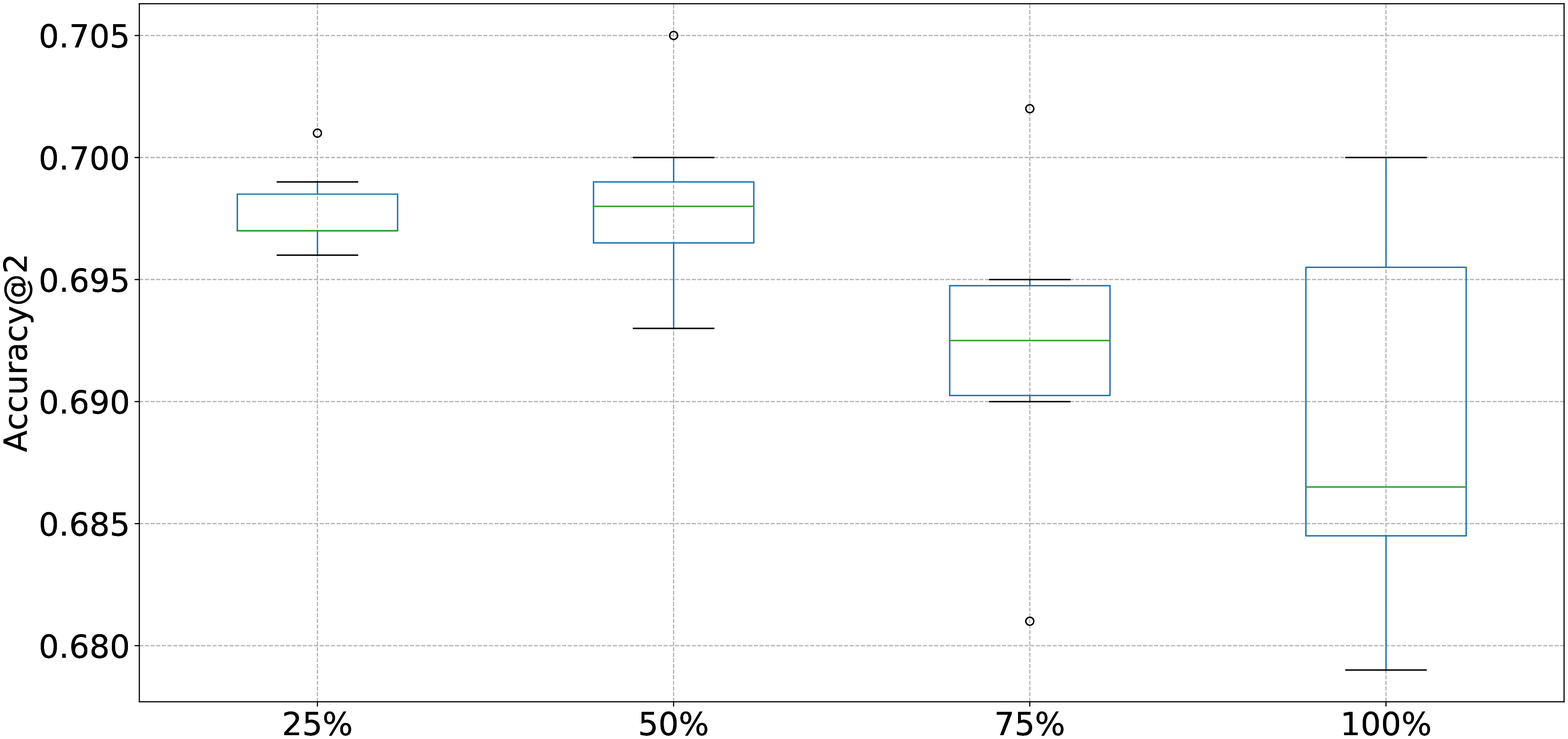}
		\end{minipage}}	
		
		\subfloat[Accuracy@3]{\begin{minipage}{9cm}
				\includegraphics[width=1\textwidth]{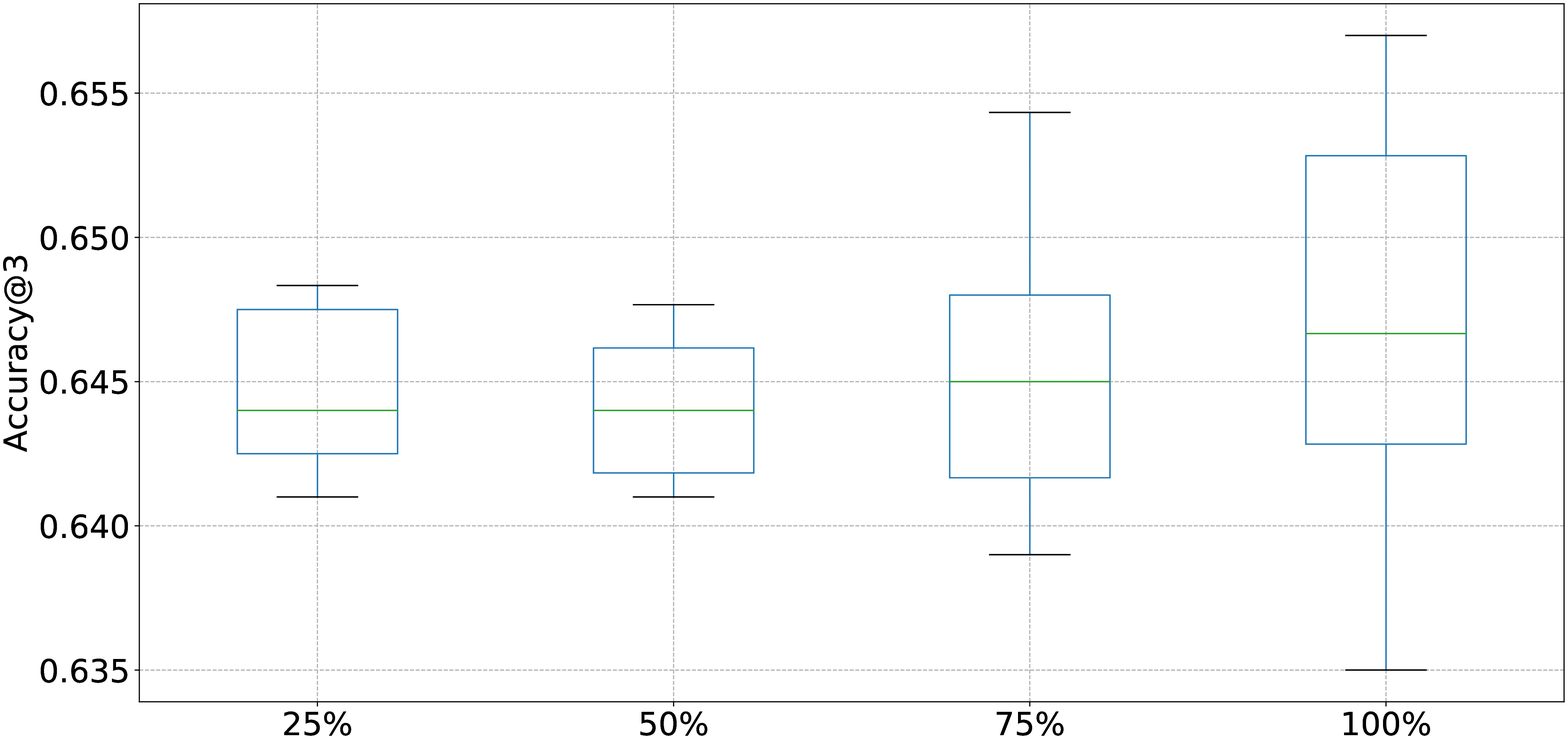}
		\end{minipage}}		
		\subfloat[Kendall's tau]{\begin{minipage}{9cm}
				\includegraphics[width=1\textwidth]{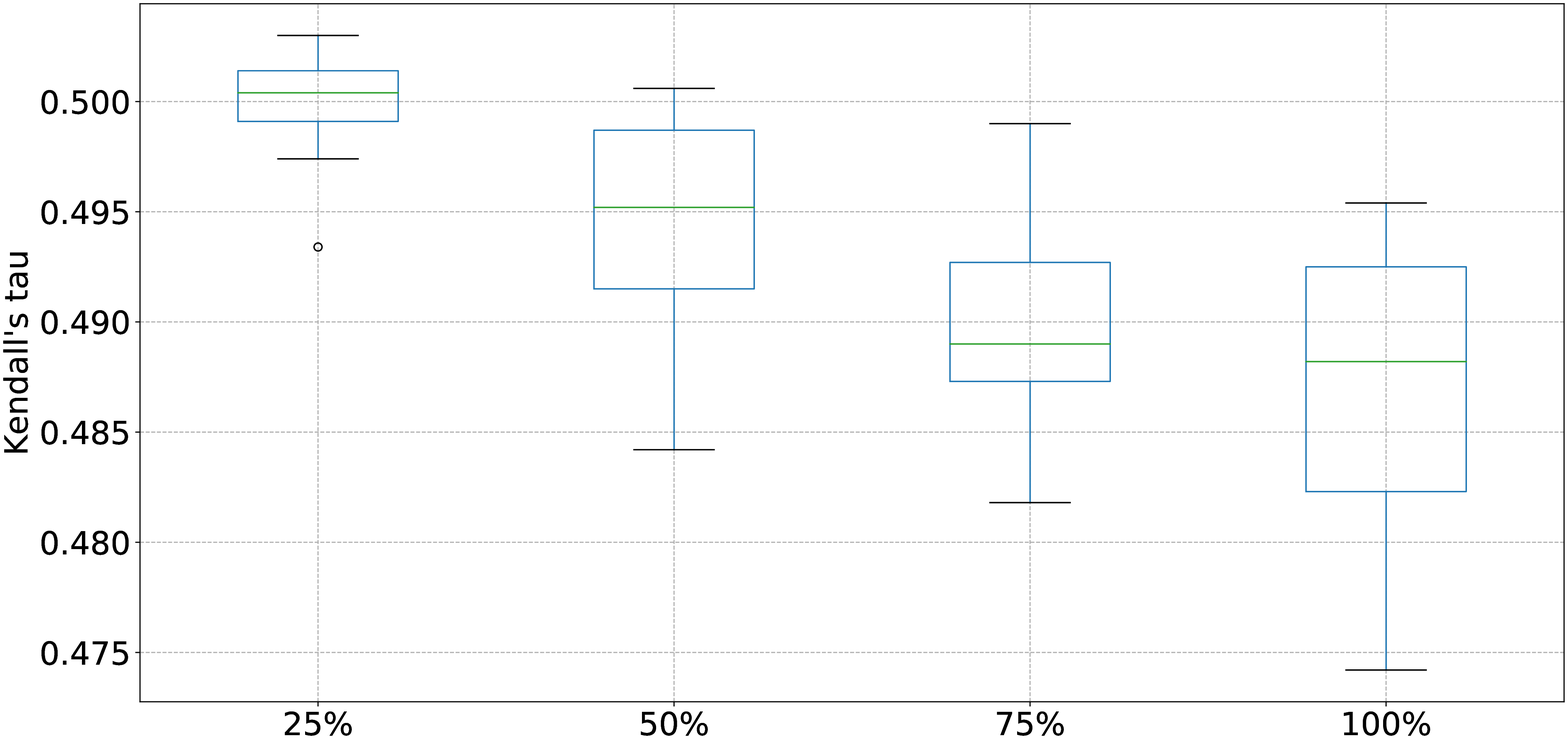}
		\end{minipage}}	
		\caption{\label{figure-7}Distribution of Top-$N$ accuracy and Kendall's tau of linear variant of proposed framework for valued decision examples generated by random value functions.}
	\end{figure}

\begin{figure}[!htbp] 
		\centering
		\subfloat[Accuracy@1]{\begin{minipage}{9cm}
				\includegraphics[width=1\textwidth]{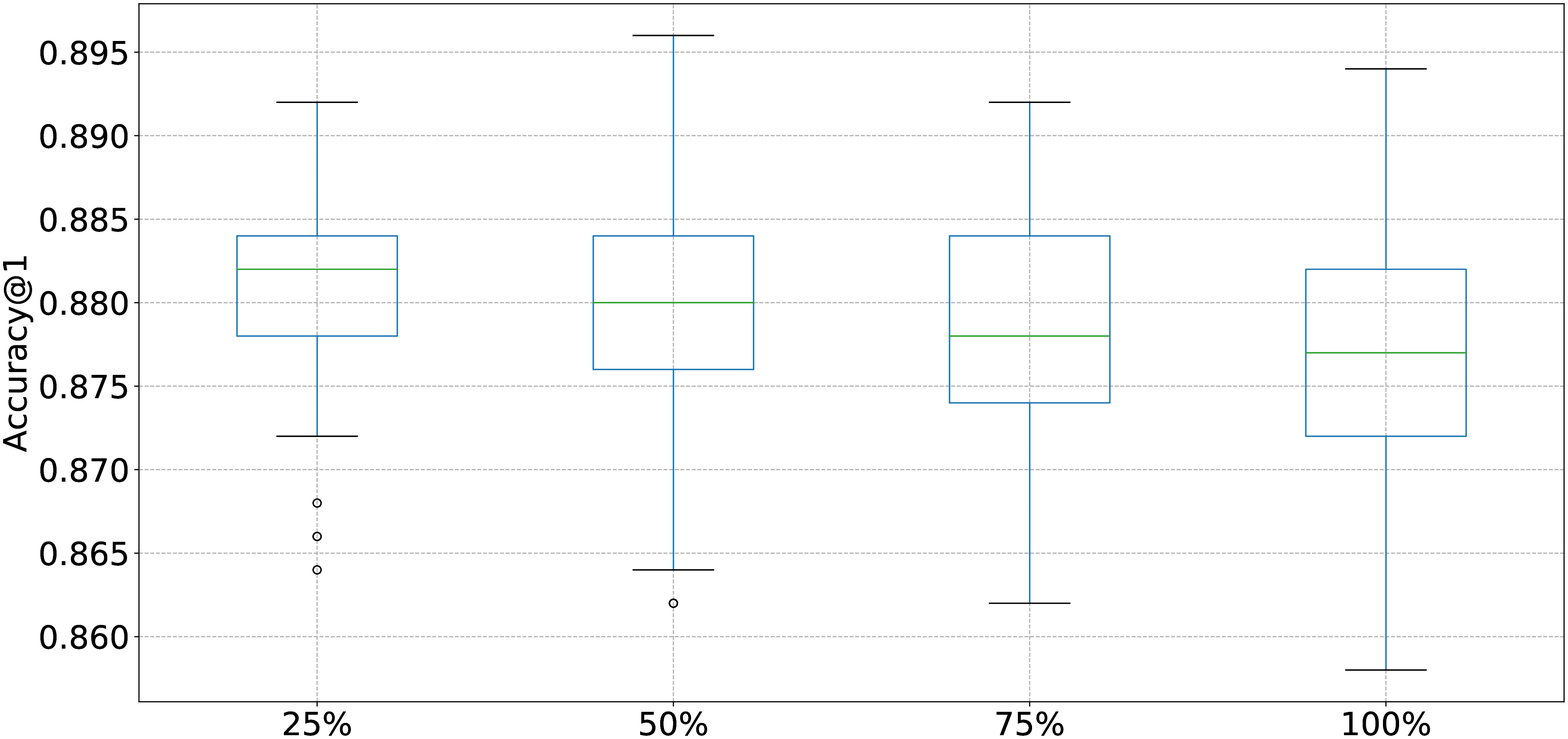}
		\end{minipage}}
		\subfloat[Accuracy@2]{\begin{minipage}{9cm}
				\includegraphics[width=1\textwidth]{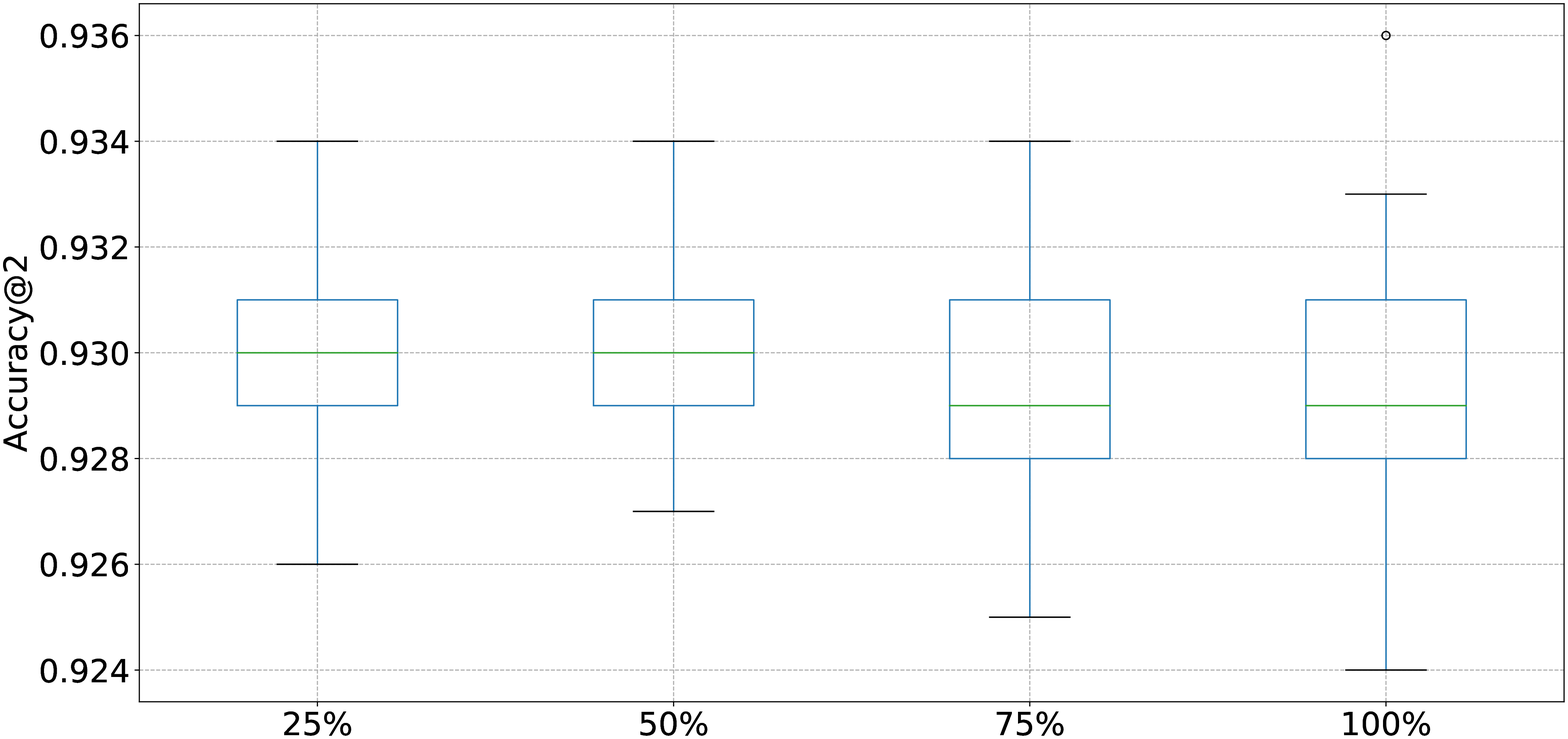}
		\end{minipage}}	
		
		\subfloat[Accuracy@3]{\begin{minipage}{9cm}
				\includegraphics[width=1\textwidth]{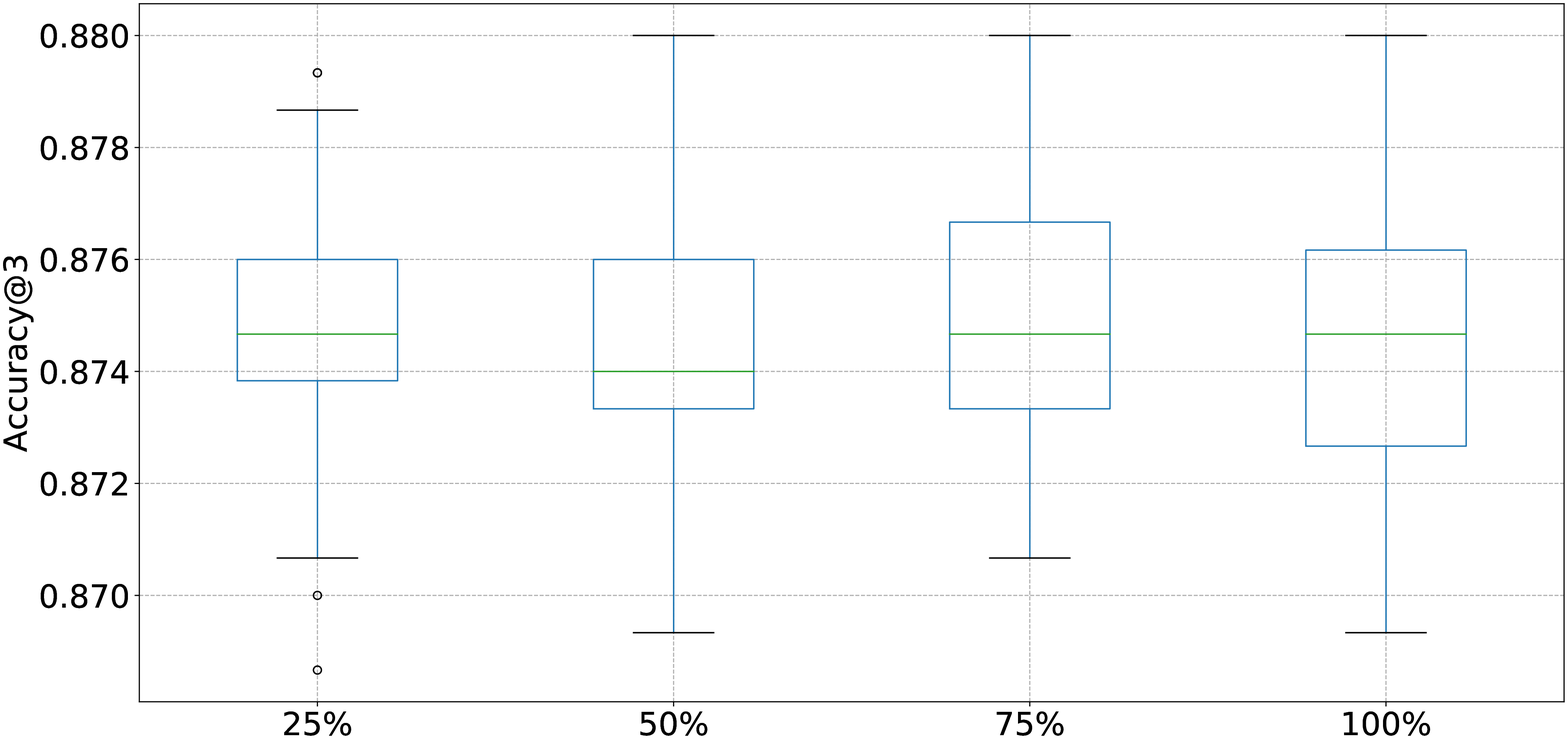}
		\end{minipage}}		
		\subfloat[Kendall's tau]{\begin{minipage}{9cm}
				\includegraphics[width=1\textwidth]{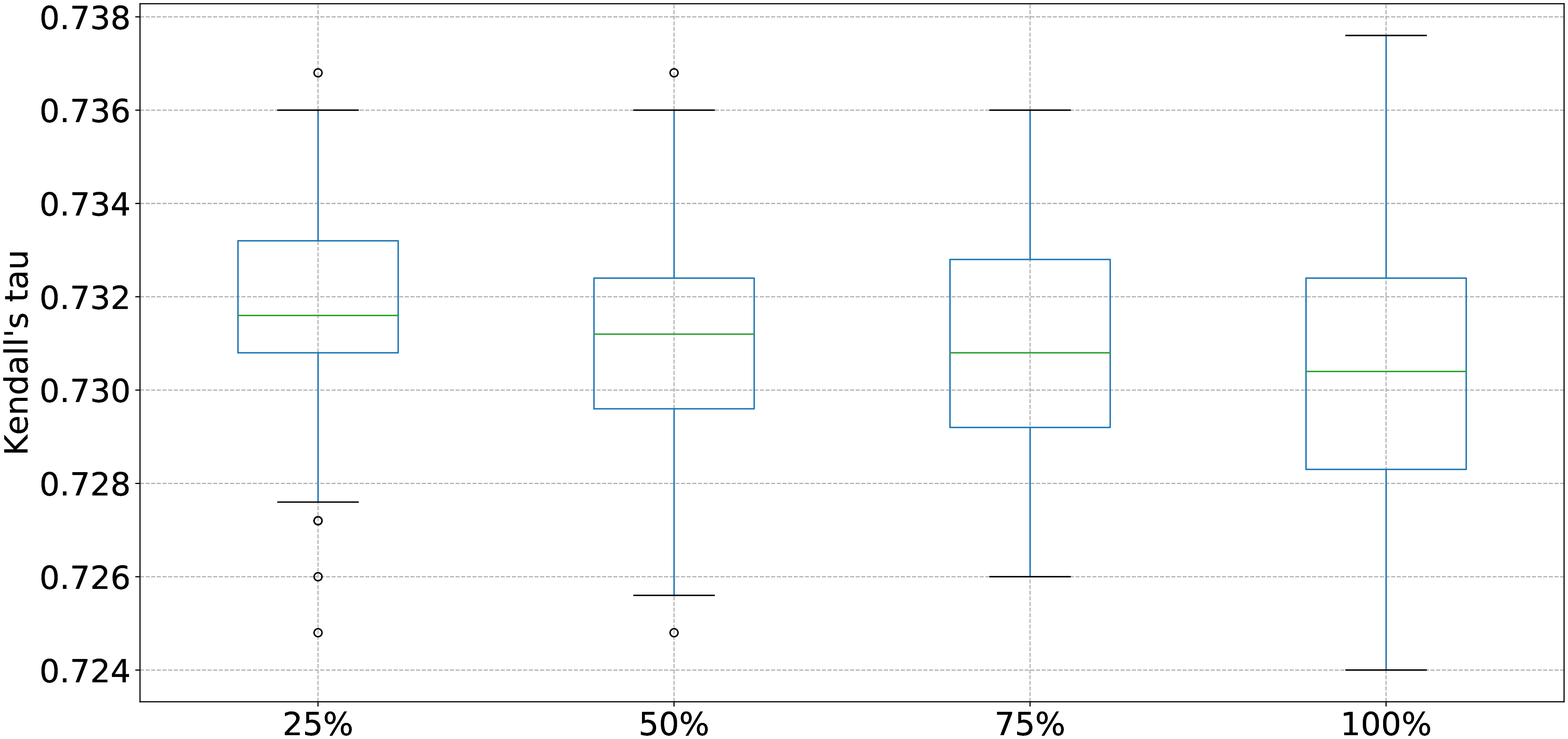}
		\end{minipage}}	
		\caption{\label{figure-8}Distribution of Top-$N$ accuracy and Kendall's tau of piecewise-linear variant of proposed framework for valued decision examples generated by random value functions.}
	\end{figure}

\begin{figure}[!htbp] 
		\centering
		\subfloat[Accuracy@1]{\begin{minipage}{9cm}
				\includegraphics[width=1\textwidth]{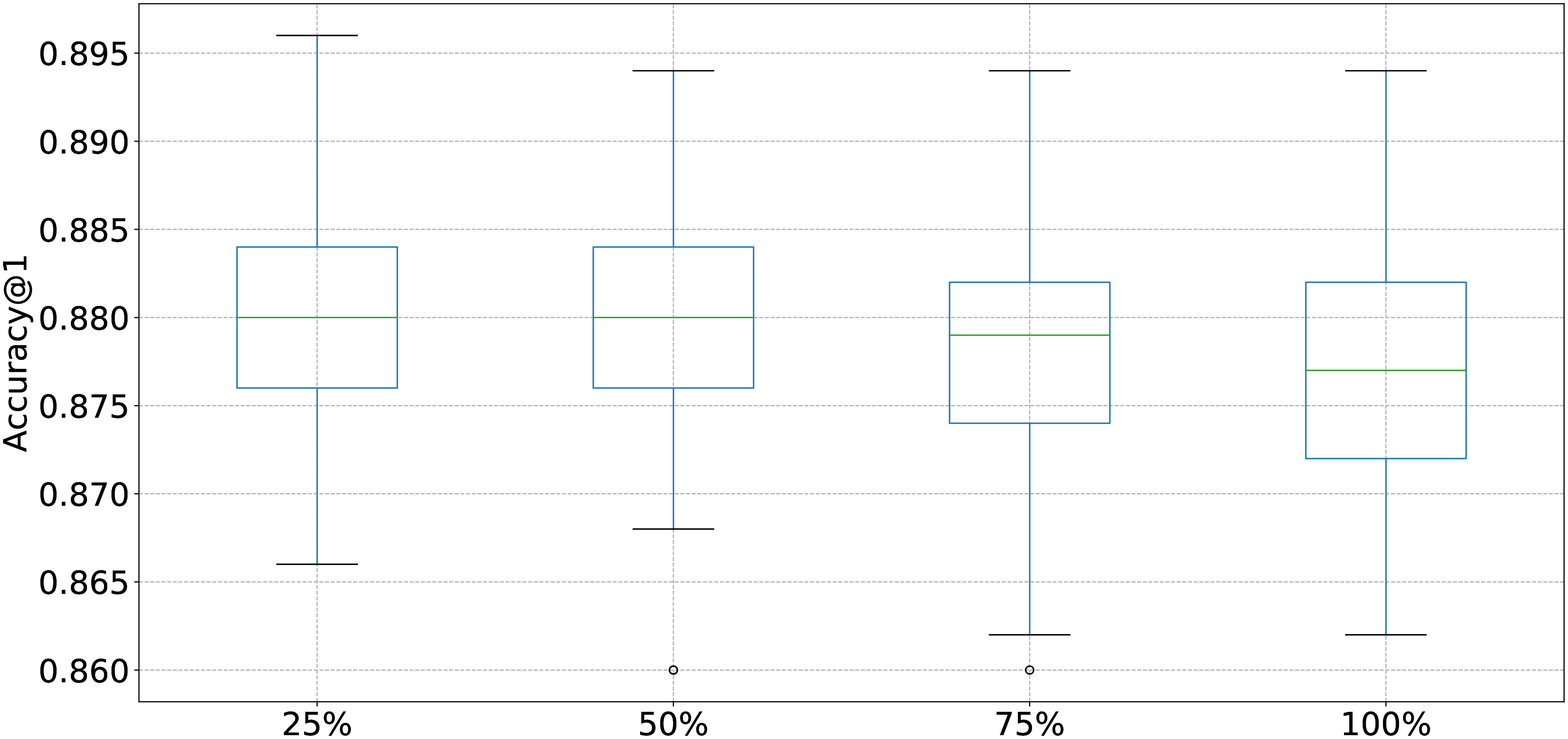}
		\end{minipage}}
		\subfloat[Accuracy@2]{\begin{minipage}{9cm}
				\includegraphics[width=1\textwidth]{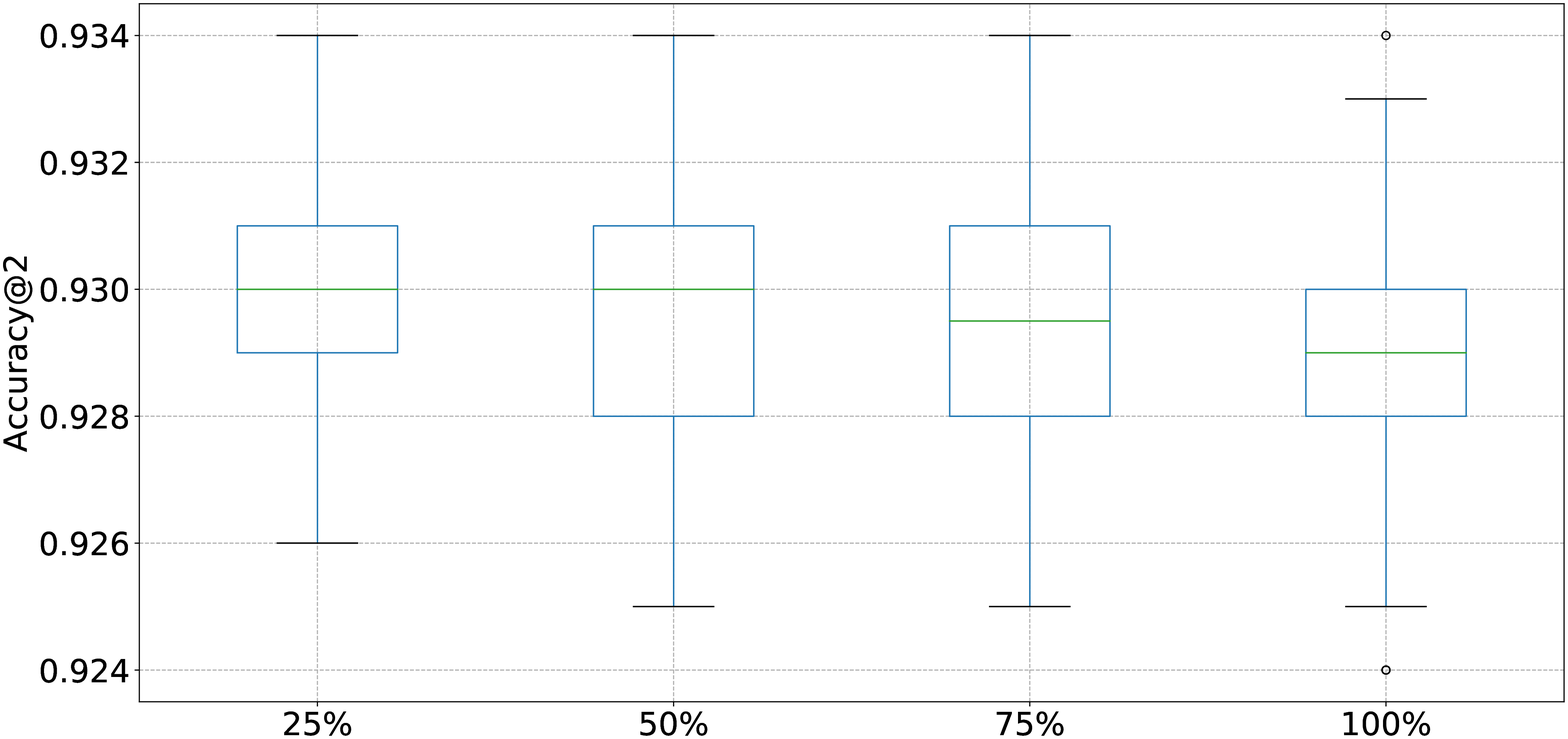}
		\end{minipage}}	
		
		\subfloat[Accuracy@3]{\begin{minipage}{9cm}
				\includegraphics[width=1\textwidth]{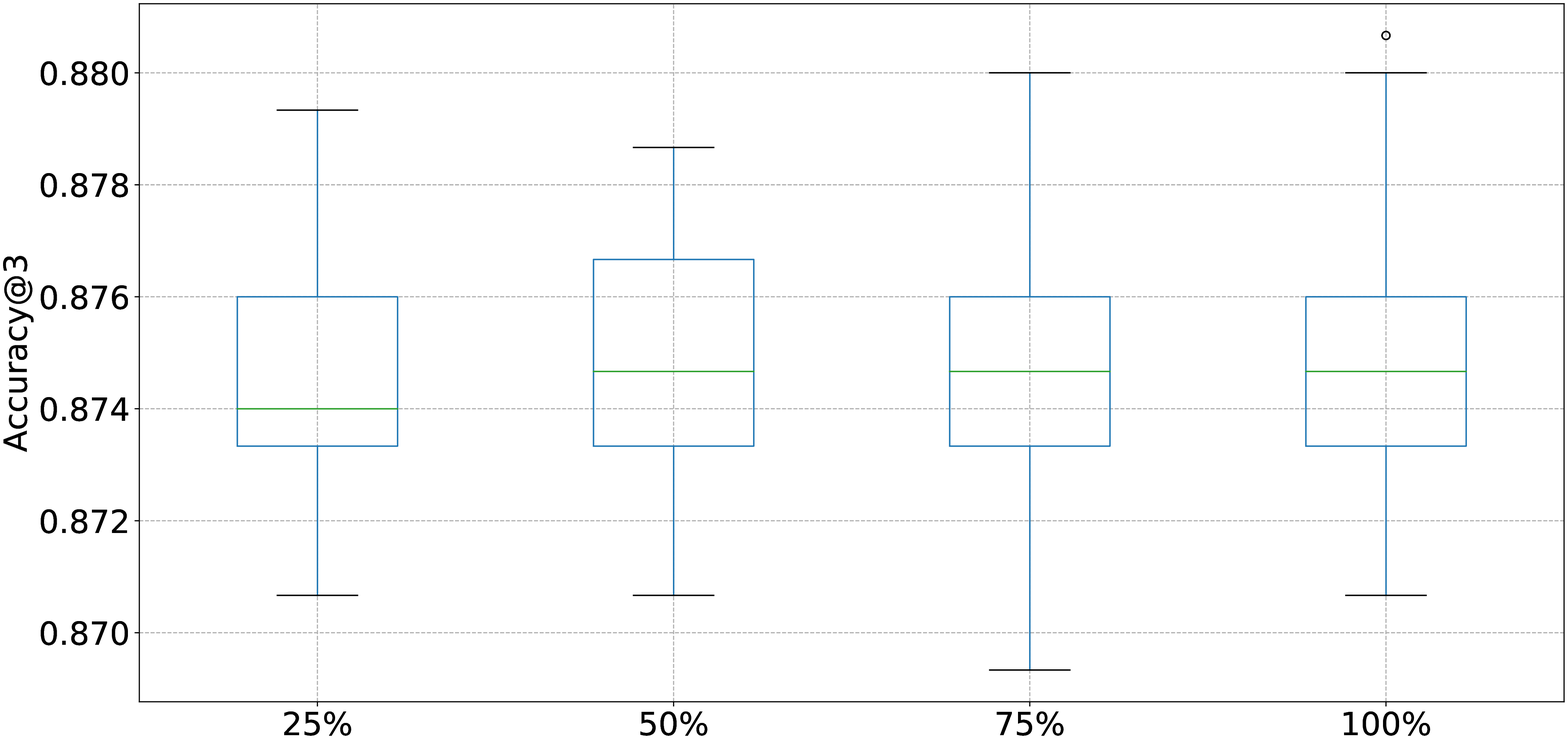}
		\end{minipage}}		
		\subfloat[Kendall's tau]{\begin{minipage}{9cm}
				\includegraphics[width=1\textwidth]{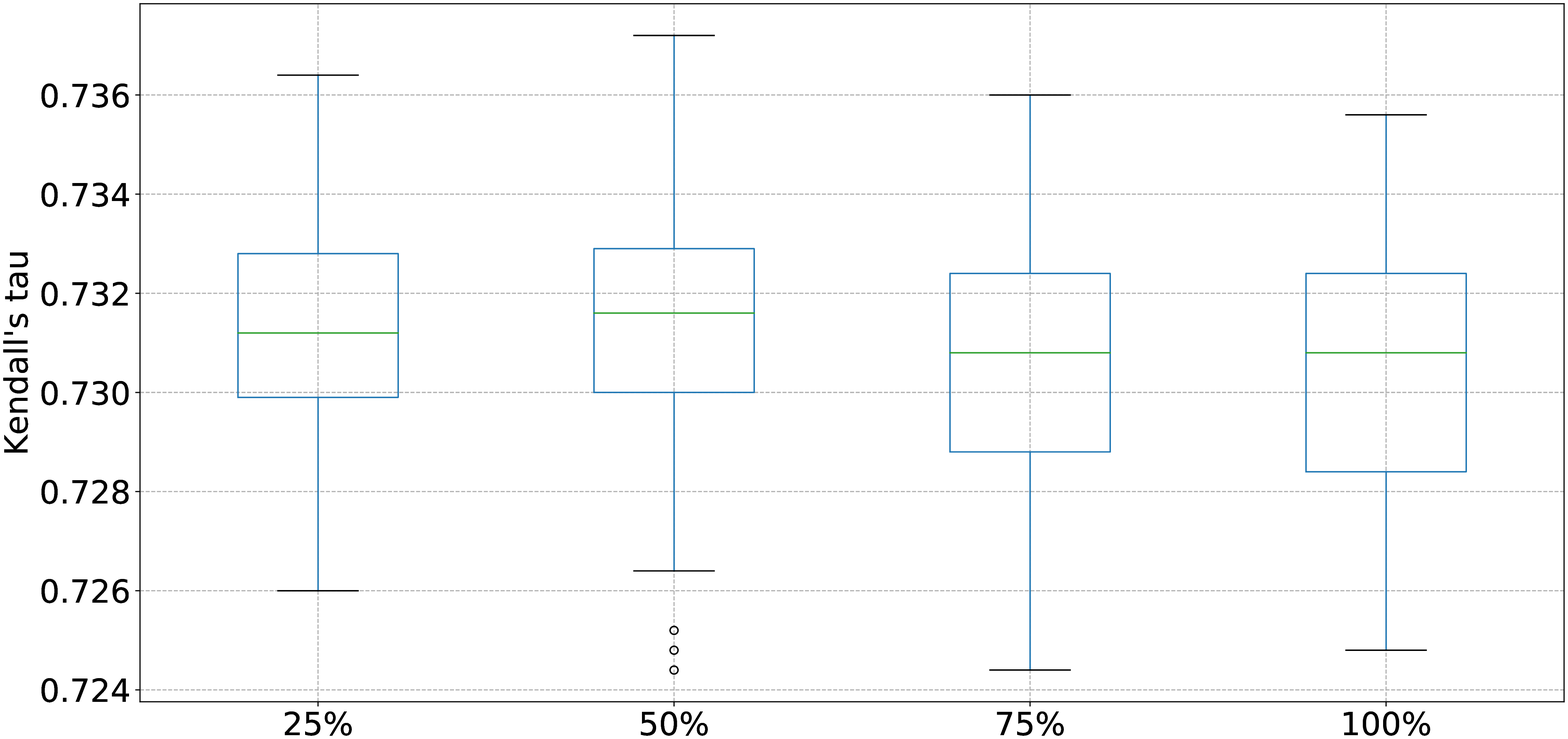}
		\end{minipage}}	
		\caption{\label{figure-9}Distribution of Top-$N$ accuracy and Kendall's tau of splined variant of proposed framework for valued decision examples generated by random value functions.}
	\end{figure}

\begin{figure}[!htbp] 
		\centering
		\subfloat[Accuracy@1]{\begin{minipage}{9cm}
				\includegraphics[width=1\textwidth]{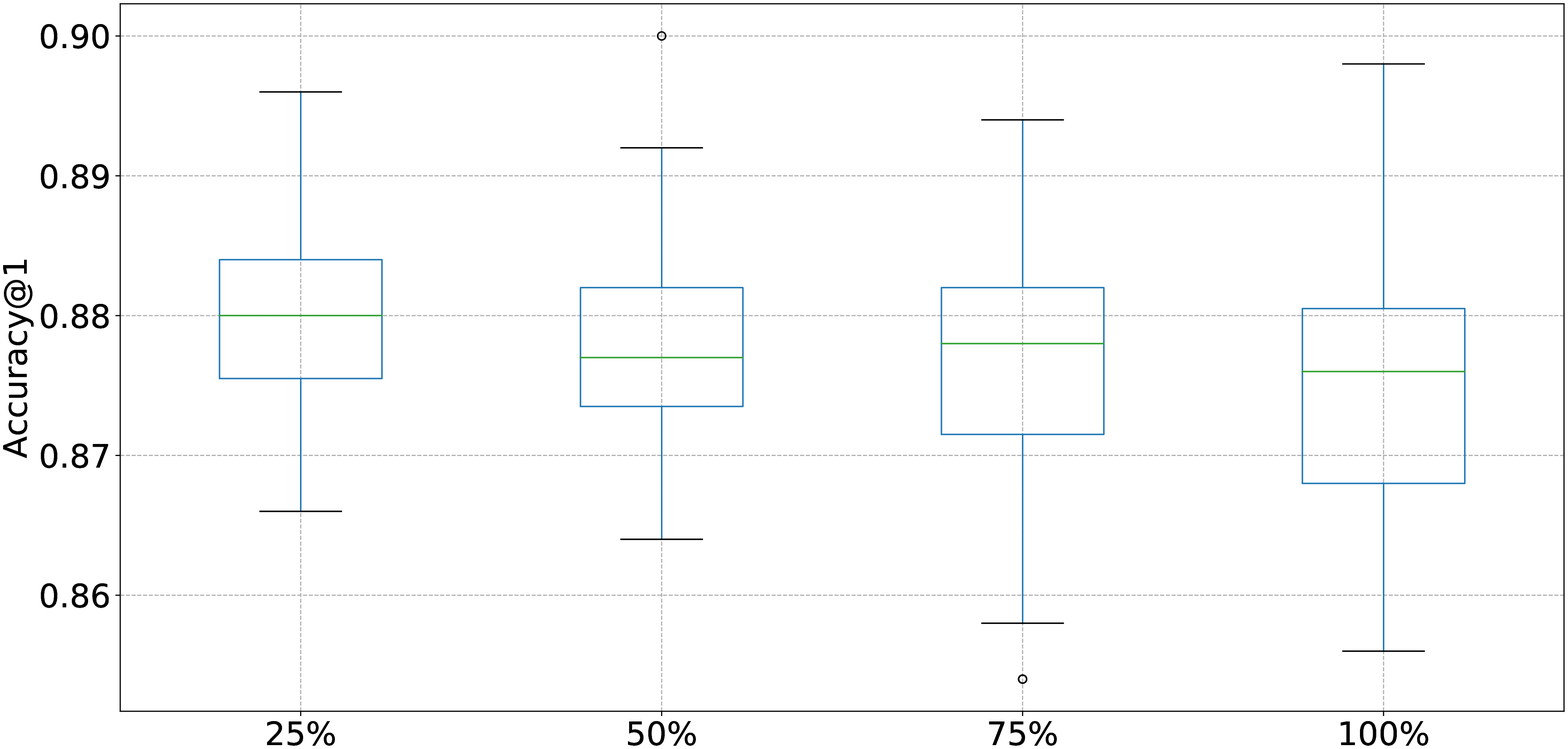}
		\end{minipage}}
		\subfloat[Accuracy@2]{\begin{minipage}{9cm}
				\includegraphics[width=1\textwidth]{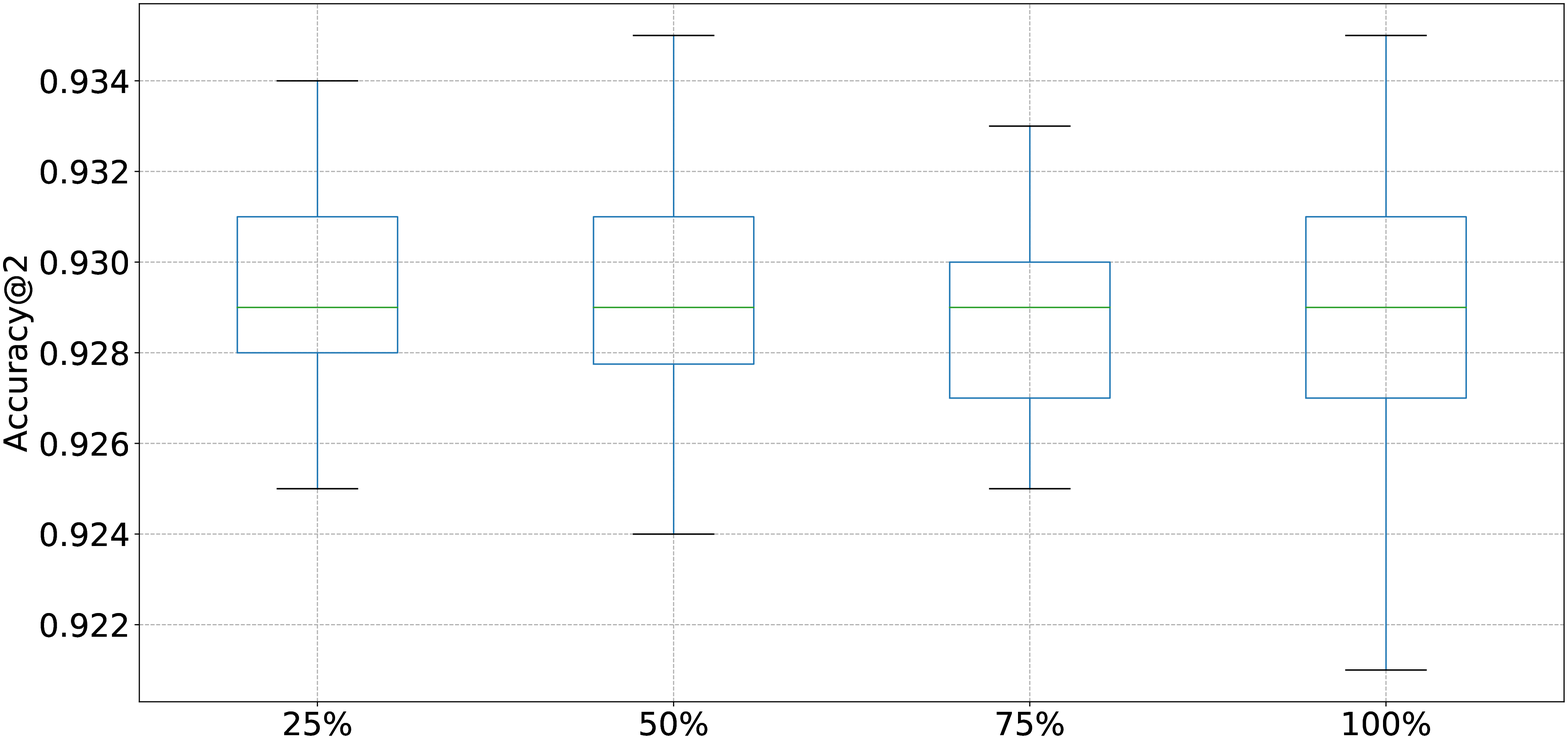}
		\end{minipage}}	
		
		\subfloat[Accuracy@3]{\begin{minipage}{9cm}
				\includegraphics[width=1\textwidth]{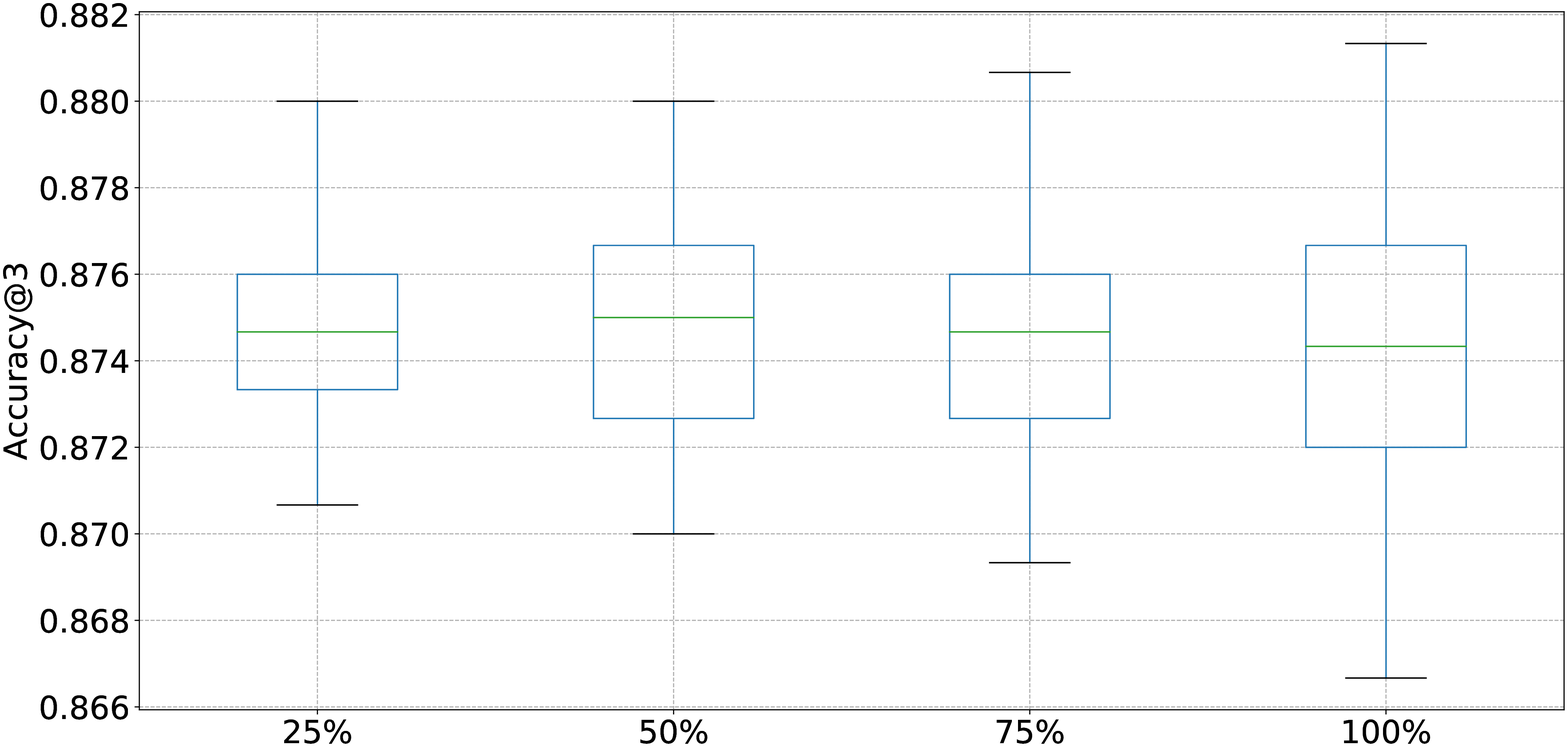}
		\end{minipage}}		
		\subfloat[Kendall's tau]{\begin{minipage}{9cm}
				\includegraphics[width=1\textwidth]{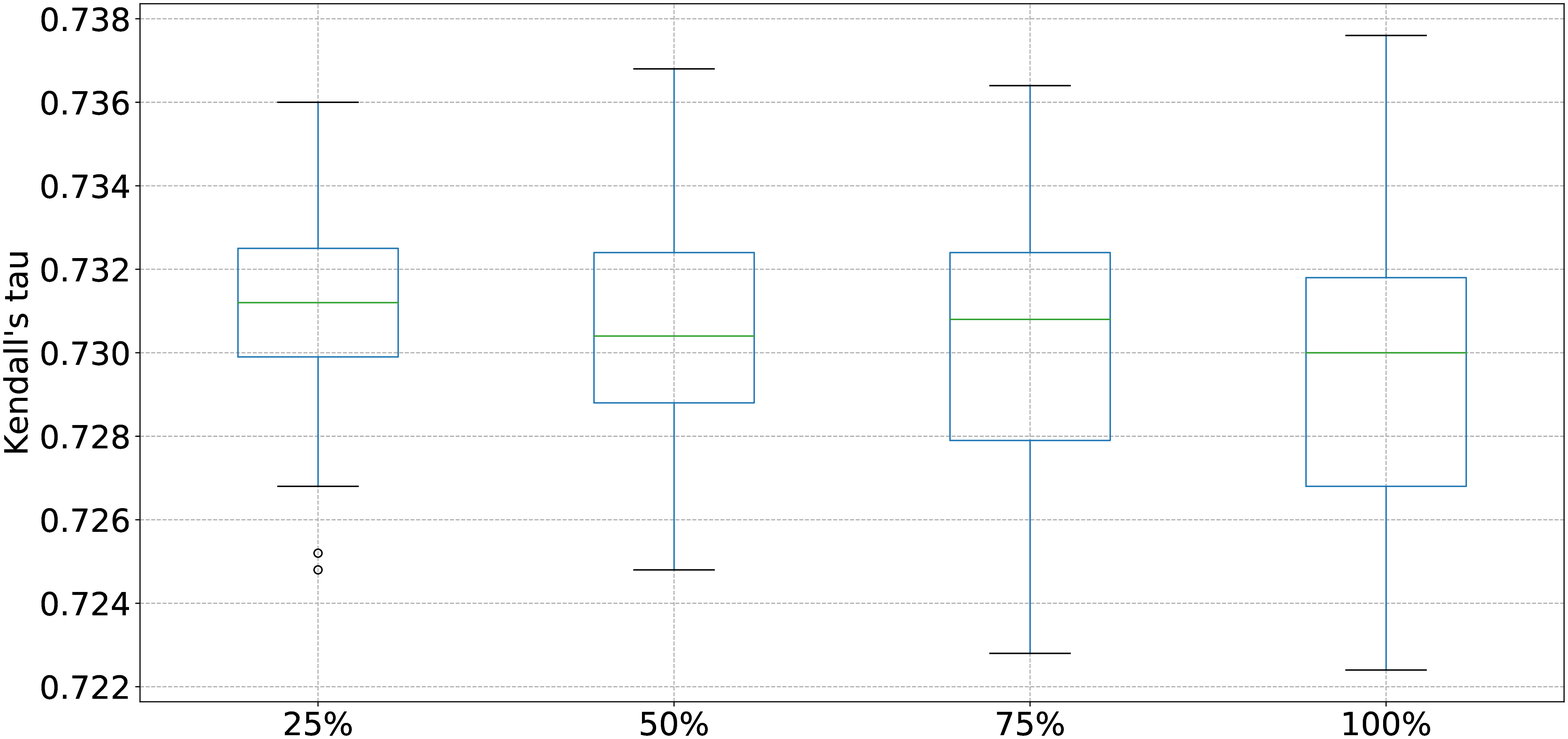}
		\end{minipage}}	
		\caption{\label{figure-10}Distribution of Top-$N$ accuracy and Kendall's tau of general monotone variant of proposed framework for valued decision examples generated by random value functions.}
	\end{figure}

\begin{figure}[!htbp] 
		\centering
		\subfloat[Linear marginal value function]{\begin{minipage}{9cm}
				\includegraphics[width=1\textwidth]{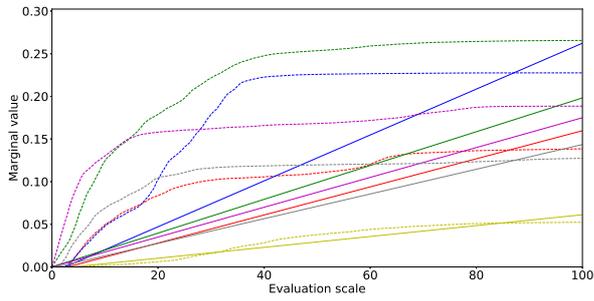}
		\end{minipage}}
		\subfloat[Piecewise-linear marginal value function]{\begin{minipage}{9cm}
				\includegraphics[width=1\textwidth]{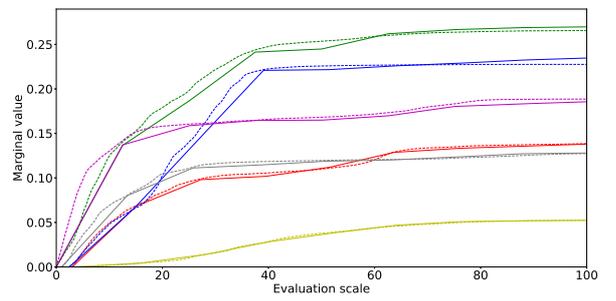}
		\end{minipage}}	
		
		\subfloat[Splined marginal value function]{\begin{minipage}{9cm}
				\includegraphics[width=1\textwidth]{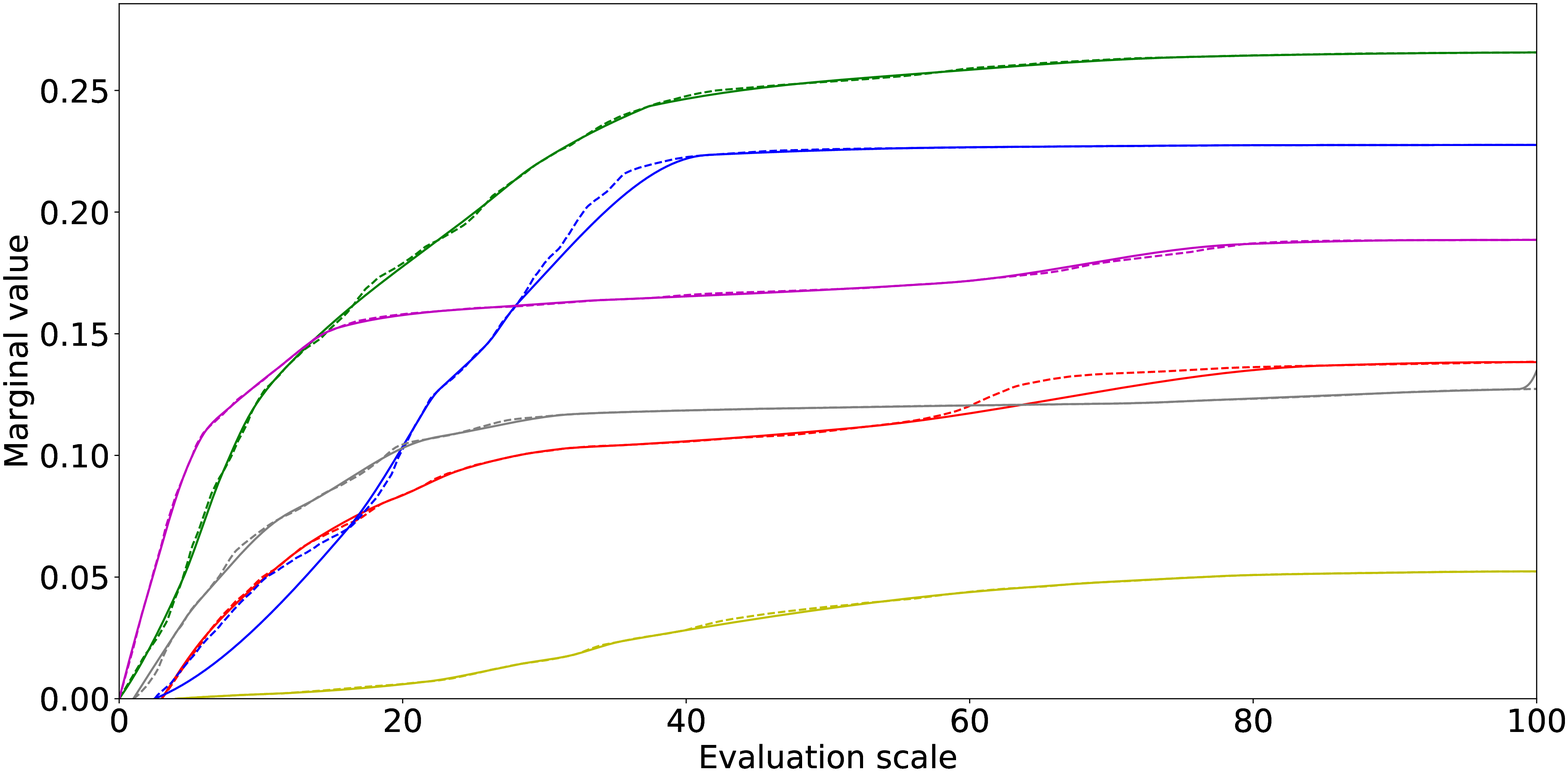}
		\end{minipage}}		
		\subfloat[General monotone marginal value function]{\begin{minipage}{9cm}
				\includegraphics[width=1\textwidth]{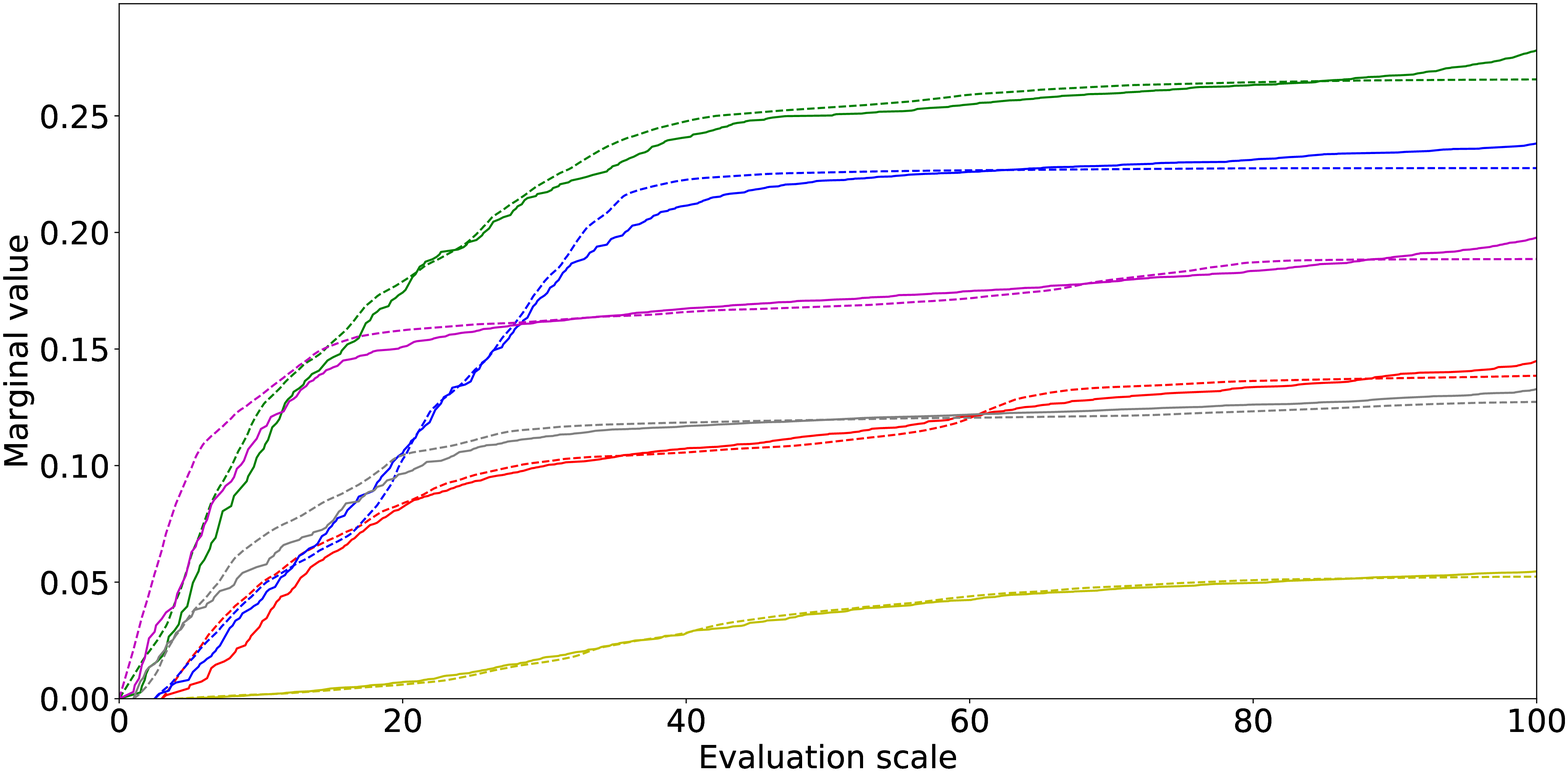}
		\end{minipage}}	
		\caption{\label{figure-11}An example of non-linear actual marginal value functions and marginal value functions derived from four variants of proposed framework. The dashed and solid lines refer to actual marginal value functions and marginal value functions derived from four variants of proposed framework, respectively. The marginal value functions on each criteria are represented using different colors: red -- $g_1$, blue -- $g_2$, yellow -- $g_3$, green -- $g_4$, purple -- $g_5$, gray -- $g_6$. (For interpretation of the references to color in this figure, the reader is referred to the web version of this article.)}
	\end{figure}

	The results of generating random value functions for testing the performance of the four variants of the proposed framework are presented in Table \ref{table-3} and Figures~\ref{figure-7}--\ref{figure-10}. For all variants of the proposed framework, when increasing the complexity of the underlying value function model by selecting a greater $\rho$, the performance of the sorting methods decreases on all evaluation metrics, which is reflected in the decline of mean and the rise of standard deviation. This observation is easy to understand, because a more complex actual value function model makes it difficult for the sorting methods to fit the valued decision examples well. When it comes to the comparison among the four variants of the proposed framework, we observe that the linear variant is significantly outperformed by the others, because its ability to fit non-linear marginal value functions is rather weak. Moreover, it is noted that the piecewise-linear and splined variants slightly outperform the general monotone variant. A~possible reason for this observation can be that the fitting ability of piecewise-linear and splined marginal value functions is sufficient to construct a preference model that is close to the actual one, while the general monotone variant is too flexible and the decision examples are insufficient to help it infer a ``close'' value function since the number is reference alternatives is 350. An example of a randomly generated actual value function and the marginal value functions derived from the four variants of the proposed framework is illustrated in Figure~\ref{figure-11}, where we can compare the actual value function and the constructed ones directly. It is apparent that the constructed linear marginal value functions have great divergence to the actual ones, while the piecewise-linear, splined, and general monotone marginal value functions are similar to the actual ones. Particularly, the piecewise-linear functions have a~significant change in slope at breakpoints in this example, while the splined marginal value functions are completely smooth over the whole performance scales. In this perspective, the splined marginal value functions have the advantage of interpretability and descriptive character.

	\begin{table}[!htbp] \caption{\label{table-3}Top-$N$ accuracy and Kendall's tau in terms of mean and standard deviation of four variants of proposed framework for valued decision examples generated by random value functions.}
		\centering
		\scriptsize
		\begin{tabular}{rccccc}
			\hline
			Method & $\rho$   & Accuracy@1 & Accuracy@2 & Accuracy@3 & Kendall's tau \\
			\hline
			\multirow{4}[0]{*}{Linear variant} & 25\% & 0.6474$\pm$0.0061 & 0.6977$\pm$0.0014 & 0.6446$\pm$0.0027 & 0.4998$\pm$0.0026 \\
			& 50\% & 0.6224$\pm$0.0169 & 0.6982$\pm$0.0029 & 0.6440$\pm$0.0024 & 0.4944$\pm$0.0050 \\
			& 75\% & 0.6093$\pm$0.0178 & 0.6923$\pm$0.0050 & 0.6454$\pm$0.0046 & 0.4896$\pm$0.0046 \\
			& 100\% & 0.5970$\pm$0.0188 & 0.6891$\pm$0.0065 & 0.6471$\pm$0.0067 & 0.4865$\pm$0.0068 \\
			\multicolumn{6}{l}{} \\
			\multirow{4}[0]{*}{Piecewise-linear variant} & 25\% & 0.8811$\pm$0.0056 & 0.9300$\pm$0.0016 & 0.8749$\pm$0.0019 & 0.7318$\pm$0.0020 \\
			& 50\% & 0.8791$\pm$0.0064 & 0.9300$\pm$0.0017 & 0.8745$\pm$0.0021 & 0.7311$\pm$0.0023 \\
			& 75\% & 0.8782$\pm$0.0061 & 0.9293$\pm$0.0018 & 0.8749$\pm$0.0020 & 0.7309$\pm$0.0023 \\
			& 100\% & 0.8764$\pm$0.0073 & 0.9293$\pm$0.0021 & 0.8746$\pm$0.0025 & 0.7302$\pm$0.0029 \\
			\multicolumn{6}{l}{} \\
			\multirow{4}[0]{*}{Splined variant} & 25\% & 0.8801$\pm$0.0058 & 0.9301$\pm$0.0017 & 0.8744$\pm$0.0018 & 0.7312$\pm$0.0019 \\
			& 50\% & 0.8796$\pm$0.0064 & 0.9297$\pm$0.0017 & 0.8748$\pm$0.0019 & 0.7313$\pm$0.0022 \\
			& 75\% & 0.8784$\pm$0.0063 & 0.9294$\pm$0.0017 & 0.8747$\pm$0.0021 & 0.7308$\pm$0.0024 \\
			& 100\% & 0.8770$\pm$0.0067 & 0.9291$\pm$0.0020 & 0.8748$\pm$0.0021 & 0.7304$\pm$0.0026 \\
			\multicolumn{6}{l}{} \\
			\multirow{4}[0]{*}{General monotone variant} & 25\% & 0.8792$\pm$0.0055 & 0.9296$\pm$0.0017 & 0.8747$\pm$0.0019 & 0.7311$\pm$0.0021 \\
			& 50\% & 0.8776$\pm$0.0068 & 0.9290$\pm$0.0021 & 0.8749$\pm$0.0024 & 0.7305$\pm$0.0025 \\
			& 75\% & 0.8765$\pm$0.0082 & 0.9288$\pm$0.0018 & 0.8747$\pm$0.0023 & 0.7301$\pm$0.0029 \\
			& 100\% & 0.8749$\pm$0.0087 & 0.9288$\pm$0.0026 & 0.8744$\pm$0.0030 & 0.7295$\pm$0.0033 \\
			\hline
		\end{tabular}
	\end{table}
	
	\subsection{Accounting for class priorities}
\begin{figure}[!htbp] 
		\centering
		\subfloat[$\text{CardPf}_r$ on reference alternatives]{\begin{minipage}{9cm}
				\includegraphics[width=1\textwidth]{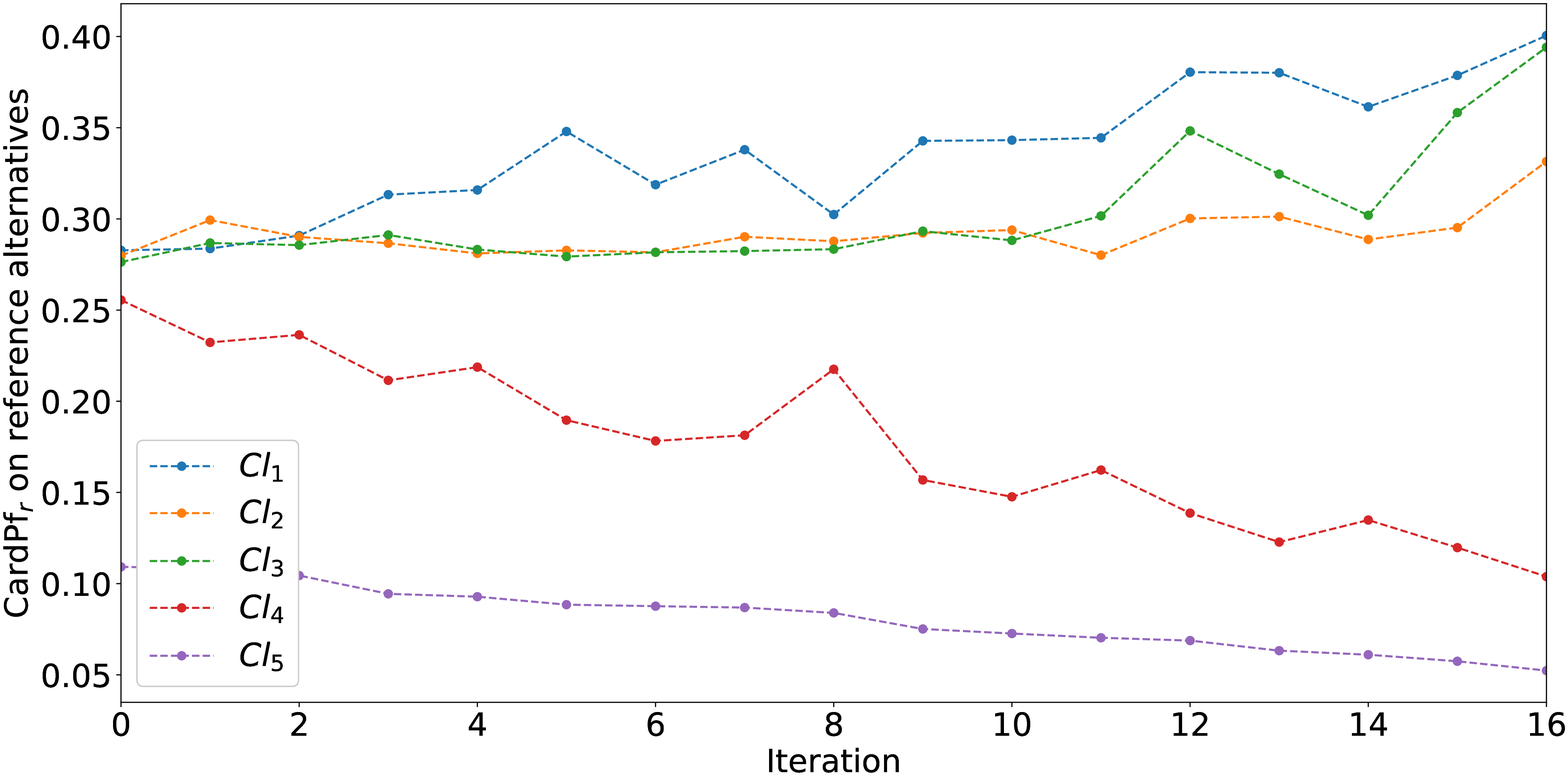}
		\end{minipage}}
		\subfloat[$\text{OrdPf}_r$ on reference alternatives]{\begin{minipage}{9cm}
				\includegraphics[width=1\textwidth]{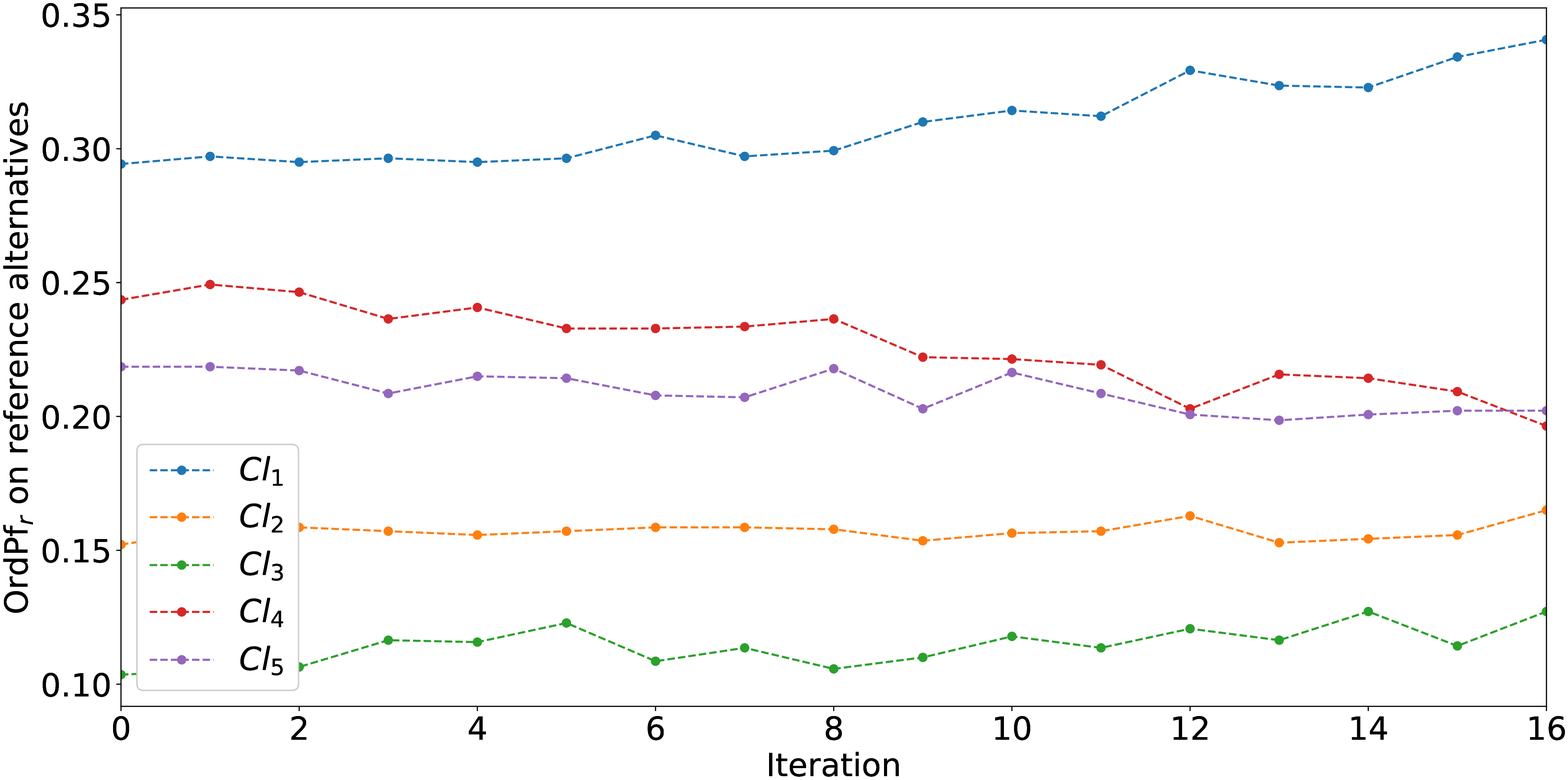}
		\end{minipage}}
		
		\subfloat[$\text{CardPf}_r$ on non-reference alternatives]{\begin{minipage}{9cm}
				\includegraphics[width=1\textwidth]{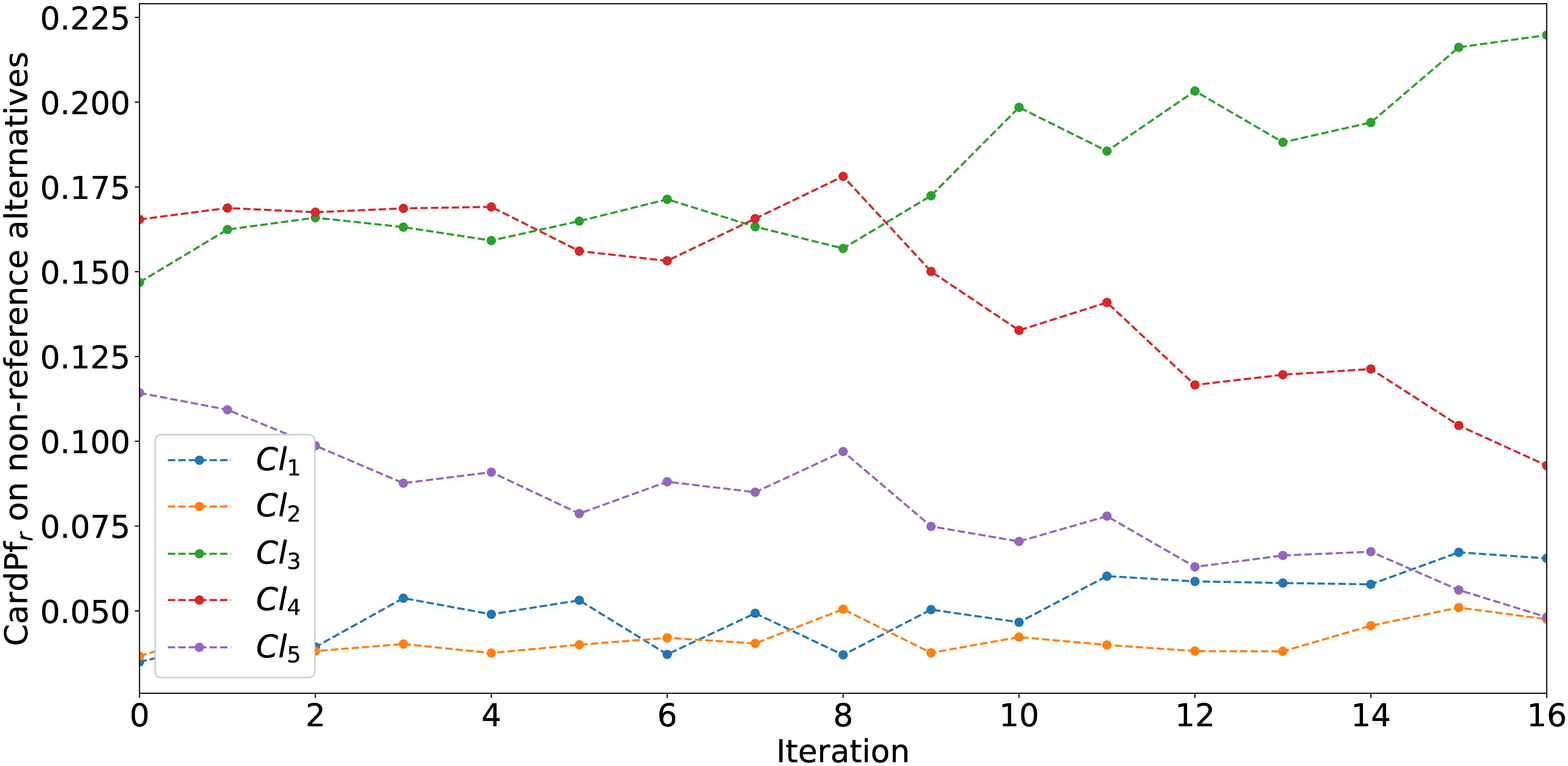}
		\end{minipage}}	
		\subfloat[$\text{OrdPf}_r$ on non-reference alternatives]{\begin{minipage}{9cm}
				\includegraphics[width=1\textwidth]{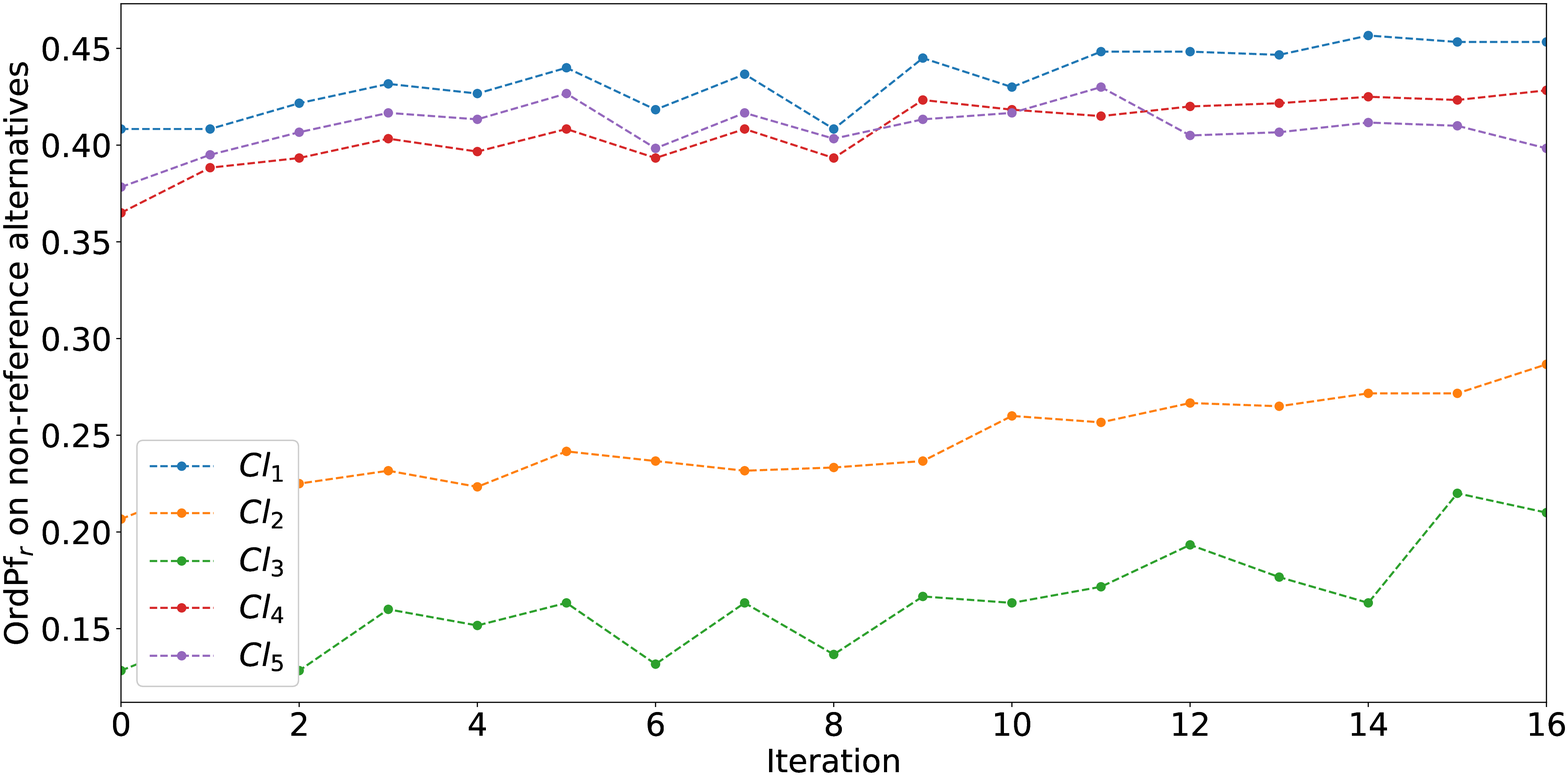}
		\end{minipage}}	
		\caption{\label{figure-12}Variation of $\text{CardPf}_r$ and $\text{OrdPf}_r$ for each class in an example of adjusting classification performance across classes according to class priorities.}
	\end{figure}

	\noindent To illustrate the method for adjusting classification performance across classes according to class priorities, we give an example which is summarized in Table \ref{table-4} and Figure \ref{figure-12}. In this example, the actual preference model is a randomly generated general monotone value function with the setting $\rho=25\%$. Then, we use this preference model to determine the actual assignment for each alternative and construct the training set $A^R$ and the test set~$A^T$. Particularly, we construct valued decision examples by allocating 20\% credibility degree of each reference alternative to the classes adjacent to its actual assignment. Then, we use the splined variant of the proposed framework to derive the initial value function model, according to which the initial performance measures $\text{CardPf}_r$ and $\text{OrdPf}_r$ for each class on the reference and non-reference alternatives are obtained (see columns ``Initial'' in Table \ref{table-4}). In the following procedure for adjusting classification performance across classes, we consider the priority ranking of classes as $Cl_5, Cl_4, Cl_3, Cl_2, Cl_1$, where the classes $Cl_5$ and $Cl_1$ have the greatest and least priorities, respectively. According to the specified class priorities, we apply the proposed method to adjust classification performance across classes, in which threshold $\zeta$ for controlling the complexity of the adjusted preference model is determined using cross-validation by checking $\text{CardPf}_r$ on the two classes with the greatest priorities. Specifically, if we observed no improvement of $\text{CardPf}_r$ on the two classes with the greatest priorities at a certain iteration on the validation set, we terminate the adjustment process. Note that one can choose different measures for using cross-validation to set threshold~$\zeta$. Finally, we obtain the adjusted classification performance across classes (see columns ``Final'' in Table \ref{table-4}). The variation of $\text{CardPf}_r$ and $\text{OrdPf}_r$ for each class during the whole process is depicted in Figure \ref{figure-12}. 
	
	We observe the measures $\text{CardPf}_r$ and $\text{OrdPf}_r$ for each class on the reference alternatives are adjusted according to the specified priority ranking. Particularly, $\text{CardPf}_r$ and $\text{OrdPf}_r$ for $Cl_4$ and $Cl_5$ decrease during the whole process as they have the greatest two priorities, whereas $\text{CardPf}_r$ and $\text{OrdPf}_r$ for $Cl_1$, $Cl_2$ and $Cl_3$ increase because they have relatively low priorities. This indicates that the classification performance on $Cl_1$, $Cl_2$ and $Cl_3$ is sacrificed to improve that on $Cl_4$ and $Cl_5$. When it comes to the performance for each class on the non-reference alternatives, the measure $\text{CardPf}_r$ for each class varies as it does on the reference alternatives, while the measure $\text{OrdPf}_r$ deteriorates slightly for each class. It suggests that the predictive ability of the constructed model for each class is adjusted in terms of the measure $\text{CardPf}_r$, but the performance in terms of the measure $\text{OrdPf}_r$ does not achieve the desired effect.
	
	\begin{table}[!htbp] \caption{\label{table-4}Initial and final $\text{CardPf}_r$ and $\text{OrdPf}_r$ for each class in an example of adjusting classification performance across classes according to class priorities.}
		\centering
		\scriptsize
		\begin{tabular}{rccccc}
			\hline
			\multirow{2}[2]{*}{} & \multirow{2}[2]{*}{Class} & \multicolumn{2}{c}{$\text{CardPf}_r$}       & \multicolumn{2}{c}{$\text{OrdPf}_r$}  \\
			\cmidrule{3-6}      &       & \multicolumn{1}{l}{Initial} & \multicolumn{1}{l}{Final} & \multicolumn{1}{l}{Initial} & \multicolumn{1}{l}{Final} \\
			\hline
			\multirow{5}[0]{*}{Reference 
				alternatives} & $Cl_1$    & 0.2827 & 0.4005 & 0.2942 & 0.3407 \\
			& $Cl_2$    & 0.2797 & 0.3314 & 0.1521 & 0.1651 \\
			& $Cl_3$    & 0.2763 & 0.3941 & 0.1035 & 0.1271 \\
			& $Cl_4$    & 0.2555 & 0.1038 & 0.2435 & 0.1964 \\
			& $Cl_5$    & 0.1092 & 0.0523 & 0.2185 & 0.2021 \\
			\multicolumn{6}{l}{} \\
			\multirow{5}[0]{*}{Non-reference 
				alternatives} & $Cl_1$    & 0.0349 & 0.0654 & 0.4083 & 0.4533 \\
			& $Cl_2$    & 0.0366 & 0.0475 & 0.2066 & 0.2866 \\
			& $Cl_3$    & 0.1468 & 0.2197 & 0.1283 & 0.2100 \\
			& $Cl_4$    & 0.1654 & 0.0928 & 0.3651 & 0.4283 \\
			& $Cl_5$    & 0.1142 & 0.0481 & 0.3783 & 0.3983 \\
			\hline
		\end{tabular}
	\end{table}
	
	In the above example, we observe that the predictive ability of the constructed model for each class cannot necessarily be adjusted according to the specified class priorities. To test the validity of the proposed method for adjusting classification performance across classes, we conduct a further simulation experiment to examine whether the measures $\text{CardPf}_r$ and $\text{OrdPf}_r$ for each class on non-reference alternatives can be adjusted according to the specified class priorities. In this simulation experiment, we simulate 100 general monotone value functions for each complexity level ($\rho$=0\%, 25\%, 50\%, 75\%, 100\%) as the actual preference models to generate valued decision examples with different distribution of credibility degrees (100\%-0\%, 90\%-10\%, 80\%-20\%, 70\%-30\%, 60\%-40\%). We use the four variants of the proposed framework to construct the initial value functions and then use the method for adjusting classification performance across classes. Particularly, we consider all possible rankings of class priorities (since we consider five classes in this problem, there are 120 possible rankings). Moreover, the stopping criterion of the adjustment procedure is either that there is no improvement of $\text{CardPf}_r$ on the two classes with the greatest priorities at a certain iteration on the validation set, or that the maximum iteration (100) is reached. Finally, we count the percentage of problem instances in which the measures $\text{CardPf}_r$ and $\text{OrdPf}_r$ for the classes with the greatest priorities on non-reference alternatives improve, which can be seen as the possibility the proposed method for adjusting classification performance achieves the desired effect. In e-Appendix C, we report the percentage of problem instances in which ${\text{CardP}}{{\text{f}}_r}$ and ${\text{OrdP}}{{\text{f}}_r}$ for classes with the greatest $N$ priorities on non-reference alternatives improve, respectively, in each problem setting, where $N=1,2,3$. Since we have 120 combinations of problem setting regarding three factors ($F_1$: $\rho$, $F_2$: distribution of credibility degrees, and $F_3$: sorting method), we first use a three-way analysis of variance (ANOVA) to analyze the obtained results, in which we do not consider the interactions between factors. The ANOVA results for the percentage of problem instances in which ${\text{CardP}}{{\text{f}}_r}$ and ${\text{OrdP}}{{\text{f}}_r}$ for classes with the greatest $N$ priorities on non-reference alternatives improve, respectively, are presented in Table \ref{table-5}. It is apparent that, among the three considered factors regarding a sorting task, only the factor of sorting method (i.e., the underlying value function model) affects whether the measures $\text{CardPf}_r$ and $\text{OrdPf}_r$ for each class on non-reference alternatives can be adjusted according to the specified class priorities.
	
	\begin{table}\centering
		\begin{threeparttable}[!htbp] \centering
			\caption{\label{table-5}ANOVA results for the percentage of problem instances in which ${\text{CardP}}{{\text{f}}_r}$ and ${\text{OrdP}}{{\text{f}}_r}$ for classes with the greatest $N$ priorities on non-reference alternatives improve, respectively.}	
			\scriptsize
			\begin{tabular}{rrccc}
				\hline
				Outcome & Indicator & $\rho$ & Credibility distribution & Sorting method \\
				\hline
				$\Delta {\text{CardP}}{{\text{f}}_r}@1$ & df    & 4     & 4     & 3 \\
				& Mean squares & 0.002 & 0.003 & 0.564 \\
				& $F$   & 0.541 & 0.943 & 169.804 \\
				& Sig.  & 0.706 & 0.443 & 0.000* \\
				&       &       &       &  \\
				$\Delta {\text{CardP}}{{\text{f}}_r}@2$ & df    & 4     & 4     & 3 \\
				& Mean squares & 0.002 & 0.003 & 0.559 \\
				& $F$   & 0.246 & 0.464 & 80.606 \\
				& Sig.  & 0.911 & 0.762 & 0.000* \\
				&       &       &       &  \\
				$\Delta {\text{CardP}}{{\text{f}}_r}@3$ & df    & 4     & 4     & 3 \\
				& Mean squares & 0.009 & 0.019 & 0.675 \\
				& $F$   & 0.783 & 1.579 & 57.545 \\
				& Sig.  & 0.541 & 0.187 & 0.000* \\
				&       &       &       &  \\
				$\Delta {\text{OrdP}}{{\text{f}}_r}@1$ & df    & 4     & 4     & 3 \\
				& Mean squares & 0.003 & 0.004 & 0.486 \\
				& $F$   & 0.523 & 0.741 & 101.439 \\
				& Sig.  & 0.719 & 0.567 & 0.000* \\
				&       &       &       &  \\
				$\Delta {\text{OrdP}}{{\text{f}}_r}@2$ & df    & 4     & 4     & 3 \\
				& Mean squares & 0.001 & 0.004 & 0.462 \\
				& $F$   & 0.096 & 0.564 & 59.471 \\
				& Sig.  & 0.984 & 0.689 & 0.000* \\
				&       &       &       &  \\
				$\Delta {\text{OrdP}}{{\text{f}}_r}@3$ & df    & 4     & 4     & 3 \\
				& Mean squares & 0.011 & 0.018 & 0.549 \\
				& $F$   & 0.867 & 1.471 & 44.488 \\
				& Sig.  & 0.487 & 0.218 & 0.000* \\
				\hline
			\end{tabular}
			\begin{tablenotes}
				\item[1] $\Delta {\text{CardP}}{{\text{f}}_r}@N$ and $\Delta {\text{OrdP}}{{\text{f}}_r}@N$ for $N=1, 2, 3$ refer to the percentage of problem instances in which ${\text{CardP}}{{\text{f}}_r}$ and ${\text{OrdP}}{{\text{f}}_r}$ for the classes with the greatest $N$ priorities on the non-reference alternatives improve, respectively.
			\end{tablenotes}
		\end{threeparttable}
	\end{table}
	
	Then, we analyze the results of applying the proposed method to adjust the initial outcomes derived by different variants of the proposed framework, which is summarized in Table \ref{table-6}. On the one hand, we observe that both $\Delta {\text{CardP}}{{\text{f}}_r}@N$ and $\Delta {\text{OrdP}}{{\text{f}}_r}@N$ decrease as $N$ increases, which indicates that the chance that the performance on a class is improved decreases with its priority. Particularly, $\Delta {\text{CardP}}{{\text{f}}_r}@N$ is greater than $\Delta {\text{OrdP}}{{\text{f}}_r}@N$ for the same $N$, which means that the measure ${\text{CardP}}{{\text{f}}_r}$ has a higher possibility to be improved than the measure ${\text{OrdP}}{{\text{f}}_r}$. On the other hand, it is noted that the proposed method has a greater possibility to adjust the initial outcomes derived by the piecewise-linear, splined, and general monotone variants of the proposed framework than to adjust the results achieved by the linear counterpart. This is because the piecewise-linear, splined, and general monotone value functions are more flexible while the linear value function is restricted to the linear form.
	
	\begin{table}\centering									
		\begin{threeparttable}[!htbp] \caption{\label{table-6}Percentage of problem instances in which ${\text{CardP}}{{\text{f}}_r}$ and ${\text{OrdP}}{{\text{f}}_r}$ for classes with the greatest $N$ priorities on non-reference alternatives improve, respectively, in terms of mean and standard deviation.}
			\centering
			\scriptsize
			\begin{tabular}{rcccccc}
				\hline
				Method & $\Delta {\text{CardP}}{{\text{f}}_r}@1$ & $\Delta {\text{CardP}}{{\text{f}}_r}@2$ & $\Delta {\text{CardP}}{{\text{f}}_r}@3$ & $\Delta {\text{OrdP}}{{\text{f}}_r}@1$ & $\Delta {\text{OrdP}}{{\text{f}}_r}@2$ & $\Delta {\text{OrdP}}{{\text{f}}_r}@3$ \\
				\hline
				Linear variant & 0.6043$\pm$0.0605 & 0.4951$\pm$0.0896 & 0.3744$\pm$0.1227 & 0.5293$\pm$0.0691 & 0.4212$\pm$0.0885 & 0.3008$\pm$0.1242 \\
				Piecewise-linear variant & 0.9040$\pm$0.0598 & 0.8124$\pm$0.0916 & 0.7288$\pm$0.1179 & 0.7988$\pm$0.0676 & 0.7117$\pm$0.0885 & 0.6200$\pm$0.1199 \\
				Splined variant & 0.9080$\pm$0.0489 & 0.7856$\pm$0.0611 & 0.6856$\pm$0.0862 & 0.8163$\pm$0.0643 & 0.6816$\pm$0.0782 & 0.5771$\pm$0.0892 \\
				General monotone variant & 0.9024$\pm$0.0534 & 0.7807$\pm$0.0709 & 0.6873$\pm$0.0966 & 0.8083$\pm$0.0661 & 0.6819$\pm$0.0801 & 0.5876$\pm$0.1015 \\
				\hline
			\end{tabular}
			\begin{tablenotes}
				\item[1] $\Delta {\text{CardP}}{{\text{f}}_r}@N$ and $\Delta {\text{OrdP}}{{\text{f}}_r}@N$ for $N=1, 2, 3$ refer to the percentage of problem instances in which ${\text{CardP}}{{\text{f}}_r}$ and ${\text{OrdP}}{{\text{f}}_r}$ for the classes with the greatest $N$ priorities on the non-reference alternatives improve, respectively.
			\end{tablenotes}
		\end{threeparttable}
	\end{table}

	\section{Conclusions}
	\label{sec-4}
	
	\noindent In this paper, we proposed a new preference learning framework for constructing an additive value function model from the given decision examples. We put the linear, piecewise-linear, splined, and general monotone value functions under a~unified analytical framework, in which the DM can select any type to equip different variants of the analytical framework. In comparison with the existing sorting methods, our analytical framework allows to consider valued decision examples and each reference alternative could be assigned to multiple classes with respective credibility degrees. We formulated the learning problem within the regularization framework in order to improve the predictive ability of the constructed preference model on new instances. Specifically, we defined the complexity for each type of value function model and use regularization terms avoid the over-fitting problem. Moreover, we introduced the advanced alternating direction method of multipliers to solve the proposed optimization model in a computationally efficient way. In addition, considering the potential lack of equivalence in class priorities, we proposed a method to adjust classification performance across classes according to the priority ranking of classes specified by the DM.
	
	The experimental study of applying the analytical framework to a real-world dataset revealed that the variants using different value functions had a~competitive advantage over the existing sorting methods in terms of a~predictive performance, but each of them had respective characteristics in interpreting human preferences. Specifically, the value function model constructed by the linear variant is easy to explain to a non-experienced DM, but has a~relatively weak ability in learning complex non-linear preferences. As for the piecewise-linear variant, the derived piecewise-linear marginal value functions can fit complex preferences well, but may exhibit a~sudden change in slope at breakpoints, which limits their interpretability in some contexts. When it comes to the splined variant, it achieves the same high classification performance as the piecewise-linear variant, but has the advantage of constructing smooth marginal value functions, which is more appreciated in some applications. Finally, the general monotone variant uses the most flexible value function model that can characterize any general monotone preferences over alternatives. 
	
	We also investigated the variation of classification performance for different credibility distribution of valued decision examples generated by simulated value functions with different degrees of complexity. Moreover, we tested the possibility of using the proposed method for adjusting classification performance to achieve desired outcomes. Overall, the analytical framework is capable of dealing with complex decision problems and reveals flexibility to fulfil personalized requirement from the DM.
	
	Our work contributes to the research at the crossroads of Multiple Criteria Decision Aiding and Machine Learning. On the one hand, we introduced the methodological and computational advances from the machine learning community as efficient tools to address several new characteristics that have never been considered in previous studies in the field of MCDA, including constructing various types of value function models under the unified framework, assigning alternatives to multiple classes with respective credibility degrees, and prioritizing importance of classes in order to adjust classification performance. On the other hand, these new characteristics also provide several possible avenues to the Machine Learning community, such as utilizing the descriptive character of value function models for interpreting the outcomes suggested by a learning procedure, considering consistency-driven procedure for ordinal classification tasks instead of using the traditional probabilistic framework, and allowing the DM to participate in the construction process to obtain customized results. 
	
	We envisage the following directions for future research. The analytical framework can be extended to consider interacting and non-monotonic preferences. Moreover, it will be interesting to apply the analytical framework to multiple criteria ranking or multi-label ranking. We will also incorporate more types of value function models and other preference models into the analytical framework. Finally, more real-world applications are needed to validate the practical performance of the four variants of the analytical framework.	
	
	\section*{References}
	\bibliographystyle{apalike}
	\bibliography{mybibfile}

	
\end{document}